\documentclass[11pt]{article}
\usepackage[letterpaper,left=0.78in,right=0.78in,top=0.72in,bottom=0.78in]{geometry}
\usepackage[utf8]{inputenc}
\usepackage[T1]{fontenc}
\ifdefined\pdfgentounicode
  \input glyphtounicode
\fi

\usepackage{amsmath,amssymb,mathtools,amsthm,bm}
\usepackage{microtype}
\usepackage[numbers,sort&compress]{natbib}
\usepackage{xurl}
\usepackage{graphicx}
\graphicspath{{figures/}}
\usepackage{booktabs,longtable,threeparttable}
\usepackage{xcolor}
\usepackage{colortbl}
\usepackage{float,placeins,needspace}
\usepackage{enumitem,multirow,array}
\usepackage{caption,subcaption}
\usepackage{fancyvrb,fvextra}
\usepackage{hyperref}
\usepackage[capitalize,noabbrev]{cleveref}
\hypersetup{
  pdftitle={Decoupling Solving from Abstention: Single-Rollout Conformal Risk Control for Hard-Budget LLM Reasoning},
  pdfauthor={Qianyi Chen and Bo Li},
  colorlinks=true,linkcolor=blue!55!black,citecolor=green!40!black,urlcolor=blue!65!black
}
\makeatletter
\renewcommand\paragraph{\@startsection{paragraph}{4}{\z@}
  {\z@}{-1em}{\normalfont\normalsize\bfseries}}
\makeatother
\setlist[itemize]{leftmargin=*,itemsep=2pt,topsep=3pt}
\setlist[enumerate]{leftmargin=*,itemsep=2pt,topsep=3pt}
\newcolumntype{L}[1]{>{\raggedright\arraybackslash}p{#1}}
\newcolumntype{C}[1]{>{\centering\arraybackslash}p{#1}}
\theoremstyle{plain}
\newtheorem{theorem}{Theorem}
\newtheorem{proposition}{Proposition}
\newtheorem{lemma}{Lemma}[section]
\newtheorem{corollary}{Corollary}[section]
\theoremstyle{remark}

\DeclareMathOperator{\E}{\mathbb{E}}
\newcommand{\Var}{\operatorname{Var}}
\newcommand{\Cov}{\operatorname{Cov}}
\newcommand{\ind}{\mathbb{I}}
\newcommand{\KL}{D_{\mathrm{KL}}}
\newcommand{\R}{\mathbb R}
\newcommand{\TailCRC}{Tail-CRC}

\title{\Large\bfseries Decoupling Solving from Abstention: Single-Rollout
Conformal Risk Control for Hard-Budget LLM Reasoning}
\author{Qianyi Chen\qquad Bo Li\thanks{Corresponding author.}\\[0.35em]
\normalsize School of Economics and Management, Tsinghua University\\[0.25em]
{\small\ttfamily\{\href{mailto:cqy22@mails.tsinghua.edu.cn}{cqy22@mails},\href{mailto:libo@sem.tsinghua.edu.cn}{libo@sem}\}.tsinghua.edu.cn}\\[0.45em]
\small
\href{https://github.com/Cqyiiii/tail-crc}{\raisebox{-0.12em}{\includegraphics[height=1em]{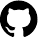}}\,GitHub}
\qquad
\href{https://cqyiiii.github.io/tail-crc/}{\raisebox{-0.12em}{\includegraphics[height=1em]{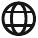}}\,Project Page}}
\date{}
\makeatletter
\renewcommand{\@maketitle}{%
  \newpage\null\vskip0.4em
  \begin{center}
    {\@title\par}
    \vskip1em
    {\large\lineskip0.4em\begin{tabular}[t]{c}\@author\end{tabular}\par}
  \end{center}
  \par\vskip0.35em}
\makeatother
\begin{document}
\maketitle
\begin{abstract}
\noindent
\looseness=-1
Hard token limits often stop large language models (LLMs) before they
produce an answer, while forcing thinking to terminate recovers solutions
but exposes additional mistakes. Reliable budgeted reasoning must therefore
improve both solving and submission decisions. Supervised fine-tuning leaves
deployment error uncontrolled, wrong-answer penalties in reinforcement
learning (RL) can suppress weak solvers' attempts, and multi-sample conformal
methods increase inference compute. We introduce Tail-CRC, which trains a
budget-adapted solver with binary correctness rewards, freezes it, and
applies conformal risk control to the upper-tail token surprise of its
answer span. One rollout supplies the answer and score from returned token
log probabilities, providing finite-sample control of expected wrong
submissions under exchangeability and recalibration without retraining.
Our theory identifies failure-only comparisons induced by penalties,
derives an exact starvation--recovery boundary in an attempt--solver model,
and characterizes length-induced ranking reversal and single-token spike
sensitivity. Across Qwen3.5-4B and Qwen3-8B on DAPO and Omni-MATH, 50 RL
updates on 400 mathematics problems raise correct submissions by
15.0--29.6 and 4.3--12.8 percentage points over inference-only budget
forcing, with 3.5--12.6\% lower generation cost at the same cap.
Target calibration reduces 8B Omni-MATH wrong submissions from
30.3/41.8\% to 5.8/6.5\% at 4K/8K. In a blinded LiveCodeBench study,
the mathematics-trained 8B solver transfers without code training:
correct submissions rise by 7.2/4.5 percentage points over the same-budget
base model, and calibrating both models gives net-score gains of 6.5/5.8
points with lower generation cost. At 8K, calibration reduces its wrong submissions from
11.8\% to 4.8\%, yielding 91.9\% submission precision. Local 8B measurements
at a matched 8K cap further show 12.56\% lower generation time than Native
decoding, combining improved budgeted solving with efficient risk control.
\end{abstract}

\section{Introduction}
\label{sec:intro}

Chain-of-thought (CoT) reasoning and adaptive test-time compute have
substantially expanded what large language models (LLMs) can solve
\citep{wei2022chain,snell2025scaling}, yet real deployments rarely offer an
unbounded token budget. Under a hard output cap, a reasoning model must
convert its intermediate derivation into a finished answer before tokens run
out. This constraint turns inference into a three-way outcome: a correct
submission, a wrong submission, or no submission at all. Standard accuracy
treats the latter two identically, even though withholding an answer and
exposing a confident mistake carry very different costs for downstream users.
Reliable budgeted reasoning therefore requires solving two distinct problems:
producing more correct answers within the cap, and deciding which completed
answers are safe to submit.

\begin{figure}[tb]
 \centering
 \includegraphics[width=\linewidth]{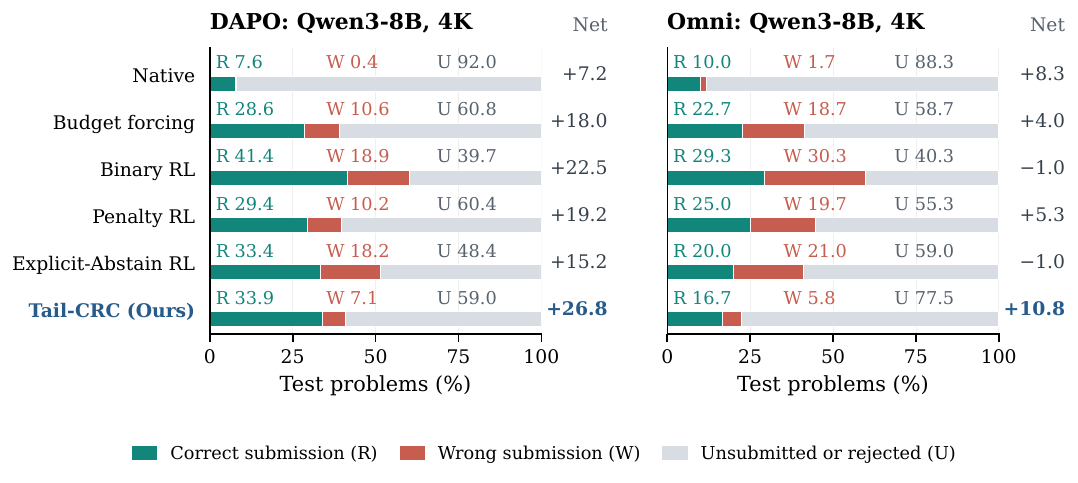}
 \caption{\textbf{Training expands the supply of correct answers; calibration
 controls which ones are submitted.}
 Qwen3-8B~\citep{yang2025qwen3} at a 4K token budget on the 500-problem DAPO and 300-problem Omni-MATH~\citep{gao2025omni} test sets.
 \emph{Native} is uninterrupted base-model generation; \emph{Budget forcing}
 applies s1-style thinking termination~\citep{muennighoff2025s1}.
 \emph{Binary RL} trains with GRPO Done Right (Dr.\ GRPO) using binary
 correctness rewards~\citep{liu2025drgrpo}; \emph{Penalty RL} adds a $-1$
 wrong-answer penalty, and \emph{Explicit-Abstain RL} additionally allows an
 explicit abstention action. The three outcomes are correct submissions ($R$), wrong submissions ($W$),
 and unsubmitted or rejected outcomes ($U$). $U$ combines budget exhaustion,
 invalid format, explicit abstention (0.6\% on DAPO and 10.0\% on Omni for
 Explicit-Abstain RL), and post-hoc rejection. \emph{Tail-CRC (Ours)}
 calibrates the frozen Binary RL solver at $\alpha=0.065$. It preserves
 Binary RL's candidate distribution and rejects candidates after generation
 on 19.3\%/37.2\% of DAPO/Omni problems; the underlying generation-time
 non-submission shares remain 39.7\%/40.3\%. Uniform annotations report
 $R$, $W$, and $U$; right-hand values give net score ($R-W$). Tail-CRC
 increases correct submissions and coverage while reducing wrong submissions
 relative to Budget forcing and Penalty RL on DAPO; on Omni, it reduces
 wrong submissions to 5.8\% and yields the highest displayed net score.
 Table~\ref{tab:audited-penalty} compares calibrated competitors.}
 \label{fig:teaser}
\end{figure}

We evaluate Qwen3.5-4B and Qwen3-8B
\citep{qwen2026qwen35,yang2025qwen3} on two verifiable mathematics domains
and test transfer in a blinded cross-domain code-generation study on
LiveCodeBench~\citep{jain2025livecodebench}. The mathematics results
highlight how sharply solving and submission goals collide.
At a 4K token cap, uninterrupted Native decoding exhausts the budget
without submitting an answer on 88.0--99.2\% of problems across two
mathematics benchmarks. Following s1's budget-forcing strategy
\citep{muennighoff2025s1}, injecting an end-of-thinking delimiter reserves
tokens for answer synthesis and immediately recovers many solutions: on
Qwen3-8B DAPO~\citep{yu2025dapo}, correct submissions jump from 7.6\% to
28.6\%. At the same time, however, wrong submissions surge more than
twentyfold, from 0.4\% to 10.6\% (Figure~\ref{fig:teaser}). Budget forcing
makes many more answers available, but its error exposure turns reliable
submission into a separate deployment problem.

Existing remedies struggle because they either leave deployment risk
uncontrolled or pay a steep price in capability or compute. Supervised
fine-tuning (SFT) can teach a model to reason from worked examples
\citep{muennighoff2025s1}, but it leaves the deployment error rate
uncontrolled. Incorporating wrong-answer penalties or
abstention rewards into reinforcement learning (RL) couples caution directly
into training~\citep{mohamadi2025hesitation}; under a hard cap, however, a
weak solver can avoid penalties simply by exhausting the budget, starving
itself of the very attempts needed to improve. Conformal methods can
calibrate a rule for constructing candidate sets or aggregating reasoning
paths~\citep{quach2024clm,gu2026pause}. Each additional sample adds
inference cost.

\textbf{We instead decouple learning to solve from deciding when to abstain.}
We introduce \textbf{Tail-CRC}, a two-stage, single-rollout framework that
keeps candidate generation and risk control separate. In the first stage, we
train the solver inside the budget-forcing environment using RL with pure
binary correctness rewards, encouraging it to finish as many correct
solutions as possible within the cap. We then freeze the solver weights and
apply conformal risk control (CRC)~\citep{angelopoulos2024crc} to an
uncertainty score computed from the upper tail of token surprises across the
answer span. Because a single rollout supplies both the candidate answer and
its score using only the token log probabilities routinely returned by hosted
APIs, deployment requires no extra model calls and no auxiliary verifier.
Under exchangeability, CRC guarantees finite-sample control of expected wrong
submissions regardless of solver accuracy. Because calibration is decoupled
from training, operators can change risk targets by adjusting the cutoff,
or adapt to a new domain by recalibrating on a few hundred labeled target
problems, without retraining the solver.

The three-state accounting in Figure~\ref{fig:teaser} combines generation-time
non-submission and post-hoc rejection in $U$, but these arise at distinct decision
stages. Penalty RL and Explicit-Abstain RL leave more problems without a valid
answer during generation, whereas Tail-CRC preserves Binary RL's entire candidate
distribution and filters high-score candidates only after generation.
On DAPO at 4K, Tail-CRC actually submits \emph{more} answers overall than
both Budget forcing and Penalty RL (41.0\% vs.\ 39.2\% and 39.6\% coverage),
yet achieves higher correct submissions (33.9\% vs.\ 28.6\% and 29.4\%) and
lower wrong submissions (7.1\% vs.\ 10.6\% and 10.2\%)---strictly dominating
both baselines. On the harder out-of-domain Omni-MATH benchmark, Budget forcing and the
unfiltered RL policies expose wrong answers on roughly one in five to nearly
one in three problems (18.7--30.3\%). There, Tail-CRC targets a 6.5\%
wrong-submission rate by withholding more predictions, cutting errors to
5.8\% and more than doubling net score ($R-W$) over Penalty RL
(10.8\% vs.\ 5.3\%). When
every penalty and explicit-abstention baseline is granted the same
post-hoc tail-score calibration, Binary RL with Tail-CRC still achieves the
highest net score across all twelve matched 4K comparisons
(Section~\ref{sec:audited-results}, Table~\ref{tab:audited-penalty}).

Both halves of this design are grounded in concrete failure modes. On the
training side, we show exactly how wrong-answer penalties activate
failure-only group comparisons in group-relative RL, and derive a sharp
exploration-starvation boundary in an attempt--solver model explaining how
premature penalties can halt improvement. On the scoring side, relying only on
final boxed-answer likelihood can miss uncertainty in the derivation and favor
short wrong integers over longer correct symbolic expressions, while summing
negative log likelihood (NLL) over the answer span introduces a length
multiplier that can invert rankings altogether. Averaging the largest 10\% of
token surprises across the answer span removes this explicit length factor
and limits the influence of any single lexical spike. As illustrated in
Section~\ref{sec:score-theory} (Figure~\ref{fig:theory-intuition}(b)), this
allows Tail-CRC to preserve a correct radical answer rejected by Box NLL
while filtering out a wrong integer that Box NLL confidently accepts.

The paper makes three contributions:
\begin{itemize}
 \item \textbf{A decoupled, single-rollout framework for hard-budget LLM reasoning.}
 Tail-CRC separates binary-reward budget adaptation from answer-span CRC,
 providing finite-sample wrong-submission guarantees and retraining-free
 risk adaptation using one rollout's returned token log probabilities.
 \item \textbf{Theoretical mechanisms for training and single-rollout scoring.}
 Exact group-advantage identities and an analytically solved attempt--solver
 flow explain penalty-induced exploration starvation, while score-anatomy
 results isolate length-induced ranking reversal and single-token spike
 sensitivity (complete proofs in Appendix~\ref{app:proofs}).
 \item \textbf{Empirical gains in accuracy, controlled risk, and inference efficiency.}
 Across Qwen3.5-4B and Qwen3-8B, two mathematics domains, and 4K/8K budgets, 50 RL
 updates improve correct submissions by 15.0--29.6 and 4.3--12.8 percentage
 points over budget forcing while lowering generation cost by 3.5--12.6\%;
 target calibration cuts 8B Omni-MATH errors from 30.3/41.8\% to 5.8/6.5\%.
 Without code training, the same 8B policies improve LiveCodeBench correct
 submissions by 7.2/4.5 points and calibrated net score by 6.5/5.8 points
 at 4K/8K. At 8K, 8B takes 12.56\% less time than Native.
\end{itemize}

\section{Problem Formulation}
\label{sec:setting}
We study a reasoner that receives one problem, generates one trajectory
within a hard token budget, and either submits its answer or withholds it.
The objective is to retain as many correct submissions as possible at a
chosen wrong-submission risk, with the generation cost of one rollout.

\paragraph{Budgeted generation.}
Let $x$ be a problem, $z^\star$ its reference answer, and $\pi_\theta$ an
autoregressive model. A fixed stochastic decoder
$K_{\theta,B}(\cdot\mid x)$ generates trajectory $y$ under total token cap
$B$. \emph{Native} decoding runs continuously; \emph{budget forcing}
reserves tokens for answer synthesis by inserting the end-of-thinking
delimiter \texttt{</think>} if thinking exceeds its allocation.
The \emph{answer span} contains all sampled explanation and answer tokens
after the first end-of-thinking delimiter and before end-of-sequence (EOS),
whether thinking ended naturally or externally. Injected tokens and EOS
are excluded from scoring. Exact 4K/8K allocations and early-ending rules
are specified in Appendix~\ref{app:protocols}.

\paragraph{Outcomes and submission.}
The observable indicator $v(x,y)$ records an EOS-terminated, extractable
answer with consistent boxed contents. A valid candidate is right ($Y=R$)
or wrong ($Y=W$) by mathematical equivalence; truncation, missing or
conflicting answers, and explicit abstention give $Y=U$. Labels require a
reference answer during calibration and evaluation; validity and deployment
selection do not. For a nonconformity score $S(x,y)$, larger means more
uncertain, and invalid or empty spans receive $S=+\infty$. We submit iff
\begin{equation}
 g_\tau(x,y)=\ind\{v(x,y)=1,\ S(x,y)<\tau\}.
 \label{eq:submission-rule}
\end{equation}
Define $R_\tau=P(Y=R,g_\tau=1)$, $W_\tau=P(Y=W,g_\tau=1)$, and
$U_\tau=1-R_\tau-W_\tau$. Suppressing the cutoff subscript, coverage,
net score, precision, and selective error are
\begin{equation}
 C=R+W,\qquad \mathrm{Net}=R-W,\qquad
 \mathrm{Prec}=\frac{R}{R+W},\qquad r_{\rm sel}=\frac{W}{R+W}.
 \label{eq:metrics}
\end{equation}
Marginal rates use all problem families; precision and selective error use
submitted answers. Generation cost includes rejected trajectories.
Rates are percentages; differences are percentage points (pp).

\paragraph{Calibration target.}
The loss $L_\tau=\ind\{Y=W,S<\tau\}$ records an exposed wrong answer.
With the solver, decoder, validity rule, and score frozen independently
of calibration, we select a cutoff from one rollout per labeled family so
that
\begin{equation}
 \E_{\mathcal D_{\rm cal},\,{\rm test}}[L_{\widehat\tau}]\le\alpha.
 \label{eq:target-risk}
\end{equation}
This expected marginal risk averages over calibration, a new test family,
and stochastic generation; it differs from the error rate conditional on
submission. Sections~\ref{sec:pipeline} and~\ref{sec:guarantees} give the
construction and guarantee.

\section{Tail-CRC: Method and Theory}
\label{sec:method}

\subsection{Learning to solve before choosing when to submit}
\label{sec:pipeline}

\begin{figure}[tb]
\centering
\includegraphics[width=\linewidth]{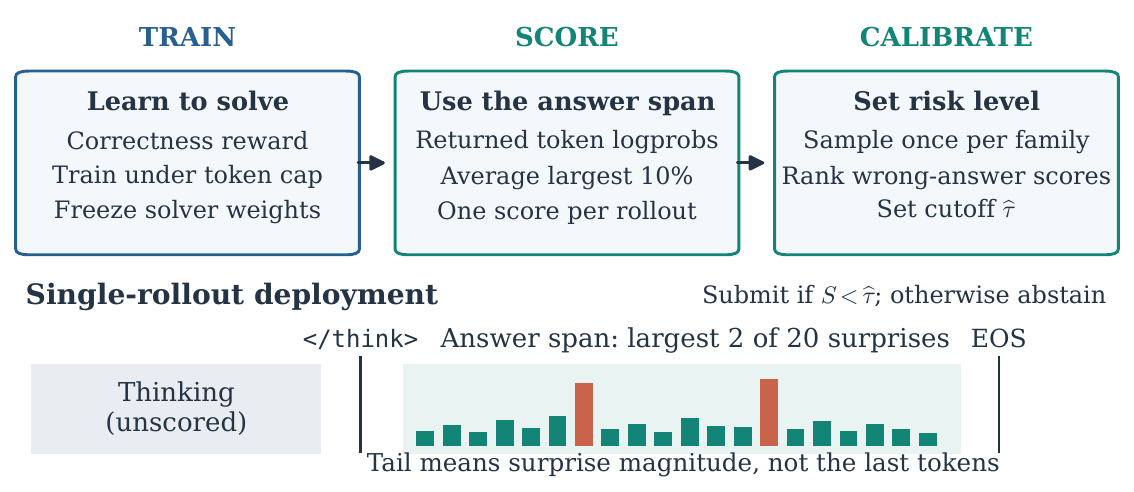}
\caption{\textbf{Train, score, and calibrate with Tail-CRC.}
Binary RL adapts the solver to the hard budget; a fixed answer-span score
then measures token surprise. Labeled calibration families determine a
cutoff for that score. Deployment generates one rollout and either
submits or abstains. The answer span
begins when thinking ends, naturally or through budget forcing; tail
aggregation selects surprises across that span. Calibration is offline.}
\label{fig:method-overview}
\end{figure}

\TailCRC\ first trains a solver to produce correct answers within the hard
budget, then calibrates which candidates to submit. Separating these stages
lets the submission rule adapt to a new risk target or population while
keeping candidate generation fixed. Figure~\ref{fig:method-overview}
summarizes the pipeline.

Binary RL uses terminal reward $r_0(Y)=\ind\{Y=R\}$ in the same two-stage
budget-forcing environment used at deployment. Dr.\ GRPO~\citep{liu2025drgrpo},
a variant of Group Relative Policy Optimization (GRPO)~\citep{shao2024deepseekmath},
centers each group of $G$ rollout rewards,
$A_g=r_g-G^{-1}\sum_jr_j$, without standard-deviation normalization.
Sampled thinking and answer tokens receive the terminal advantage with a
fixed token denominator; deterministic prompt and injected delimiter
tokens are masked. We freeze the final solver after training.
The objective rewards a successful budgeted trajectory as a whole. Giving
both sampled phases its terminal advantage allows training to improve the
partial derivation and its final synthesis together. The learned solver
thus supplies candidates before any deployment-specific choice about how
many of their errors to expose.

A boxed answer can be a single confident token, whereas its preceding
answer span exposes the transitions used to construct it. Let $\ell_t$
be the negative log likelihood (NLL) of sampled answer token $t$, after the
first end-of-thinking delimiter and before EOS. Natural termination and
budget forcing use the same span. Order the $L$ token surprises as
$\ell_{(1)}\ge\cdots\ge\ell_{(L)}$; Tail NLL is
\begin{equation}
 S=\frac1k\sum_{t=1}^k\ell_{(t)},\qquad k=\lceil0.1L\rceil.
 \label{eq:tail-score}
\end{equation}
The tail contains the largest surprises anywhere in the answer span.
Invalid outputs and empty spans receive $S=+\infty$.
The API version uses token log probabilities from the generated trajectory;
Tail entropy instead uses full-vocabulary entropy at the same prefixes
(Appendix~\ref{app:surprise-proof}).

With the solver, decoder, and score frozen, calibration generates and
grades one trajectory per held-out family and applies standard CRC to
choose a cutoff. Deployment generates one trajectory and submits it when
it is valid and $S<\widehat\tau$. References are needed for calibration;
submission uses only the output and score. Changing the risk target
changes the cutoff without retraining or extra generation.

\subsection{Why binary rewards preserve learning opportunities}
\label{sec:penalties}

Under a hard cap, avoiding a wrong answer can mean exhausting the budget
without submitting. To identify the training signal behind this behavior,
consider Penalty RL rewards $r_\lambda(R)=1$, $r_\lambda(W)=-\lambda$,
and $r_\lambda(U)=0$, for $\lambda\ge0$. Probabilities below refer to the
actual stochastic rollout distribution.

\begin{proposition}[Penalty signal and failure-only comparisons]
\label{prop:penalty-signal}
For $G\ge2$ independent rollouts of a fixed prompt with probabilities
$(p_R,p_W,p_U)$, set $\gamma_G=(G-1)/G$. For outcomes of positive probability,
\begin{align}
 \E[A_1\mid R]&=\gamma_G(1-p_R+\lambda p_W),\nonumber\\
 \E[A_1\mid W]&=-\gamma_G\{\lambda(1-p_W)+p_R\},\nonumber\\
 \E[A_1\mid U]&=\gamma_G(\lambda p_W-p_R).
 \label{eq:conditional-group-advantage}
\end{align}
When $p_R+p_W>0$, the last expression is positive exactly when
$p_R/(p_R+p_W)<\lambda/(1+\lambda)$.
For $\lambda>0$, a reward-varying group with no correct rollout has probability
\begin{equation}
 (1-p_R)^G-p_W^G-p_U^G.
 \label{eq:failure-only-probability}
\end{equation}
Every such group has zero advantages under binary reward. For any
integrable per-rollout feature $\psi$,
\begin{equation}
 \E\!\left[G^{-1}\sum_g A_g\psi_g\right]
 =\gamma_G\Cov(r_\lambda,\psi).
 \label{eq:group-covariance}
\end{equation}
\end{proposition}
\emph{Proof.} Appendix~\ref{app:penalty-proof} derives all three identities.

Non-submission has positive expected advantage below 50\% submitted-answer
precision for $\lambda=1$, or 20\% for $\lambda=0.25$.
Failure-only groups then favor $U$ over wrong attempts without containing a
successful solution. Binary reward gives the whole group zero advantage.
The comparison is prompt-specific: a difficult prompt can fall below the
break-even precision even when the benchmark average is higher.
A common non-submission reward $0\le\rho<1$ is equivalent to effective
penalty $(\lambda+\rho)/(1-\rho)$ \emph{up to positive reward rescaling},
which group centering retains (Corollary~\ref{cor:positive-u-main}).

An explanatory attempt--solver model connects this signal to learning
opportunities. Let $a$ be attempt probability and $q$ success probability
given an attempt, so $(p_R,p_W,p_U)=(aq,a(1-q),1-a)$. Attempts decline
below break-even precision, while solver improvement slows with fewer
attempts. The next theorem solves this interaction exactly in a product
Bernoulli mirror-gradient model, with $\kappa$ scaling solver learning
relative to the attempt decision.

\begin{theorem}[Starvation--recovery dichotomy in an attempt--solver flow]
\label{thm:starvation-boundary}
Fix $\lambda\ge0$, $\kappa>0$, and $(a_0,q_0)\in(0,1)^2$. Consider
the flow
\begin{equation}
 \dot a=a(1-a)\{(1+\lambda)q-\lambda\},\qquad
 \dot q=\kappa(1+\lambda)a q(1-q).
 \label{eq:attempt-solver-main}
\end{equation}
For $\lambda>0$, put $q_c=\lambda/(1+\lambda)$. Starvation occurs
if and only if $q_0<q_c$ and
\begin{equation}
 D_{\rm Ber}(q_c\Vert q_0)\ge-\kappa\log(1-a_0),
 \label{eq:barrier-main}
\end{equation}
where $D_{\rm Ber}$ is Bernoulli Kullback--Leibler (KL) divergence.
In this regime, $a(t)\to0$ and $q(t)\to q_\infty\le q_c$, and
both $\int_0^\infty a(t)\,dt$ and $\int_0^\infty a(t)q(t)\,dt$ are finite.
Every other interior initial state converges to $(a,q)=(1,1)$.
With binary reward ($\lambda=0$), every interior initial state also
converges to $(1,1)$.
\end{theorem}
\emph{Proof.} Appendix~\ref{app:attempt-solver-proof} derives the invariant,
the boundary cases, and the finite-exposure identities.

The boundary quantifies the race between improving the solver and
suppressing attempts: a larger initial distance to break-even precision
requires a larger attempt probability or faster solver learning.

Figure~\ref{fig:theory-intuition}(a) shows the exact boundary at
$\lambda=1$, $\kappa=0.5$: for $q_0=0.2$, recovery requires $a_0>0.36$.
Starting at $a_0=0.25$ gives vanishing attempts and
$q_\infty\approx0.309$; starting at $a_0=0.50$ leads to recovery.
Appendix Figure~\ref{fig:penalty-barrier} compares boundaries for
$\kappa\in\{0.25,0.5,1\}$ and shows recovery and starvation trajectories
at $\kappa=0.5$.
The model explains how attempts can disappear before solver quality
crosses break-even precision. The reward ablation tests the corresponding
observable pattern: Penalty RL produces more truncation, fewer correct
submissions, and longer rollouts (Table~\ref{tab:audited-penalty}).

Equal rewards for explicit abstention and truncation express no preference
between them: moving probability between these outcomes at fixed
$(p_R,p_W)$ leaves the objective unchanged. Rewarding explicit refusal
alone would break that tie. Deployment selection is a separate choice:
the optimal correctness threshold generally depends on both the target
and population, even when posterior correctness is unchanged.
Proposition~\ref{prop:penalty-selection} in Appendix~\ref{app:selection-proof}
formalizes this dependence. \TailCRC\ adjusts submission by recalibration;
we compare training rewards after matched calibration.

\subsection{Why score the upper tail of the answer span?}
\label{sec:score-theory}

With the solver fixed, selection depends on score orderings. Box NLL omits the reasoning leading
to an answer; Sum NLL incorporates that span but also its length; a maximum
can be dominated by one unusual token. The following identities isolate
these effects. For independent valid wrong and right draws, the area under
the receiver operating characteristic curve (AUROC) is
$\operatorname{AUC}(S)=P(S_W>S_R)+\tfrac12P(S_W=S_R)$, with wrong answers
as the positive class.

\begin{proposition}[Length factor and ranking identity]
\label{prop:length-discrimination}
For a nonempty span, $S_{\rm sum}=L\bar\ell$. With positive average
surprises, define $D=\log\bar\ell_W-\log\bar\ell_R$,
$B_\ell=\log L_R-\log L_W$, and
$\psi(u)=\ind\{u>0\}+\tfrac12\ind\{u=0\}$.
\begin{equation}
 \operatorname{AUC}(S_{\rm sum})=\E\psi(D-B_\ell),\qquad
 \operatorname{AUC}(\bar\ell)=\E\psi(D).
 \label{eq:exact-length-auroc}
\end{equation}
\end{proposition}
\emph{Proof.} Appendix~\ref{app:length-proof} factors the pairwise comparison
and gives a Gaussian surrogate for ranking reversal.

\begin{figure}[t]
\centering
\includegraphics[width=\linewidth]{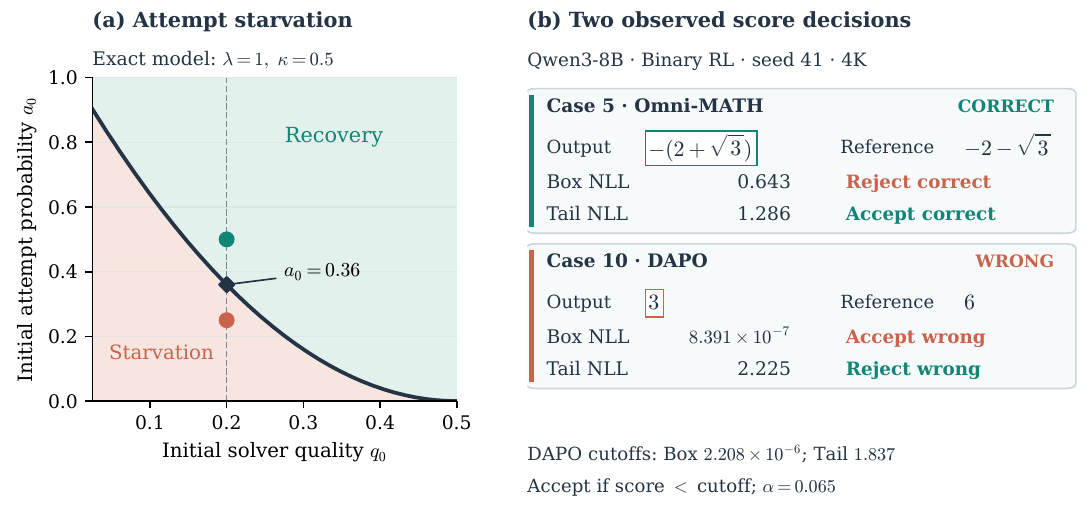}
\caption{\textbf{A learning mechanism and two observed scoring decisions.}
\textbf{(a)} The explanatory attempt--solver flow at $\lambda=1$,
$\kappa=0.5$ has an exact recovery--starvation boundary; for $q_0=0.2$,
the critical initial attempt probability is $a_0=0.36$.
\textbf{(b)} Two 8B Binary RL examples at 4K: a correct Omni radical
answer and a wrong DAPO integer answer. The same DAPO-calibrated cutoffs
at $\alpha=0.065$ reject the radical and accept the integer under Box NLL;
Tail NLL reverses both decisions (\hyperref[case:audited-05]{Case 5}
and~\hyperref[case:audited-10]{Case 10}, Appendix~\ref{app:audited-cases}).}
\label{fig:theory-intuition}
\end{figure}

A correct symbolic derivation can have lower average surprise yet greater
total surprise than a short wrong answer. Mean and tail averages remove
the explicit length multiplier. Figure~\ref{fig:theory-intuition}(b)
illustrates a complementary problem with final-box scoring: it rejects a
correct radical answer while accepting a confident wrong integer.
Answer-span tail scoring reverses both decisions in these observed cases.

For $T_k(z)$, the average of the $k$ largest vector entries, the standard
conditional value-at-risk (CVaR) representation
\citep{rockafellar2002conditional} quantifies sensitivity to single tokens.

\begin{proposition}[Tail aggregation and spike sensitivity]
\label{prop:tail-aggregation}
For $z\in\R^L$ and $k\in\{1,\ldots,L\}$,
\begin{equation}
 T_k(z)=\min_\eta\left\{\eta+\frac1k\sum_{t=1}^L(z_t-\eta)_+\right\}
       =\max_{\substack{0\le w_t\le1/k\\\sum_tw_t=1}}w^\top z.
 \label{eq:tail-primal-dual}
\end{equation}
This is empirical CVaR at tail mass $k/L$. A coordinate increase
$u\ge0$ changes $T_k$ by at most $u/k$.
\end{proposition}
\emph{Proof.} Appendix~\ref{app:tail-proof} gives the proof;
Lemma~\ref{lem:rounded-cvar} bounds rounding and replication errors.

Tail weights cap each token's influence at $1/k$; a maximum permits the
full change $u$. Both statements hold under arbitrary token dependence.
Mean NLL is the natural length-normalized baseline. Tail averaging can
improve separation when discarded background adds noise without a
class-mean signal. Under the fixed-length, separated-support model in
Appendix~\ref{app:sparse-proof}, let $V_C$ and $V_B$ be the summed
critical and background noise variances. For $V_C>0$,
Corollary~\ref{cor:tail-gain-main} gives a standardized-separation gain of
$\sqrt{1+V_B/V_C}$.
At $V_B=0$, common deterministic background makes the mean a positive
affine transform of the tail, so their rankings coincide. Small $V_B/V_C$
instead predicts a small separation gain.

Corollary~\ref{cor:bounded-operating-point}
turns the removal of background concentration terms into a stronger
correct-submission lower bound at a fixed wrong-submission budget;
Corollary~\ref{cor:score-calibration-slack} transfers that bound to the CRC
cutoff when the population threshold has sufficient calibration slack.

This heterogeneous-token view connects to \citet{wang2025beyond8020}, who
identify high-entropy reasoning forks and show that a minority of tokens
can support effective RL updates amid a largely low-entropy background.
Tail-CRC applies this perspective to inference-time submission through
sampled-surprise aggregation, with the mathematics and code analyses in
Section~\ref{sec:audited-results} examining both gains and near-equal rankings.

\subsection{Single-rollout finite-sample risk control}
\label{sec:guarantees}

To calibrate submission, treat each problem family and its fixed-decoder
output as one unit. Internal token dependence and variable trace length do
not prevent exchangeability across families. Let $n$ be the
calibration size, $0<\alpha<1$ the marginal wrong-submission target,
$j=\lfloor(n+1)\alpha\rfloor$, and $S^W_{(1)}\le\cdots\le S^W_{(m)}$
the wrong calibration scores. The bounded-loss CRC rule
\citep{angelopoulos2024crc} specializes to
\begin{equation}
 \widehat\tau_\alpha=
 \begin{cases}
 -\infty,&j=0,\\
 +\infty,&j\ge1,\ m<j,\\
 S^W_{(j)},&j\ge1,\ m\ge j.
 \end{cases}
 \label{eq:crc-cut}
\end{equation}
Strict acceptance rejects cutoff ties. When $j\ge1$, the rule allows at
most $j-1$ wrong calibration submissions, accounting for the finite-sample
correction for the unseen family.

\Needspace{9\baselineskip}
\begin{theorem}[Single-rollout marginal risk control]
\label{thm:crc}
Let calibration and test families be exchangeable. Fix the solver,
decoder, score, and verifier independently of these data. Applying the
same stochastic kernel independently to each family preserves
outcome--score exchangeability, and the cutoff~\eqref{eq:crc-cut} satisfies
\begin{equation}
 \E_{\mathcal D_{\rm cal},\,{\rm test}}
 \ind\{Y_{\rm test}=W,\ S_{\rm test}<\widehat\tau_\alpha\}
 \le\frac{\lfloor(n+1)\alpha\rfloor}{n+1}\le\alpha.
 \label{eq:crc-guarantee}
\end{equation}
This holds at any solver accuracy, with one test trajectory.
\end{theorem}
\emph{Proof.} Appendix~\ref{app:crc-proof} proves decoder-preserved
exchangeability and the rank bound, including ties.

The guarantee averages over calibration and test draws. Useful retention
still depends on the solver supplying correct candidates and the score
ranking them above wrong ones in confidence. For $n=300$ and
$\alpha=0.065$, $j=19$: at most 18 wrong calibration answers pass, and
expected test risk is at most $19/301\approx6.31\%$. A realized test-panel
rate, or its confidence bound, can exceed the target.
Choosing the largest admissible cutoff retains every lower-score candidate
consistent with the corrected calibration count. Better discrimination
therefore creates room for more correct submissions under the same risk
rule. This separates two empirical questions: whether training increases
the correct-candidate supply, and whether the score retains that supply
when submission risk is controlled.

For source and target laws $P,Q$ with outcome--score marginals $P_Z,Q_Z$, define
$F_W^P(t)=P(Y=W,S<t)$.
\begin{equation}
 \E_{\mathcal D_{\rm cal}\sim P^n}F_W^Q(\widehat\tau_\alpha)
 \le\alpha+\sup_t\{F_W^Q(t)-F_W^P(t)\}
 \le\alpha+\operatorname{TV}(P_Z,Q_Z).
 \label{eq:shift}
\end{equation}
\emph{Proof.} Appendix~\ref{app:shift-proof} bounds the additional target
wrong-answer mass below source cutoffs. Target-domain calibration restores
the exchangeable setting without changing the solver.
This motivates testing solver transfer and cutoff transfer separately:
a model can retain its ability to solve target problems while the source
cutoff admits too much of the target's wrong-answer score distribution.

CRC controls marginal wrong submissions. Learn-then-Test (LTT) certifies error
\emph{among submitted answers} with high probability over calibration
\citep{angelopoulos2025ltt}. Even zero errors require 44 accepted examples
for a 10\% conditional-error target with ten fixed cutoffs and $\delta=0.1$.
Appendix~\ref{app:ltt-proof} gives the test and Theorem~\ref{thm:ltt}.

\section{Experimental Setup}
\label{sec:audited-experiments}
\paragraph{Models and controls.}
We evaluate Qwen3.5-4B and Qwen3-8B at total caps of 4K and 8K tokens
\citep{qwen2026qwen35,yang2025qwen3}. \textbf{Binary RL} uses 50
Dr.\ GRPO updates on 400 DAPO families: 16 prompts with eight rollouts each
per update, or 6,400 trajectories per seed. Rank-32 LoRA~\citep{hu2022lora}
updates attention, MLP, and output projections. Training uses 4K external
budget forcing, binary correctness reward, no KL penalty, and loss
denominator $16\times8\times4096$; the final weights serve both caps.
Controls are the pretrained model with \textbf{Native} decoding or
\textbf{Budget forcing}, \textbf{Answer-SFT}, and \textbf{Penalty RL}
($\lambda=1$). Two ablations use \textbf{Mild-Penalty RL}
($\lambda=0.25$) or \textbf{Explicit-Abstain RL}, which adds an abstention
instruction and action to the $\lambda=1$ reward. Binary RL has seeds
41 and 73; penalty arms use seed 41. All decoding uses temperature 1,
top-$p=0.95$, and top-$k=20$. \textbf{Tail-CRC} filters Binary RL with
calibrated Tail NLL.

\paragraph{Families and splits.}
DAPO supplies 400 training, 200 development, 500 calibration, and 500 test families in
disjoint sets~\citep{yu2025dapo}. Independent Omni-MATH evaluation uses 600 verifiable
problems held out from earlier studies, split into 300 calibration and 300 test
families without answer-shape quotas~\citep{gao2025omni}; test references include 37
bare integers and 263 other forms. To preserve blinding, all confirmation trajectories
were generated before calibration labels were opened, and calibration cutoffs were
locked before test labels were revealed. Additional experiments on these benchmarks
examine Native decoding, penalty variants, white-box scoring, and retrospective DAPO
Tail NLL comparisons; full protocols appear in Appendix~\ref{app:protocols}.

\paragraph{Cross-domain code-generation transfer.}
LiveCodeBench~\citep{jain2025livecodebench} tests whether mathematics-trained budget
adaptation transfers to executable Python. We use 1,052 eligible deduplicated problems
dated May 2023--April 6, 2025, partitioned into a 20-problem pilot, 300 calibration
problems, 300 test problems, and a 432-problem held-out reserve, all disjoint. After
the pilot, the main evaluation uses Qwen3-8B base and both frozen Binary RL models at
4K/8K with budget forcing; replaying their 8K first segments also supplies paired
Native 4K controls without new generation. Reference-free validity $v(x,y)$ requires
normal end-of-sequence termination, a complete nonempty Python code block, and valid
Python syntax; correctness then requires passing the official public and private tests
in an isolated sandbox. All trajectories and scores were recorded before calibration
labels were opened, and thresholds were fixed before test labels were revealed. The 4B
pilot results motivate focusing the main evaluation on 8B
(Section~\ref{sec:discussion}, Appendix~\ref{app:lcb-extended}); the full protocol
appears in Appendix~\ref{app:lcb-protocol}.

\paragraph{Estimation.}
Binary RL metrics average the two trained runs \emph{within family}, estimating single-run deployment without answer aggregation.  $R$, $W$, net score
$R-W$, and cost use all test families: 500 on DAPO and 300 each on
Omni-MATH and LiveCodeBench;
conditional error uses accepted submissions. Rates are percentages,
absolute differences are percentage points (pp), and cost savings are
relative to the stated comparator. AUROC uses valid right/wrong submissions
with wrong answers positive; two-seed AUROC averages the seed-specific
values. Pointwise 95\% confidence intervals (CIs) use 20,000 paired family
bootstrap resamples conditional on trained models and fitted thresholds;
white-box AUROC uses up to 2,000 estimable resamples. Full estimation and
calibration diagnostics appear in Appendices~\ref{app:protocols}
and~\ref{app:audited-diagnostics}.
The mathematics score analysis is retrospective; LiveCodeBench prespecified
its answer-span Tail NLL score and auxiliary comparisons before opening test labels.

\section{Results}
\label{sec:audited-results}
\subsection{Q1: Does budget adaptation improve solving?}

Binary RL increases correct submissions over Budget forcing in all eight
settings, by 15.0--29.6 pp for 4B and 4.3--12.8 pp for 8B, with
positive paired lower bounds throughout (Figure~\ref{fig:teaser},
Table~\ref{tab:audited-main}); Appendix Table~\ref{tab:audited-raw-contrasts}
provides all paired intervals and per-family costs. Budget forcing itself recovers completed
answers but also exposes errors: across the four 8B domain/cap settings,
it produces 6.3--26.5 times as many wrong submissions as Native. Binary RL's API generation
cost per family falls by 3.5--12.6\% against the same-budget comparator.
At 4K, its correct-submission gains over Answer-SFT are 12.30/8.33 pp
on DAPO/Omni for 4B and 5.80/7.33 pp for 8B. The independent Omni-MATH
confirmation validates both these solving gains and the score ranking
analyzed next.
Under Native decoding, 4B also gains 1.60--7.20 pp in correct submissions
across all four settings; 8B changes span $-0.40$ to $+0.60$ pp
(Appendix Table~\ref{tab:native-transfer}).

\begin{table}[t]
\centering\small
\begingroup%
\renewcommand{\arraystretch}{1.20}%
\setlength{\tabcolsep}{4.1pt}%
\begin{tabular*}{\linewidth}{@{\extracolsep{\fill}}lll*{8}{C{2.0em}}*{2}{>{\columncolor{gray!8}}C{2.0em}}r@{}}
\toprule
\multirow{2}{*}{Domain} & \multirow{2}{*}{Model} & \multirow{2}{*}{Cap}
& \multicolumn{2}{c}{Native}
& \multicolumn{2}{c}{\shortstack{Budget\\forcing}}
& \multicolumn{2}{c}{\shortstack{Answer-\\SFT}}
& \multicolumn{2}{c}{Binary RL}
& \multicolumn{2}{>{\columncolor{gray!8}}c}{\textbf{Tail-CRC}}
& \multirow{2}{*}{\shortstack{Binary RL\\cost saving}} \\
\cmidrule(lr){4-5}\cmidrule(lr){6-7}\cmidrule(lr){8-9}\cmidrule(lr){10-11}\cmidrule(lr){12-13}
& & & $R$ & $W$ & $R$ & $W$ & $R$ & $W$ & $R$ & $W$ & $R$ & $W$ & \\
\midrule
DAPO & 4B & 4K & 0.8 & 0.0 & 21.2 & 2.4 & 35.6 & 1.6 & 47.9 & 5.0 & 47.9 & 5.0 & 6.5\% \\
DAPO & 4B & 8K & 7.0 & 0.4 & 28.0 & 1.0 & 49.2 & 1.8 & 57.6 & 6.3 & 57.6 & 6.3 & 12.6\% \\
\addlinespace[3pt]
DAPO & 8B & 4K & 7.6 & 0.4 & 28.6 & 10.6 & 35.6 & 13.2 & 41.4 & 18.9 & 33.9 & 7.1 & 4.7\% \\
DAPO & 8B & 8K & 35.4 & 0.8 & 53.2 & 14.6 & 56.6 & 18.4 & 57.5 & 26.4 & 39.1 & 5.9 & 8.2\% \\
\addlinespace[3pt]
Omni & 4B & 4K & 2.3 & 0.7 & 10.3 & 1.7 & 17.0 & 4.3 & 25.3 & 9.8 & 17.8 & 3.7 & 5.8\% \\
Omni & 4B & 8K & 8.7 & 0.3 & 14.3 & 2.7 & 21.7 & 6.3 & 30.5 & 12.8 & 24.3 & 5.8 & 9.9\% \\
\addlinespace[3pt]
Omni & 8B & 4K & 10.0 & 1.7 & 22.7 & 18.7 & 22.0 & 23.3 & 29.3 & 30.3 & 16.7 & 5.8 & 3.5\% \\
Omni & 8B & 8K & 17.7 & 5.0 & 31.0 & 31.3 & 34.7 & 35.0 & 38.8 & 41.8 & 18.0 & 6.5 & 8.2\% \\
\bottomrule
\end{tabular*}%
\endgroup%

\caption{\textbf{Correct/wrong submissions under a hard cap.}
$R$/$W$ are percentages of all 500 DAPO or 300 Omni test families.
Tail-CRC filters Binary RL outputs
at $\alpha=0.065$ with same-domain calibration. Answer-SFT training sets
are documented in Appendix~\ref{app:protocols}. Cost saving compares
Binary RL with Budget forcing at the same cap. For 4B DAPO, the calibration
error counts meet CRC's corrected allowance, so $\widehat\tau=+\infty$
retains all Binary RL answers.}
\label{tab:audited-main}
\end{table}

\begin{figure}[tb]
\centering
\includegraphics[width=\linewidth]{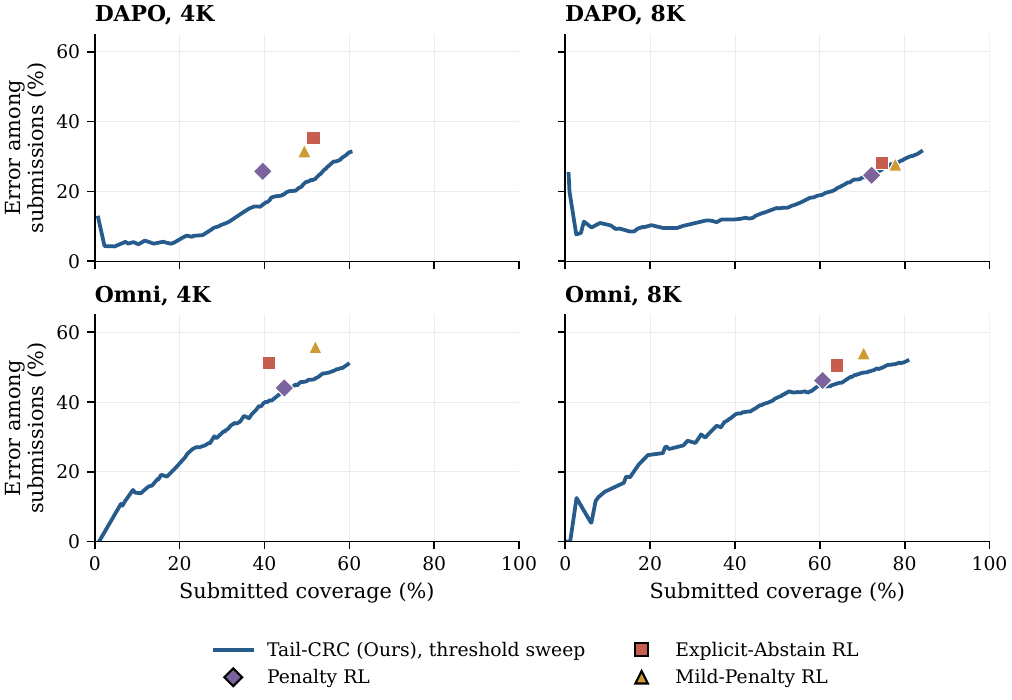}
\caption{\textbf{Selection improves the observed risk--coverage frontier.}
The Binary RL threshold sweep dominates seven of eight raw
Explicit-Abstain RL/Mild-Penalty RL points across two domains and caps.
The exception is Mild-Penalty RL on DAPO 8K: 77.8\% coverage and
27.76\% error, versus 28.11\% error for Binary RL at at least that coverage.
Calibration orders thresholds; the test sweep describes the frontier
without selecting a deployment policy.}
\label{fig:frontier}
\end{figure}

Training-time penalties trade away correct answers without directly
setting deployment risk. On 8B DAPO 4K, Penalty RL lowers $W$ from
Binary RL's 18.9\% to 10.2\%, but also lowers $R$ from 41.4\% to
29.4\%. Truncation rises from 36.1\% to 58.2\%, and sampled output
length from 3,423 to 3,628 tokens per family. This pattern is consistent
with the incentive isolated in Proposition~\ref{prop:penalty-signal}.
The training curves in Appendix Figure~\ref{fig:training} show that Penalty RL keeps
more reward-varying groups active, yet its first/last-ten-update truncation
means fall only from 64.9\% to 62.7\%; Binary RL seed 41 reaches 47.3\%.
Failure-only comparisons therefore need not supply productive solving
evidence.
Explicit-Abstain RL uses its added action on 0.6\%/1.0\% of DAPO
families at 4K/8K, versus 10.0\%/14.3\% on Omni. Of its 43 Omni 8K
abstentions, Penalty RL had answered 11 correctly; among its remaining
submissions, 50.5\% are wrong. The added action therefore trades away some correct answers while
leaving substantial accepted error.

\begin{table}[t]
\centering\small
\begingroup%
\renewcommand{\arraystretch}{1.20}%
\setlength{\tabcolsep}{4pt}%
\begin{tabular*}{\linewidth}{@{\extracolsep{\fill}}llrrrrrr@{}}
\toprule
\multirow{2}{*}{Domain} & \multirow{2}{*}{Arm}
& \multicolumn{4}{c}{Unfiltered (\%)}
& \multicolumn{2}{c}{Calibrated net (\%)} \\
\cmidrule(lr){3-6}\cmidrule(l){7-8}
& & $R$ & $W$ & Trunc. & Abstain & $\alpha=0.065$ & $\alpha=0.10$ \\
\midrule
DAPO & Binary RL & 41.4 & 18.9 & 36.1 & 0.0 & 26.8 & 28.2 \\
DAPO & Penalty RL & 29.4 & 10.2 & 58.2 & 0.0 & 19.6 & 19.2 \\
DAPO & Explicit-Abstain RL & 33.4 & 18.2 & 44.8 & 0.6 & 21.2 & 18.8 \\
DAPO & Mild-Penalty RL & 33.8 & 15.6 & 46.2 & 0.0 & 21.8 & 19.2 \\
\addlinespace[3pt]
Omni & Binary RL & 29.3 & 30.3 & 36.2 & 0.0 & 10.8 & 11.3 \\
Omni & Penalty RL & 25.0 & 19.7 & 52.7 & 0.0 & 8.3 & 11.0 \\
Omni & Explicit-Abstain RL & 20.0 & 21.0 & 44.3 & 10.0 & 8.3 & 7.7 \\
Omni & Mild-Penalty RL & 23.0 & 29.0 & 43.3 & 0.0 & 7.0 & 7.0 \\
\bottomrule
\end{tabular*}%
\endgroup%

\caption{\textbf{8B reward ablations at 4K.}
Raw $R$, $W$, truncation, and explicit-abstention rates use all 500 DAPO
or 300 Omni test families. Final columns give net score after separate
same-domain Tail NLL/CRC fits at the stated marginal-error targets.
Binary RL uses both training seeds; the other arms use seed 41.}
\label{tab:audited-penalty}
\end{table}

Giving every competitor the same calibration procedure preserves the
main ordering. At 4K, Tail-CRC has higher net-score point estimates than
all three separately calibrated penalty baselines in all twelve comparisons:
two domains, three arms, and targets $\alpha=0.065/0.10$.
Against Penalty RL, the gains at these two targets are 7.20/9.00 pp on
DAPO and 2.50/0.33 pp on Omni. Against Explicit-Abstain RL and
Mild-Penalty RL, the four DAPO gains are 5.00--9.40 pp, all with positive
paired lower bounds; the four Omni gains are 2.50--4.33 pp, with intervals
crossing zero. Section~\ref{sec:discussion} discusses the 8K Omni boundary.

\begin{samepage}
Complementing these operating points, Binary RL's observed
selection frontier dominates seven of eight raw ablation points
(Figure~\ref{fig:frontier}). Solving and adjustable selection are useful
separate controls; a fixed reward penalty does not replace either.
\end{samepage}

\subsection{Q2: Which single-rollout score supports selection?}

In an earlier preregistered DAPO study using Box NLL, the 8B model did not jointly meet its prespecified
risk and gain criteria (Appendix~\ref{app:study-history}), motivating
our shift to answer-span tail scoring. Box NLL loses information when
correctness depends on a long or symbolic
answer. The observed cases in Figure~\ref{fig:theory-intuition}(b)
illustrate this mismatch; the independent Omni test quantifies it.
For 8B Binary RL, Tail NLL versus Box NLL AUROCs are 0.756 versus
0.628 at 4K and 0.724 versus 0.622 at 8K. The paired improvements are
\textbf{0.128} (95\% CI $[0.043,0.216]$) and \textbf{0.102}
($[0.025,0.180]$). Under source-calibrated cuts at $\alpha=0.065$, accepted correct
answers whose references are not written as a single integer increase
from 42 to 90 at 4K and 44 to 94
at 8K, across 526 seed--family records (263 families, two seeds).

Sum NLL has a separate length problem: its AUROC is 0.111 for the 4B
Budget forcing baseline on DAPO 4K and 0.289 for Answer-SFT.
Proposition~\ref{prop:length-discrimination} isolates this multiplicative
length confounding. Across 36 model--domain--budget cells (nine model
arms $\times$ two domains $\times$ two caps; Appendix Table~\ref{tab:audited-uncertainty}),
Tail NLL and Mean NLL have AUROC
ranges 0.630--0.862 and 0.626--0.854, respectively, with an 18--18 split.
A retrospective analysis of these mathematics trajectories
locates when tail averaging adds value (Appendix Table~\ref{tab:math-stage1}).
The equal-weight Tail NLL minus Mean NLL AUROC difference is
$-0.0003$ ($[-0.0061,0.0056]$) on the answer span, but rises to
$+0.0055$ ($[0.0008,0.0105]$) on the full trajectory, including thinking.
The span interaction is $+0.0058$ ($[0.0003,0.0115]$).
On full trajectories, Tail NLL exceeds Max NLL by
$0.0649$ ($[0.0342,0.0952]$), with gains in 31 of 36
settings, consistent with
Proposition~\ref{prop:tail-aggregation}.

The gain over Mean NLL is strongest in long 8B DAPO 8K derivations:
all five model arms favor answer-span Tail NLL, averaging
$+0.0167$ AUROC ($[0.0108,0.0228]$).

\begin{figure}[t]
\centering
\includegraphics[width=\linewidth]{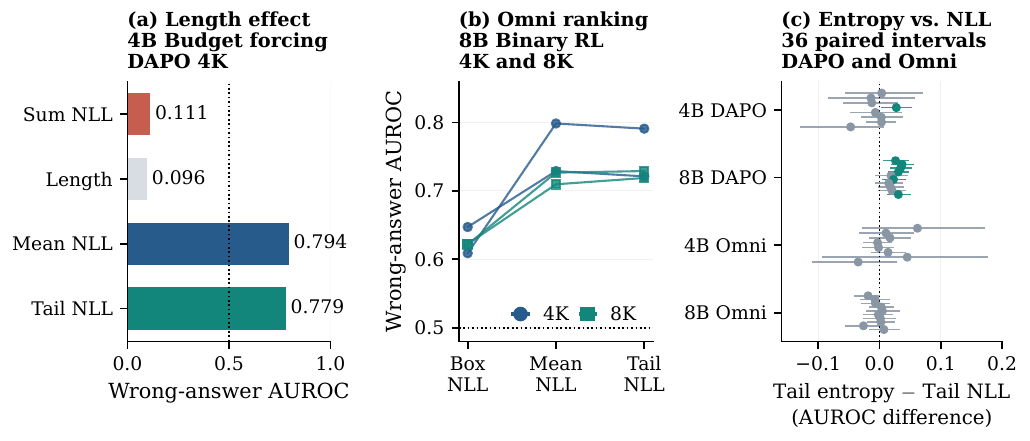}
\caption{\textbf{Single-trajectory score behavior.}
(a) Sum NLL reverses discrimination on 4B Budget forcing, DAPO 4K.
(b) Mean NLL and Tail NLL improve on Box NLL for both 8B Binary RL
seeds on Omni. (c) Tail entropy minus Tail NLL AUROC: seven of 36
paired intervals are positive; the other 29, including all 18 Omni
intervals, cross zero.}
\label{fig:score}
\end{figure}

\begin{samepage}
This ranking improvement translates into retained solutions. For the tail-fraction comparison, Tail-10 denotes the default top-10\% Tail NLL score and Tail-5 its top-5\% variant.
At $\alpha=0.065$, Binary RL's $R/W$ changes from Mean NLL's
$37.7/6.2\%$ to Tail-10's $39.1/5.9\%$, increasing net score by
1.7 pp ($[0.7,2.8]$). Tail-5 retains $41.5/6.4\%$,
for a 3.6 pp net gain ($[2.0,5.2]$).
Together with the Box and Sum failures in Figure~\ref{fig:score},
these results connect tail averaging to stronger separation and
correct-answer retention.
\end{samepage}

\paragraph{What white-box access adds.}
Full-vocabulary access provides an optional score upgrade. Replaying
28,800 trajectories and 15,461,536 valid answer tokens finds seven
significant Tail entropy gains: six 8B DAPO cells and one 4B DAPO cell.
The other intervals cross zero, including all 18 Omni cells.
At 8B DAPO 8K, replacing Tail NLL
with Tail entropy at $\alpha=0.065$ raises Binary RL's $R$ from 39.1\%
to 43.7\%, changes $W$ from 5.9\% to 6.2\%, and raises net score from
33.2\% to 37.5\%.
Within the white-box scores themselves, Tail entropy exceeds Mean entropy
in all five 8B DAPO 8K arms, averaging $+0.0393$ AUROC;
the pooled Binary RL net gain is 4.2 pp ($[2.5,6.0]$;
Appendix Table~\ref{tab:math-stage1}).

Writing $p_t$ for the model's next-token distribution at step $t$ and
$Y_t$ for the sampled token, sampled surprise decomposes as
$-\log p_t(Y_t)=H(p_t)+b_t+\xi_t$, where $H(p_t)$ is full-vocabulary
Shannon entropy, $b_t$ is predictable drift from the truncated decoder,
and $\xi_t$ is conditionally centered sampling noise. Under raw-model
sampling, the conditional variance of $\xi_t$ is varentropy.
Tail selection can retain extreme drift or noise as well as entropy,
so tokenwise centering does not equate the tail scores.
Proposition~\ref{prop:decoder-decomposition} gives the decomposition and
pathwise spike bound. Lemma~\ref{lem:predictable-isometry} explains why
random-length normalization and EOS removal need care, while
Lemma~\ref{lem:auc-margin} connects score perturbations to AUROC through
pairwise margins: a small AUROC difference alone does not imply low
sampling variance.

\subsection{Q3: How do calibration and cost affect deployment?}

Target calibration substantially reduces exposed errors. On 8B Omni,
Tail-CRC at $\alpha=0.065$ lowers $W$ from 30.33\% to 5.83\% at 4K
and from 41.83\% to 6.50\% at 8K, while net score rises from $-1.00\%$
to 10.83\% and from $-3.00\%$ to 11.50\%. Source-calibrated cuts
transfer less effectively: they accept every valid 4B Binary RL Omni
output, giving $W=9.83\%$ at 4K and 12.83\% at 8K. Under transferred
source cuts on 8B Omni, Tail NLL improves net score over Box NLL at 8K
(+5.17 pp, 95\% CI $[0.33,10.17]$), but its 4K interval crosses zero
(+3.50 pp, $[-1.00,8.33]$; Appendix~\ref{app:study-history}).

\begin{figure}[t]
\centering
\includegraphics[width=\linewidth]{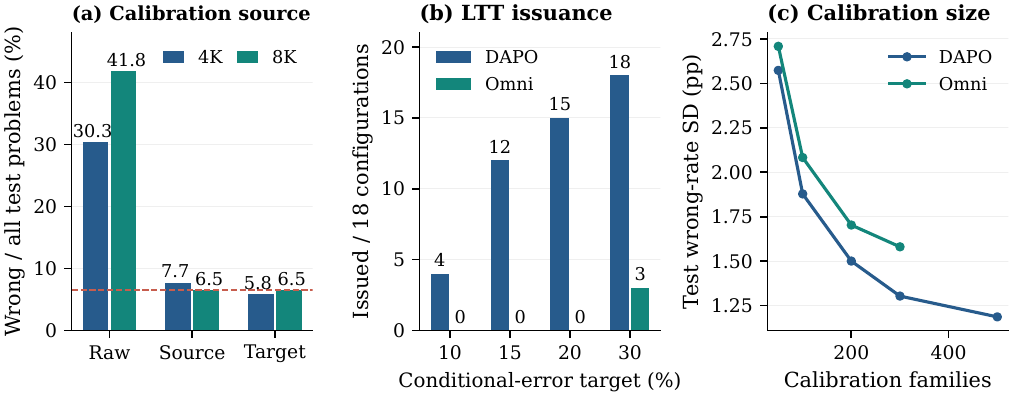}
\caption{\textbf{Target calibration and conditional certification.}
(a) 8B Binary RL Omni wrong-submission rates, raw and with source- or
target-calibrated Tail-CRC at $\alpha=0.065$.
(b) Tail NLL LTT issuance out of 18 configurations per domain/target
(nine models, two caps); DAPO is retrospective and Omni uses independently
registered grids. (c) Test-$W$ SD over 2,000 repartitions at each
calibration size, for 8B 4K and $\alpha=0.065$. Fixed pools have 1,000
DAPO and 600 Omni families; those outside calibration form the test set.}
\label{fig:calibration}
\end{figure}

Among 36 target-calibrated settings (nine individual models, two caps,
targets $\alpha=0.065/0.10$), 28 realized test wrong-submission rates meet
target. Across 2,000 finite-pool splits at 8B 4K and
$\alpha=0.065$, mean $W$ is 6.41\% on DAPO and 6.31\% on Omni
(Figure~\ref{fig:calibration}). Conditional certification needs enough
accurate accepted answers: Tail NLL issues 49/72 certificates in the
retrospective DAPO analysis, while Tail NLL and Mean NLL each issue
3/72 on independently registered Omni grids, all at a 30\% conditional-error
target. Appendix Tables~\ref{tab:audited-ltt} and~\ref{tab:ncal-all} give full counts
and calibration-size sensitivity.

The cost trade-off depends on model size. Against Native 8K of the same
size, 4K Tail-CRC raises 4B net score by 36.30 pp ($[31.70,40.80]$)
on DAPO and 5.83 pp ($[2.33,9.50]$) on Omni, while API cost falls
47.4\%/45.1\%. The 8B cost savings are 43.0\%/44.3\%, but net-score
differences are $-7.80$ pp ($[-11.80,-3.80]$) and $-1.83$ pp
($[-5.83,2.17]$), respectively.

\begin{samepage}
The models differ in how often they finish within 8K:
4B Native 8K still truncates 92.6\% of DAPO and 91.0\% of Omni trajectories,
giving budget adaptation room to improve both net score and cost.
In contrast, 8B Native 8K
already solves 35.4\% of DAPO problems with 97.8\% submission precision;
compressing its budget to 4K exchanges some net score for lower wall time
and 43.0\% lower API cost.
\end{samepage}

\begin{table}[t]
\centering\small
\begingroup%
\renewcommand{\arraystretch}{1.16}%
\setlength{\tabcolsep}{6pt}%
\begin{tabular}{@{}llrr@{}}
\toprule
\multirow{2}{*}{Model} & \multirow{2}{*}{Generation configuration} & \multicolumn{1}{c}{Generation time} & \multicolumn{1}{c}{Relative to} \\
& & \multicolumn{1}{c}{(s/family)} & \multicolumn{1}{c}{Native 8K} \\
\midrule
4B & Native, 8K & 20.88 & -- \\
4B & Binary RL, 4K & 10.20 & 48.9\% of time (4K cap) \\
\addlinespace[3pt]
8B & Native, 8K & 13.60 & -- \\
8B & Binary RL, 4K & 6.11 & 44.9\% of time (4K cap) \\
8B & Binary RL, 8K & 11.90 & 12.56\% less (8K cap) \\
\bottomrule
\end{tabular}%
\endgroup%

\caption{\textbf{Warm-engine generation time on 50 held-out DAPO problems.}
Binary RL uses seed 41. Times are amortized seconds per family on
vLLM/A100, with Native 8K using the original model weights as the
baseline for relative times. The slower 4B Native 8K run uses a different,
hybrid-attention architecture~\citep{qwen2026qwen35} and produces longer
traces here (7,980.50 versus 7,037.34 sampled tokens per family).
Its 8K truncation rate on the full DAPO test set is also higher (92.6\% versus 62.8\%). Relative times use the same model size; 4K rows compress the
budget, while the 8K row compares matched caps. Offline entropy
extraction is excluded.}
\label{tab:audited-latency}
\end{table}

On vLLM~\citep{kwon2023vllm}, 8B Binary RL at a matched 8K cap uses
\textbf{12.56\% less generation time} than Native 8K:
11.90 versus 13.60 seconds per family (Table~\ref{tab:audited-latency}).
Compressing the budget to 4K lowers time to 48.9\% of Native 8K for
4B and 44.9\% for 8B. These are budget-compression wall-clock trade-offs,
read together with the net-score changes above. A precomputed score
requires about 26 ns per family for thresholding.

\begin{samepage}
With Tail entropy at $\alpha=0.065$, 8B Binary RL on DAPO 8K achieves
$R/W/\mathrm{Net}=43.7/6.2/37.5\%$, versus $35.4/0.8/34.6\%$ for
Native 8K. The post hoc paired differences are
\textbf{$\Delta R=+8.3$ pp} ($[4.6,12.0]$),
$\Delta W=+5.4$ pp ($[3.8,7.0]$), and
\textbf{$\Delta\mathrm{Net}=+2.9$ pp} ($[-1.3,7.0]$), all with 95\% CIs.
It submits more correct answers and more errors; the net-score interval
crosses zero. Note that the 12.56\% generation-time reduction above excludes
the offline pass required to extract full-vocabulary entropy.
\end{samepage}

\subsection{Q4: Do budget adaptation and Tail-CRC transfer to algorithmic code?}
\label{sec:lcb-results}

\paragraph{Budget bottleneck and zero-shot RL transfer.}
Code generation exposes the same budget bottleneck, and mathematics-trained
Binary RL transfers across it. Native 4K truncates 72.33\% of base
trajectories and 70.33\% of Binary RL trajectories; budget forcing restores
many completed programs but exposes 8.33--11.83\% wrong submissions
(Table~\ref{tab:lcb-main}). With no code training, Binary RL improves
raw $R$ over the same-budget base by 7.17 pp ($[3.00,11.33]$) at 4K
and 4.50 pp ($[1.33,7.83]$) at 8K, while reducing truncation by
5.83/5.50 pp and lowering per-family generation cost.
Applying the same answer-span Tail-CRC procedure to both models preserves
net-score gains of 6.50 pp ($[1.00,11.84]$) and 5.83 pp
($[0.50,11.33]$). Both gains retain positive paired lower bounds after prespecified errata exclusions
(Appendix Table~\ref{tab:lcb-sensitivity}).

\begin{table}[tb]
\centering\small
\begingroup
\setlength{\tabcolsep}{6pt}
\renewcommand{\arraystretch}{1.16}
\begin{tabular}{@{}llrrrr@{}}
\toprule
Solver & Submission rule & $R$ (\%) & $W$ (\%) & Net (\%) & \shortstack[r]{API token cost\\(USD/problem)} \\
\midrule
\multicolumn{6}{@{}l}{\textbf{Native 4K (paired first-segment replay)}} \\
Base & Raw & 27.67 & 0.00 & 27.67 & 0.00218 \\
Binary RL & Raw & 29.67 & 0.00 & 29.67 & 0.00216 \\
\addlinespace[5pt]
\multicolumn{6}{@{}l}{\textbf{Budget forcing, 4K}} \\
Base & Raw & 42.00 & 9.33 & 32.67 & 0.00234 \\
Base & Tail-CRC (answer) & 41.00 & 6.33 & 34.67 & 0.00234 \\
\addlinespace[3pt]
Binary RL & Raw & 49.17 & 8.33 & 40.83 & 0.00228 \\
Binary RL & Tail-CRC (answer) & 48.50 & 7.33 & 41.17 & 0.00228 \\
\addlinespace[5pt]
\multicolumn{6}{@{}l}{\textbf{Budget forcing, 8K}} \\
Base & Raw & 58.33 & 11.33 & 47.00 & 0.00373 \\
Base & Tail-CRC (answer) & 50.33 & 6.33 & 44.00 & 0.00373 \\
Base & Tail-CRC (full trajectory) & 55.00 & 6.00 & 49.00 & 0.00373 \\
\addlinespace[3pt]
Binary RL & Raw & 62.83 & 11.83 & 51.00 & 0.00356 \\
Binary RL & Tail-CRC (answer) & 54.67 & 4.83 & 49.83 & 0.00356 \\
Binary RL & Tail-CRC (full trajectory) & 57.67 & 5.83 & 51.83 & 0.00356 \\
\bottomrule
\end{tabular}
\endgroup

\caption{\textbf{Mathematics-trained solving transfers to LiveCodeBench.}
Qwen3-8B on 300 held-out test problems. Binary RL metrics average two independently
trained runs within each problem; deployment uses one trajectory. $R$, $W$, and Net
are percentages of all test problems. Tail-CRC uses separate cutoffs fitted on 300
calibration problems at $\alpha=0.065$; the answer span is primary and the full
trajectory is a prespecified auxiliary comparison. Native 4K replays the corresponding
8K first segment without new requests; its API token cost is computed for an equivalent
standalone deployment of that segment. Filtering reuses the rollout, preserving generation cost.}
\label{tab:lcb-main}
\end{table}

\paragraph{Single-rollout risk control on code.}
At 8K, answer-span Tail-CRC lowers Binary RL's wrong submissions from
11.83\% to 4.83\%, retaining 54.67\% correct submissions with 91.88\%
submission precision. At 4K, the original 300-problem test split has a 7.33\% wrong-submission rate; the two seeds have only 21 and 26 calibration errors around the
rank-$19$ cutoff, making the operating point sensitive to the split.
Across 2,000 common 300/300 repartitions of the 600-problem pool,
mean $W$ is 6.26\% at 4K and 6.34\% at 8K, with respective standard
deviations of 1.52 and 1.67 pp. These finite-pool averages illustrate
calibration variation around the marginal target in
Theorem~\ref{thm:crc}; the original fixed-split outcomes remain distinct.
On the 41-problem 2025 subset, the same locked 8K rule reduces $W$ from
19.51\% to 6.10\%, raising net score from 17.07\% to 24.39\%
without year-specific calibration.

\paragraph{Score concentration and span choice.}
The code scores also clarify when Mean NLL already captures the tail's information.
Answer- and code-span Tail NLL AUROCs range from 0.827 to 0.871
(Appendix Table~\ref{tab:lcb-score-ranking}), while all twelve Tail NLL minus Mean NLL intervals
across three spans, two caps, and base/Binary RL include zero (Appendix Table~\ref{tab:lcb-score-anatomy}).
For Binary RL, the largest 10\% of code-token surprises contribute 96.09\%/96.78\% of total
NLL at 4K/8K. The corresponding Spearman correlations between Tail NLL and Mean NLL round to 0.999 at both caps.

For a span of $L$ tokens with positive total NLL, let $\overline{\ell}=L^{-1}\sum_{t=1}^L\ell_t$ be its
mean NLL, $T_k(\ell)$ the average of its $k=\lceil0.1L\rceil$ largest token NLLs, and
$M=\sum_{t=1}^k\ell_{(t)}/\sum_{t=1}^L\ell_t$ the fraction of total NLL in those $k$ tokens.
Then $\overline{\ell}=(k/L)\,T_k(\ell)/M$; when $M$ and $k/L$ vary little, the two scores convey
nearly the same ranking information, consistent with Corollary~\ref{cor:tail-gain-main}'s common-background limit.

\begin{samepage}
Averaging remains valuable against isolated spikes: full-trajectory
Tail NLL exceeds Max NLL by 0.066 AUROC ($[0.021,0.112]$) at 4K and
0.056 ($[0.019,0.093]$) at 8K. Including thinking yields 51.83\% net
at 8K, compared with 49.83\% for answer-span scoring and 51.00\% for
raw submission.
\end{samepage}

\FloatBarrier
\section{Related Work}
\label{sec:related}

\paragraph{Reasoning under a compute budget.}
CoT prompting and self-consistency use intermediate computation and
multiple reasoning paths~\citep{wei2022chain,wang2023selfconsistency};
test-time scaling studies how to allocate that computation
\citep{snell2025scaling}. STaR trains on self-generated rationales that
yield correct answers~\citep{zelikman2022star}. Rationale distillation uses
teacher-generated explanations as supervision~\citep{hsieh2023distilling},
and RL can build reasoning behavior directly~\citep{guo2025deepseekr1}.
For explicit budgets, s1 introduces inference-time forcing, L1 trains for
chosen lengths, and ThinkPrune trains under a token limit
\citep{muennighoff2025s1,aggarwal2025l1,hou2026thinkprune}.
Tail-CRC couples budget adaptation to a separate decision about which
completed answers a deployment should expose.

\paragraph{Learning abstention.}
The reject option and selective classification formalize prediction with
withholding~\citep{chow1970reject,geifman2017selective}. Refusal-Aware
Instruction Tuning (R-Tuning) teaches uncertainty responses
\citep{zhang2024rtuning}; reinforced hesitation distinguishes correct,
incorrect, and abstaining outputs through rewards~\citep{mohamadi2025hesitation}.
We analyze how these rewards interact with hard-budget
non-submission under GRPO and Dr.\ GRPO
\citep{shao2024deepseekmath,liu2025drgrpo,zhou2026grpo}.
Matched calibration tests which training objective supplies better
answers at the same risk target.

\begin{table}[t]
\centering\small
\setlength{\tabcolsep}{5pt}
\renewcommand{\arraystretch}{1.2}
\begin{tabular}{@{}L{.25\linewidth}L{.41\linewidth}L{.29\linewidth}@{}}
\toprule
Approach & Decision and risk target & Deployment information\\
\midrule
Risk-controlling prediction sets~\citep{bates2021distribution}
 & Set size chosen to meet a specified population-loss target
 & Predictor scores for candidate outputs\\
\addlinespace[4pt]
Conformal language modeling~\citep{quach2024clm}
 & Candidate set containing at least one acceptable answer
 & Additional sampled candidates as needed\\
\addlinespace[4pt]
Conformal factuality~\citep{mohri2024factuality}
 & Claim-filtered response with marginal factuality control
 & Parsed claims and claim-level confidence scores\\
\addlinespace[4pt]
Enhanced conformal validity~\citep{cherian2024validity}
 & Filtered claims with level-adaptive validity over specified features
 & Claim scores and prompt/response features\\
\addlinespace[4pt]
Conformal CoT aggregation~\citep{gu2026pause}
 & Aggregated answer with controlled marginal wrong-submission risk
 & Multiple scored reasoning paths\\
\addlinespace[4pt]
\TailCRC\ (Ours)
 & One hard-budget answer with controlled marginal wrong-submission risk
 & One trajectory and its returned token log probabilities\\
\bottomrule
\end{tabular}
\caption{\textbf{Calibrated decisions and the information needed to make them.}
Guarantees concern different objects, even when they use related
calibration tools. Deployment requirements describe the cited constructions.}
\label{tab:related}
\end{table}

\paragraph{Conformal control of generated outputs.}
Conformal prediction calibrates set coverage
\citep{vovk2005algorithmic,lei2018distribution}; risk-controlling prediction
sets and CRC extend calibration to loss targets
\citep{bates2021distribution,angelopoulos2024crc}, while LTT tests candidate
policies for risk certification~\citep{angelopoulos2025ltt}.
Table~\ref{tab:related} compares the decisions these methods calibrate.

Language-generation methods control candidate-set success, retained claims,
or an answer aggregated from several paths.
Tail-CRC controls the wrong-submission event from a single budget-adapted
trajectory, using its answer-span token probabilities.

\paragraph{Uncertainty from one response or several.}
Confidence can come from answer probabilities, self-evaluation, or verbal
uncertainty~\citep{jiang2021know,kadavath2022language,lin2022teaching,ren2023selfevaluation}.
Semantic entropy compares meaning across sampled answers
\citep{farquhar2024semantic}; Deep Think with Confidence (DeepConf) filters
parallel reasoning traces by token confidence~\citep{fu2026deepconf}.
Our single-rollout comparison separates representation and length effects
from the choice of token aggregation. CVaR supplies a precise tail-score
interpretation~\citep{rockafellar2002conditional}; the entropy extension
relates sampled surprise to decoder drift and token varentropy
\citep{kontoyiannis2014optimal}.
\citet{wang2025beyond8020} show that a high-entropy minority of reasoning
tokens can carry effective RL updates. Tail-CRC studies the inference-time
counterpart: whether concentrating sampled surprise improves submission
decisions, with Corollary~\ref{cor:tail-gain-main} and the mathematics/code
comparisons showing when tail averaging helps and when rankings are nearly identical.
\FloatBarrier

\section{Discussion and Limitations}
\label{sec:discussion}
\paragraph{Three deployment controls.}
Training supplies correct candidates, scoring separates them from errors,
and calibration sets the marginal wrong-submission target. Binary RL
with Tail-CRC is useful under tight budgets and API log-probability
access. Tail entropy improves the demonstrated 8B DAPO operating point
when full-vocabulary access is available. With costly errors and an 8K
budget, conservative Native decoding remains competitive. The deployment
choice therefore depends jointly on $R$, $W$, and cost.

\paragraph{The precision boundary.}
Calibration cannot create correct answers or recover separation absent
from the score. On 8B Omni at $\alpha=0.065$, Tail-CRC's 4K net-score
gains over identically calibrated Budget forcing and Answer-SFT are
0.17 and 0.50 pp; at 8K it retains 4.0 pp fewer correct answers than
calibrated Penalty RL ($[-7.67,-0.33]$ pp).
Better raw solving does not guarantee the best small accepted subset.

Tail averaging has distinct operating regimes. Across full mathematics and code
trajectories it improves on Max NLL, which is exposed to isolated spikes. Relative to
Mean NLL, gains emerge in the long 8B DAPO 8K derivations and the aggregate
full-trajectory mathematics analysis; code spans instead concentrate 96--97\% of NLL
mass in their top 10\%, leaving Mean NLL and Tail NLL with nearly identical empirical
rankings. This is consistent with the common-background limit of
Corollary~\ref{cor:tail-gain-main}, although mass concentration alone does not
establish its deterministic-background assumption. The mathematical background itself
carries correctness information and correlates with the tail
(Appendix~\ref{app:math-stage1}), so the sparse-signal model explains a possible
benefit without identifying the full data-generating mechanism. The retrospective
Tail-5 comparison supports a narrower tail fraction for these long derivations.

Explanatory language can enter the upper tail, and changing verbosity can also change
the answer, so added text need not be label-neutral background. The pilot did not
attain its planned length increases, leaving unresolved how a label-preserving
expansion would affect score ranking (Appendix~\ref{app:math-stage1}). White-box gains
likewise leave drift and sampling noise unresolved.

\paragraph{Experimental scope.}
The study covers two Qwen model generations, two verifiable mathematics domains, and a
blinded algorithmic code-generation study on LiveCodeBench. The benchmark problems
span May 2023--April 6, 2025; pretraining exposure is unknown. Year strata and
official-errata exclusions fixed in advance are reported in Appendix
Table~\ref{tab:lcb-sensitivity}. The separate 4B pilot on 20 problems truncates
85\%/70\% at 4K/8K and 100\% under Native 4K, with no wrong valid programs to support
ranking estimation. Its 8K generation cost is \$0.00905 per problem, 2.23 times the 8B
base cost, motivating the main 8B evaluation (Appendix~\ref{app:lcb-extended}).

Binary RL has two training seeds; penalty arms have one, and
Explicit-Abstain RL changes instruction and parser together. Equal zero
rewards for refusal and truncation leave their relative value unspecified;
rewarding prompt refusal would be a different intervention.
The solved dynamics isolate a mechanism rather than identify the actual
AdamW trajectory. A longer-budget Native-GRPO training control remains
untested. Appendix Tables~\ref{tab:pilot-local-extbf}--\ref{tab:audited-voting-pilot} give
exploratory training, model-transfer, and two-draw comparisons.

\paragraph{Calibration and systems scope.}
Guarantees require the stated calibration/deployment relationship and correct labels.
Three hundred calibration examples with reference answers cost an estimated \$0.73 for
8B or \$1.31 for 4B at single-seed 4K generation rates; trustworthy labels in
open-ended domains may be harder to obtain. Supplementary mathematics analyses reuse
the existing benchmark sets; the code study tests transfer with thresholds fixed
before test labels were revealed. At a matched 4K cap, extra prefill can make budget
forcing slightly more expensive than Native; rejecting an answer does not recover
generated tokens. Full-vocabulary extraction remains untimed, and generation
equivalence across engines remains unverified. Future work can improve solving,
ranking, and calibration under shift.

\section*{Acknowledgement}
The authors thank the Tinker Research Grant program for supporting this research.

\section*{AI Use Statement}
This work involved extensive discussions with GPT 6 Astra and Gemini 4 Neon.
GPT 6 Astra implemented the experiments.
GPT 6 Astra also contributed to manuscript polishing and appendix organization.
The authors used GPT 6 Astra to help organize and verify the paper's theoretical results.
The authors take full responsibility for the paper's content, results, and
conclusions.

\FloatBarrier
\bibliographystyle{plainnat}
\bibliography{references}
\clearpage
\appendix
\numberwithin{equation}{section}
\numberwithin{theorem}{section}
\numberwithin{proposition}{section}
\renewcommand{\theHtheorem}{\thesection.\arabic{theorem}}
\renewcommand{\theHproposition}{\thesection.\arabic{proposition}}
\section{Complete proofs and theoretical extensions}
\label{app:proofs}
The first three sections establish finite-sample guarantees. The next
sections prove the reward and score results and their explanatory models.
Counterexamples clarify the assumptions behind each conclusion.

\subsection{Proof of Theorem~\ref{thm:crc}}
\label{app:crc-proof}
\paragraph{Step 1: preserve exchangeability through the decoder.}
Let $F_i$ be the complete problem family and $K(\cdot\mid F_i)$ the fixed
kernel for trajectory generation, budget forcing, score extraction, and
outcome grading. The distribution of
$((F_i,Y_i,S_i))_{i=1}^{n+1}$ integrates the product
$\prod_iK(dy_i,ds_i\mid F_i)$ against the exchangeable law of
$(F_i)_{i=1}^{n+1}$. Permuting all indices leaves both factors unchanged.
Therefore the augmented observations are exchangeable.
Token dependence, stochastic decoding, and deterministic external delimiters
are internal to $K$; they do not change this argument.
Independent random seeds across families implement the product-kernel
assumption; pairing seeds across \emph{comparison arms} preserves each
arm's argument.

\paragraph{Step 2: reduce calibration to a rank rule.}
Write $N=n+1$ and define $Z_i=S_i$ when $Y_i=W$, and $Z_i=+\infty$ otherwise.
For $j=\lfloor N\alpha\rfloor\geq1$, the threshold from $n$ calibration
observations is their $j$th ordered $Z$ value: when fewer than $j$ wrong
scores are finite, it is $+\infty$. Since $\alpha<1$, $j\leq n$.
For each $i\leq N$, let $Z_{(j)}^{-i}$ be the $j$th order statistic after
omitting $i$, and set
\[
E_i=\mathbf1\{Y_i=W,\ Z_i<Z_{(j)}^{-i}\}.
\]
For any deterministic collection of $N$ observations,
\begin{equation}
\sum_{i=1}^N E_i\leq j.
\label{eq:rank-count}
\end{equation}
To see this with ties, group finite wrong scores by value. A score in a
group whose cumulative count is $r$ is strictly below the $j$th
leave-one-out value only if $r\leq j$: the other $r-1$ observations at or
below that value would otherwise already fill the first $j$ ranks.
Thus only entire groups contained within the first $j$ ranks can count.
If there are at most $j$ finite wrong scores, all of them can count and
their number is still at most $j$. Infinite scores never satisfy a strict
comparison with an infinite threshold.

\paragraph{Step 3: average over the test position.}
Exchangeability gives $\mathbb E E_N=N^{-1}\mathbb E\sum_iE_i$.
Equation~\eqref{eq:rank-count} yields
$\mathbb E E_N\leq j/N\leq\alpha$.
When $j=0$, the declared reject-all rule has risk zero.
This proves every case in \eqref{eq:crc-cut}.
If all wrong scores are finite and almost surely distinct, and $M_W$ is the
number of wrong outputs among all $N$ observations, the proof additionally
gives the exact identity
\[
\mathbb E E_N=\frac{\mathbb E[\min\{j,M_W\}]}{N}.
\]
Consequently the rule can accept every valid answer when wrong outputs are
already rare; reaching the risk target is not required for validity.

\paragraph{Step 4: relation to the corrected empirical risk.}
For $j\geq1$, the inequality
$1+\sum_i\mathbf1\{Y_i=W,S_i<\tau\}\leq N\alpha$
is equivalent to allowing at most $j-1$ calibration wrong submissions.
The largest admissible threshold is the $j$th wrong score, unless fewer
than $j$ such scores exist, in which case all finite valid scores can pass.
With strict acceptance, $\mathbf1\{S<\tau\}$ is \emph{left-continuous} in
$\tau$, and a tied threshold rejects the entire tie group.
The rank proof above avoids any continuity assumption.
\qed

\subsection{Risk under shift, repeated test items, and seed averaging}
\label{app:shift-proof}
For any fixed threshold $t$,
$F_W^Q(t)\leq F_W^P(t)+d_W$ where
$d_W=\sup_t(F_W^Q(t)-F_W^P(t))$.
Taking the expectation over a source calibration set and using
Theorem~\ref{thm:crc} proves the first inequality in \eqref{eq:shift};
the second follows because each event $\{Y=W,S<t\}$ is measurable and
bounded by total variation.
When both domains use the same decoder and verifier kernel,
the data-processing inequality further bounds induced total variation
by that between the complete problem-family distributions.
The bound is an \emph{expectation over calibration sets}; it does not assert
$F_W^P(\widehat\tau)\leq\alpha$ for every realized calibration split.

If calibration and the $m$ future test families are jointly exchangeable,
linearity gives
$\mathbb E[m^{-1}\sum_{i=1}^m L_i(\widehat\tau)]\leq\alpha$.
Under i.i.d.\ test draws, conditionally on calibration this test average has
mean $F_W^P(\widehat\tau)$ and variance
$F_W^P(\widehat\tau)(1-F_W^P(\widehat\tau))/m$.
The conditional mean itself varies with calibration.
This is why an empirical test rate or its one-sided confidence limit can
exceed $\alpha$ without contradicting the expected-risk statement.

If two frozen training seeds are calibrated separately and one is selected
uniformly for deployment, the deployed risk is the average of their risks.
Linearity preserves the same expected-risk bound.
Reporting this average within each family estimates the randomized
single-seed deployment policy; it is not a two-rollout voting procedure,
and the seeds do not double the independent family count.

\subsection{Conditional selective-error certificate and proof}
\label{app:ltt-proof}
Fix a solver, decoder, verifier, score, and calibration-independent grid
$\Lambda$ of $M$ cutoffs, a selective-error target
$\alpha_{\rm sel}\in(0,1)$, and error probability $\delta\in(0,1)$.
For $n$ calibration families, define
\[
 N_\tau=\sum_{i=1}^n\ind\{S_i<\tau\},\qquad
 W_\tau=\sum_{i=1}^n\ind\{Y_i=W,S_i<\tau\}.
\]
The one-sided exact binomial test is
\begin{equation}
 p_\tau=\Pr\{\operatorname{Bin}(N_\tau,\alpha_{\rm sel})\le W_\tau\},
 \qquad p_\tau=1\ \text{if }N_\tau=0.
 \label{eq:ltt-test}
\end{equation}

\begin{theorem}[Fixed-grid selective-error certificate]
\label{thm:ltt}
Let calibration families and an independent test family be independent and
identically distributed (i.i.d.). Apply the same frozen decoding kernel
independently to each family, with the score and grid fixed independently
of these data. Select the largest
$\tau\in\Lambda$ satisfying $p_\tau\le\delta/M$, or issue no certificate.
With probability at least $1-\delta$ over calibration, every issued cutoff
has $P(Y_{\rm test}=W\mid S_{\rm test}<\widehat\tau,
\mathcal D_{\rm cal})\le\alpha_{\rm sel}$.
\end{theorem}
\emph{Proof.} The conditional binomial argument and simultaneous testing
steps follow.

\paragraph{Step 1: condition on the accepted sample size.}
For a fixed calibration-independent threshold $\tau$, set
$A_i=\mathbf1\{S_i<\tau\}$ and
$B_i=\mathbf1\{Y_i=W,S_i<\tau\}$.
Under i.i.d.\ families the pairs $(A_i,B_i)$ are i.i.d.
If $c_\tau=P(A_i=1)>0$, then conditioning on the set of accepted indices
makes their wrong indicators independent Bernoulli variables with common
parameter $r_\tau=P(B_i=1\mid A_i=1)$. Hence
\[
W_\tau\mid N_\tau=k\sim\operatorname{Bin}(k,r_\tau).
\]
If $c_\tau=0$, $N_\tau=0$ almost surely and the threshold cannot be certified.

\paragraph{Step 2: verify each p-value.}
Let $F_{k,a}$ be the binomial CDF with parameters $(k,a)$.
For $r_\tau\geq a=\alpha_{\rm sel}$ and $u\in[0,1]$, the event
$F_{k,a}(W_\tau)\leq u$ is a lower-tail event
$W_\tau\leq b$ for some integer $b$, or is empty.
By binomial stochastic ordering, a larger parameter has no larger
lower-tail probability:
\[
P_{r_\tau}\{F_{k,a}(W_\tau)\leq u\mid N_\tau=k\}
\leq F_{k,a}(b)\leq u.
\]
The $k=0$ convention $p_\tau=1$ is valid as well.
Averaging over $N_\tau$ proves super-uniformity under every bad
threshold with $r_\tau>\alpha_{\rm sel}$.

\paragraph{Step 3: control every selected threshold.}
Let $\Lambda_{\rm bad}=\{\tau\in\Lambda:r_\tau>\alpha_{\rm sel}\}$.
A union bound yields
\[
P\{\exists\tau\in\Lambda_{\rm bad}:p_\tau\leq\delta/M\}
\leq|\Lambda_{\rm bad}|\delta/M\leq\delta.
\]
Outside this event, all passing thresholds have conditional error at most
$\alpha_{\rm sel}$, so selecting the largest one is valid.
No monotonicity of $r_\tau$ is needed. If none passes, no conditional-risk
certificate is issued. The independent future family's risk conditional
on calibration and the selected threshold is $r_{\widehat\tau}$.
\qed

\paragraph{Accepted-volume requirement.}
At $W_\tau=0$, the p-value is $(1-\alpha_{\rm sel})^{N_\tau}$.
The binomial CDF increases with the wrong count, so zero errors minimize
the p-value at fixed $N_\tau$; issuance requires
\[
N_\tau\geq
\left\lceil\frac{\log(\delta/M)}{\log(1-\alpha_{\rm sel})}\right\rceil.
\]
With $M=10,\delta=0.1$, the necessary accepted counts at conditional-error
targets 10\%, 15\%, 20\%, and 30\% are respectively 44, 29, 21, and 13.
These are necessary volume conditions, not a guarantee of issuance.
The family-wise guarantee covers one declared model--domain--budget
and error-target configuration; reporting multiple configurations gives no
joint guarantee across them.

\subsection{Proofs and scope of Proposition~\ref{prop:penalty-signal}}
\label{app:penalty-proof}

\subsubsection{Group advantages and mixed failures}
Write $\mu_\lambda=p_R-\lambda p_W$ for the expected reward.
Condition on the first trajectory. Independence of the remaining
$G-1$ rollouts gives
\begin{align*}
 \mathbb E[A_1\mid Y_1=y]
 &=r_\lambda(y)-\frac{r_\lambda(y)+(G-1)\mu_\lambda}{G}\\
 &=\gamma_G\{r_\lambda(y)-\mu_\lambda\}.
\end{align*}
Substituting $r_\lambda(R)=1$, $r_\lambda(W)=-\lambda$,
and $r_\lambda(U)=0$ proves
\eqref{eq:conditional-group-advantage}. Conditional statements are
understood only for outcomes having positive probability. If
$p_R+p_W>0$, then
\[
 \lambda p_W-p_R>0
 \ \Longleftrightarrow\
 (1+\lambda)p_R<\lambda(p_R+p_W)
 \ \Longleftrightarrow\
 \frac{p_R}{p_R+p_W}<\frac{\lambda}{1+\lambda}.
\]
If $p_R=p_W=0$, all rewards and advantages are zero and the precision
ratio is undefined, so no phase-transition claim is needed.
For an actual group with counts $(N_R,N_W,N_U)$, every $U$ rollout has
\[
 A_U=\frac{\lambda N_W-N_R}{G}.
\]
At $\lambda=0$, this is nonpositive in every group, strengthening the
conditional-expectation statement. Sampled $R$ advantages are always
nonnegative, and sampled $W$ advantages are always nonpositive.

\paragraph{Reward-varying groups.}
For $\lambda=0$, a group has a nonconstant reward vector exactly when it
contains both an $R$ and a non-$R$. Its probability is
\[
 1-p_R^G-(1-p_R)^G.
\]
For $\lambda>0$, the three outcome classes have distinct rewards, so
the corresponding probability is
$1-p_R^G-p_W^G-p_U^G$. Subtracting proves
\eqref{eq:failure-only-probability}. Equivalently, the additional event
is $\{N_R=0,N_W>0,N_U>0\}$, on which $A_U=\lambda N_W/G>0$ and
$A_W=-\lambda N_U/G<0$.
The phrase reward-varying does not assert that an arbitrary shared
parameterization must have a nonzero parameter gradient on that group.

\subsubsection{Positive non-submission rewards: proof of Corollary~\ref{cor:positive-u-main}}
\label{app:positive-u-proof}

\begin{corollary}[A common non-submission reward raises the effective penalty]
\label{cor:positive-u-main}
If $r(R)=1$, $r(W)=-\lambda$, and $r(U)=\rho$ with
$\lambda\ge0$ and $0\le\rho<1$, the group advantages satisfy
\begin{equation}
 A_g^{\lambda,\rho}=(1-\rho)A_g^{\lambda_{\rm eff},0},
 \qquad \lambda_{\rm eff}=\frac{\lambda+\rho}{1-\rho}.
 \label{eq:effective-penalty-main}
\end{equation}
The break-even submitted-answer precision becomes
$(\lambda+\rho)/(1+\lambda)$.
\end{corollary}
\emph{Proof.} The conditional calculation and affine reward identity are as follows.

Let $r_{\lambda,\rho}(R)=1$, $r_{\lambda,\rho}(W)=-\lambda$, and
$r_{\lambda,\rho}(U)=\rho$, with $\lambda\geq0$ and $0\leq\rho<1$.
For the same independent group of $G\geq2$ rollouts, write
$\gamma_G=(G-1)/G$. Conditioning as in
Proposition~\ref{prop:penalty-signal} gives, whenever $p_U>0$,
\begin{equation}
 \mathbb E[A_1\mid U]
 =\gamma_G\{\rho(p_R+p_W)+\lambda p_W-p_R\}.
 \label{eq:positive-u-advantage}
\end{equation}
Indeed, the mean reward is
$\mu_{\lambda,\rho}=p_R-\lambda p_W+\rho p_U$ and the conditional
advantage is $\gamma_G(\rho-\mu_{\lambda,\rho})$.
For $p_R+p_W>0$, its sign is positive exactly when
\begin{equation}
 \frac{p_R}{p_R+p_W}<\frac{\lambda+\rho}{1+\lambda}.
 \label{eq:positive-u-threshold}
\end{equation}
If $p_R=p_W=0$, every group reward equals $\rho$ and every advantage
is zero.

The affine transformation
\[
 r'=\frac{r_{\lambda,\rho}-\rho}{1-\rho}
   =\left(1,-\lambda_{\mathrm{eff}},0\right),
 \qquad
 \lambda_{\mathrm{eff}}=\frac{\lambda+\rho}{1-\rho},
\]
has an exact groupwise consequence:
$A_g^{\lambda,\rho}=(1-\rho)A_g^{\lambda_{\mathrm{eff}},0}$.
The additive constant cancels in group centering, and the positive
rescaling multiplies the raw advantage-weighted gradient by $1-\rho$.
This is an identity of rewards, advantages, and the corresponding raw
surrogate gradient; it does not identify a shared neural policy's
trajectory under AdamW, gradient clipping, or additional loss terms.
The restriction $\rho<1$ ensures a positive scaling and keeps a correct
answer strictly preferable to non-submission. At $\rho=1$, those
outcomes tie and this reduction is undefined.

In the specified attempt--solver model, the objective becomes
\[
 J_{\lambda,\rho}(a,q)
   =\rho+a\{(1+\lambda)q-\lambda-\rho\}.
\]
With the same product Bernoulli mirror metric, the exact flow is
\[
 \dot a=a(1-a)\{(1+\lambda)q-\lambda-\rho\},\qquad
 \dot q=\kappa(1+\lambda)a q(1-q).
\]
For $\lambda+\rho>0$, put $q_c=(\lambda+\rho)/(1+\lambda)\in(0,1)$.
The invariant
$-q_c\log q-(1-q_c)\log(1-q)+\kappa\log(1-a)$ is unchanged in form.
Consequently, for $a_0,q_0\in(0,1)$ with $q_0<q_c$, the exact
starvation criterion remains
\[
 D_{\mathrm{Ber}}(q_c\Vert q_0)\geq-\kappa\log(1-a_0).
\]
The same proof gives $(a,q)\to(0,q_\infty)$ with $q_\infty\leq q_c$
under this inequality, and recovery to $(1,1)$ under the reverse
strict inequality; $q_0\geq q_c$ also gives recovery.
Thus a common positive non-submission reward raises break-even precision
even without an explicit wrong-answer penalty: at $\lambda=0$, $q_c=\rho$.

Rewarding only explicit refusal is a different four-outcome intervention.
For rewards $(1,-\lambda,\rho_A,0)$ on $(R,W,U_A,U_T)$, where $U_A$ is
explicit refusal and $U_T$ is truncation, the common-$U$ reduction above
does not apply. In particular, when both subtypes have positive
probability,
\[
 \mathbb E[A_1\mid U_A]-\mathbb E[A_1\mid U_T]
   =\gamma_G\rho_A.
\]
Positive $\rho_A$ breaks their reward tie and expressly favors refusal
over truncation at the reward level. It does not by itself specify
probability flow or a deployment marginal-risk guarantee. These
positive-reward shaping variants were not evaluated in the experiments.

\subsubsection{Gradient identity and policy parameterization}
Group-centered policy gradients also admit a U-statistic interpretation
\citep{zhou2026grpo}. The direct covariance derivation below isolates the
effect of the ternary reward and applies to arbitrary rollout features.
Exchangeability within the group gives
$\mathbb E[G^{-1}\sum_g A_g\psi_g]=\mathbb E[A_1\psi_1]$.
Using independence for $j\neq1$,
\begin{align*}
 \mathbb E[A_1\psi_1]
 &=\mathbb E[r_1\psi_1]
  -\frac1G\mathbb E[r_1\psi_1]
  -\frac1G\sum_{j=2}^G\mathbb E[r_j]\mathbb E[\psi_1]\\
 &=\gamma_G\{\mathbb E[r_\lambda\psi]
                  -\mathbb E[r_\lambda]\mathbb E[\psi]\}.
\end{align*}
Here $\psi$ need not be a score function.
If $Q_\theta$ has a differentiable density on a locally fixed support,
rewards have no direct dependence on $\theta$, and differentiation can
be interchanged with integration, then
\[
 \mathbb E_{Q_\theta}\nabla_\theta\log Q_\theta=0,\qquad
 \nabla_\theta\mathbb E_{Q_\theta}r_\lambda
 =\mathbb E_{Q_\theta}[r_\lambda\nabla_\theta\log Q_\theta],
\]
which proves the score-function specialization.

\paragraph{Connection to the implemented surrogate.}
Fix token denominator $GB$ and mask deterministic environment tokens.
The unclipped token-ratio surrogate's gradient at the behavior-policy
point is
\[
 \widehat g=\frac1{GB}\sum_{g=1}^G A_g
  \sum_{t\in\mathcal T_g^{\mathrm{sampled}}}
        \nabla_\theta\log\pi_\theta(y_{g,t}\mid h_{g,t}).
\]
Thus \eqref{eq:group-covariance} applies to the summed token feature
with the constant $1/B$. Fixed $B$ prevents per-trajectory length weighting.
Deterministic prefills, injected end-of-thinking tokens, and forced
stopping are unsampled and contribute no score term.

The experiments use top-$p$ and top-$k$ sampling. If the optimized
log-probability is that of the raw model $\pi_\theta$, while trajectories
come from a transformed law $Q_\theta$, the exact statement is the
covariance identity, not automatically
$\nabla_\theta\mathbb E_{Q_\theta}r_\lambda$.
On a region with a fixed truncation set $\mathcal V(h)$, for example,
$q_\theta(v\mid h)=
\pi_\theta(v\mid h)\mathbf1\{v\in\mathcal V(h)\}/
Z_\theta(h)$ has score
$\nabla\log q_\theta=\nabla\log\pi_\theta-\nabla\log Z_\theta$.
Prefix-dependent normalizer terms generally persist in the reward
covariance; changing truncation sets require separate analysis.

For standard positive-width PPO clipping, ratios equal to one lie in
the unclipped region, so the behavior-point derivative agrees with the
unclipped derivative above. This does not identify later clipped
inner-loop derivatives. AdamW, gradient clipping, parameter sharing,
and finite stochastic updates also prevent equating the raw expected
gradient with the trained LLM's exact probability flow.

\paragraph{Why the sign of a conditional advantage is not a universal drift theorem.}
For diagnostic purposes, consider a terminal three-action softmax with
logits $z_i$, and rescale time to absorb $\gamma_G/B$.
Exact Euclidean gradient ascent on
$J=p_R-\lambda p_W$ has
\[
 \dot z_i=p_i\{r_i-\mu_\lambda\}.
\]
Writing $S_2=p_R^2+p_W^2+p_U^2$ and differentiating $p_R$ gives
\begin{equation}
 \dot p_R=
 p_R^2(1-2p_R+S_2)
 -\lambda p_Rp_W(S_2-p_R-p_W).
 \label{eq:softmax-correct-drift}
\end{equation}
Indeed, $\dot p_R=\|\nabla_zp_R\|^2
-\lambda\langle\nabla_zp_R,\nabla_zp_W\rangle$,
$\|\nabla_zp_R\|^2=p_R^2(1-2p_R+S_2)$, and
$\langle\nabla_zp_R,\nabla_zp_W\rangle
=p_Rp_W(S_2-p_R-p_W)$.
At $(p_R,p_W,p_U)=(0.01,0.30,0.69)$ and $\lambda=1$,
\[
 \dot p_R=0.00015462-0.00076860=-0.00061398.
\]
Here the correct outcome has positive expected advantage, but the
penalty decreases its probability at this state. On a failure-only
group, the empirical logit gradient is exactly
\[
 (\widehat g_R,\widehat g_W,\widehat g_U)
   =\frac{\lambda N_WN_U}{G^2}(0,-1,1).
\]
The first-order change in $p_R$ from that group is therefore
$-p_R\lambda N_WN_U(p_U-p_W)/G^2$, which is negative when
$p_U>p_W$. This exhibits one mechanism for diminishing correct-path
exploration in this parameterization.

Even positive expected $U$ advantage need not increase $p_U$ under
parameter sharing. For scalar $\theta$, take
\[
 (z_R,z_W,z_U)
  =(\log0.4+4\theta,\log0.5,\log0.1+\theta),\quad \lambda=1.
\]
At $\theta=0$, $\mu_\lambda=-0.1$ and
$\mathbb E[A\mid U]=0.1\gamma_G>0$. However,
$\partial_\theta p_U=-0.07$ and
$\partial_\theta J=1.77$, so gradient ascent gives
$\dot p_U=-0.1239$ in rescaled time.

Conversely, the full-simplex natural-gradient/replicator flow is
\begin{equation}
 \dot p_i=p_i(r_i-\mu_\lambda),\qquad
 p_i(t)=\frac{p_i(0)e^{r_it}}{\sum_jp_j(0)e^{r_jt}}.
 \label{eq:terminal-replicator}
\end{equation}
For every interior initial state, $p_R$ increases and tends to one.
Moreover, from the same initial state,
\[
 p_R^\lambda(t)=
 \frac{p_R(0)e^t}
 {p_R(0)e^t+p_U(0)+p_W(0)e^{-\lambda t}}
 \geq p_R^0(t).
\]
Thus reward ordering alone implies neither a universal all-$U$ attractor
nor universal frontier contraction.

\subsubsection{An attempt--solver model with a sharp starvation boundary}
\label{app:attempt-solver-proof}
We now exhibit assumptions under which a genuine boundary attractor
can occur, rather than attributing one to Dr.\ GRPO without a dynamics
model. A gate attempts a solution with probability $a\in(0,1)$;
conditional success probability is $q\in(0,1)$. Otherwise it produces
$U$. Then
\[
 p_R=aq,\quad p_W=a(1-q),\quad p_U=1-a,\qquad
 J_\lambda(a,q)=a\{(1+\lambda)q-\lambda\}.
\]
Use the product Bernoulli mirror metric, with relative solver learning
rate $\kappa>0$:
\[
 M(a,q)=
 \operatorname{diag}\left(
     \frac1{a(1-a)},\,\frac1{\kappa q(1-q)}\right).
\]
Its gradient-ascent flow is
\begin{equation}
 \dot a=a(1-a)\{(1+\lambda)q-\lambda\},\qquad
 \dot q=\kappa(1+\lambda)a q(1-q).
 \label{eq:attempt-solver-flow}
\end{equation}
This is a specified block mirror-flow model. In particular, its metric
is not the Fisher information metric of the full gated trajectory law:
the latter contains an extra visitation factor $a$ in the solver block.
The factor $a$ remaining in $\dot q$ represents learning that slows
when solution attempts become rare.

For $\lambda>0$, write $q_c=\lambda/(1+\lambda)$ and suppose
$q_0<q_c$. Let
\[
 D_{\mathrm{Ber}}(q_c\Vert q_0)
 =q_c\log\frac{q_c}{q_0}
 +(1-q_c)\log\frac{1-q_c}{1-q_0}.
\]
Then the exact dichotomy is
\begin{equation}
 \begin{cases}
  a(t)\to0,\ q(t)\to q_\infty\leq q_c,
  &D_{\mathrm{Ber}}(q_c\Vert q_0)
       \geq-\kappa\log(1-a_0),\\
  a(t)\to1,\ q(t)\to1,
  &D_{\mathrm{Ber}}(q_c\Vert q_0)
       <-\kappa\log(1-a_0).
 \end{cases}
 \label{eq:starvation-boundary}
\end{equation}
The strict first inequality gives $q_\infty<q_c$ and exponentially
vanishing attempt probability. At equality, $q_\infty=q_c$.
For $q_0\geq q_c$, or for $\lambda=0$ with any interior initial
state, both $a$ and $q$ tend to one.

To prove the dichotomy, define
$h_c(q)=-q_c\log q-(1-q_c)\log(1-q)$.
Direct differentiation of \eqref{eq:attempt-solver-flow} yields
\[
 \frac d{dt}\{h_c(q)+\kappa\log(1-a)\}=0.
\]
Consequently,
\begin{equation}
 a(q)=1-(1-a_0)
       \exp\!\left\{\frac{h_c(q_0)-h_c(q)}{\kappa}\right\}.
 \label{eq:attempt-solver-invariant}
\end{equation}
The function $h_c$ decreases strictly on $(0,q_c)$, has its unique
minimum at $q_c$, and increases strictly on $(q_c,1)$.
Also $q(t)$ increases strictly, and both coordinates remain interior at
every finite time.

If $h_c(q_0)+\kappa\log(1-a_0)>h_c(q_c)$, there is a unique
$q_\infty\in(q_0,q_c)$ satisfying
$h_c(q_\infty)=h_c(q_0)+\kappa\log(1-a_0)$.
Equation~\eqref{eq:attempt-solver-invariant} prevents $q$ from crossing
this value with $a>0$. Its limit cannot be smaller, since then both
$a$ and $q(1-q)$ would remain bounded away from zero and $\dot q$
would remain positive. Thus $q\to q_\infty$ and $a\to0$.
Since $\frac d{dt}\operatorname{logit}(a)
=(1+\lambda)(q-q_c)$ tends to a strictly negative constant,
$a$ vanishes exponentially on the logarithmic scale.
The same argument at equality gives $(a,q)\to(0,q_c)$.

If $h_c(q_0)+\kappa\log(1-a_0)<h_c(q_c)$, then
\[
 a(q_c)=1-(1-a_0)
   e^{D_{\mathrm{Ber}}(q_c\Vert q_0)/\kappa}>0.
\]
The trajectory crosses $q_c$ in finite time. Thereafter $a$ and $q$
increase, $q$ cannot converge below one, and
\eqref{eq:attempt-solver-invariant} implies $a\to1$.
Finally, $h_c(q_0)-h_c(q_c)=D_{\mathrm{Ber}}(q_c\Vert q_0)$,
which proves \eqref{eq:starvation-boundary}.
For $q_0\ge q_c$ and $\lambda>0$, $q$ immediately exceeds $q_c$,
after which the same monotone-recovery argument applies.
For $\lambda=0$, $\dot a>0$ and $a\geq a_0>0$;
the equation for $q$ implies $q\to1$, and the equation for
$\operatorname{logit}(a)$ then implies $a\to1$.

\begin{figure}[t]
\centering
\includegraphics[width=.92\linewidth]{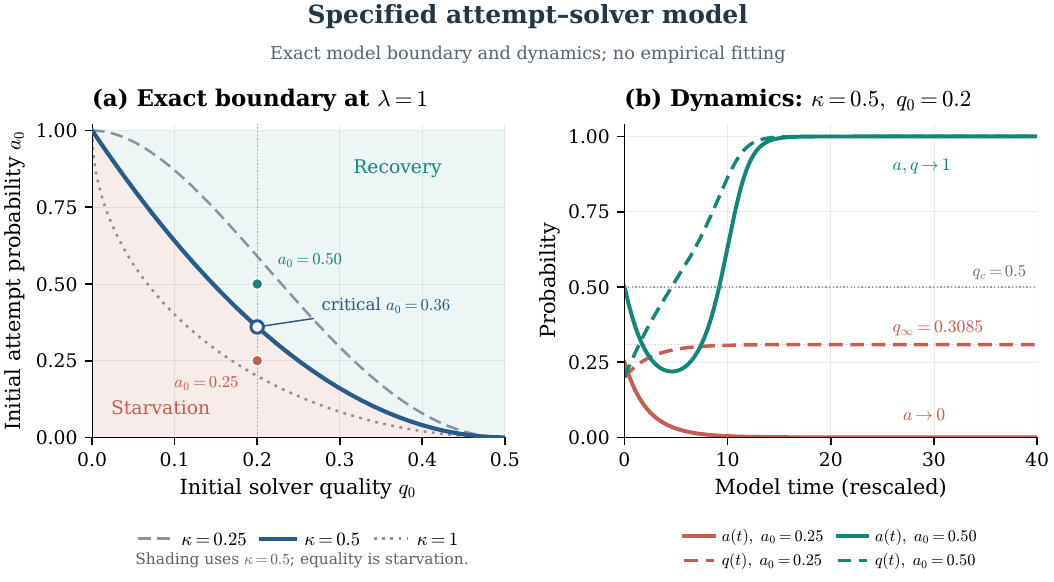}
\caption{Exact boundary and trajectories in the specified attempt--solver model.
At $\lambda=1$, $\kappa=.5$, and $q_0=.2$, recovery requires $a_0>.36$.
The $a_0=.25$ trajectory tends to $(a,q)=(0,.3085146)$; $a_0=.50$
recovers to $(1,1)$. All curves are model calculations, with no parameters
fitted to the experiments.}
\label{fig:penalty-barrier}
\end{figure}

The starvation regime has a finite exploration budget in the following
precise sense:
\begin{align}
 \int_0^\infty a(t)\,dt
 &=\frac{\operatorname{logit}(q_\infty)
            -\operatorname{logit}(q_0)}
          {\kappa(1+\lambda)}<\infty,\nonumber\\
 \int_0^\infty a(t)q(t)\,dt
 &=\frac{\log(1-q_0)-\log(1-q_\infty)}
          {\kappa(1+\lambda)}<\infty .
 \label{eq:finite-exploration}
\end{align}
These identities follow by dividing $\dot q$ by $q(1-q)$ and by
$1-q$, respectively, and integrating. Multiplication by a fixed
rollout intensity gives expected attempted and correct-rollout exposure.
The probability that a group contains any correct rollout,
$1-\{1-a(t)q(t)\}^G$, tends to zero; the probability of an all-$U$
group tends to one. This model is not fitted to the released training
curves.

An even simpler finite-horizon comparison holds when solver quality
is fixed during gate adaptation. With
$v_\lambda=q-\lambda(1-q)$,
\[
 a_\lambda(t)
 =\operatorname{logit}^{-1}
       \{\operatorname{logit}(a_0)+v_\lambda t\}.
\]
For every $\lambda>0$, $q\in(0,1)$, and $t>0$,
$a_\lambda(t)<a_0(t)$, where the subscript zero denotes binary reward
from the same initialization. Thus both the per-rollout success
probability $qa_\lambda(t)$ and
$\int_0^Tqa_\lambda(t)\,dt$ are strictly smaller for every $T>0$.
If $\delta=\lambda-(1+\lambda)q>0$, then
\[
 \int_0^T a_\lambda(t)\,dt
 =-\frac1\delta
   \log\{1-a_0+a_0e^{-\delta T}\}
 \leq\frac{-\log(1-a_0)}{\delta}.
\]
The conditional solver is fixed here; a freely adapting full policy needs
further assumptions.

\subsection{Proofs and scope of Proposition~\ref{prop:penalty-selection}}
\label{app:selection-proof}

\begin{proposition}[Deployment-dependent selection and reward-equivalent abstention]
\label{prop:penalty-selection}
For a frozen candidate generator, let
$\eta(z)=P(R\mid z,\mathrm{valid})$ be posterior correctness from selector
information $z$. For $\alpha>0$, maximizing correct submissions with
$W\le\alpha$ over randomized selectors measurable in $z$ gives
\begin{equation}
 g^\star(z)=\ind\{\eta(z)>\lambda^\star/(1+\lambda^\star)\},
 \label{eq:bayes-penalty-selector}
\end{equation}
with possible randomization at equality. The optimal dual multiplier depends
in general on both $\alpha$ and the deployment population, including when
$\eta(z)$ is unchanged under shift. If explicit abstention and truncation
both receive zero reward, redistributing probability between them at fixed
$(p_R,p_W)$ leaves the objective unchanged; their conditional expected
advantages are identical.
\end{proposition}
\emph{Proof.} We derive the optimal selector, exhibit its population dependence,
and verify reward equivalence.

\paragraph{Bayes rule and ties.}
Let $\nu_P$ be the subprobability law of valid candidates and their selector
information, so $\nu_P(\mathcal Z)=P(\mathrm{valid})$. For a randomized
selector $g\colon\mathcal Z\to[0,1]$, write
$R_P(g)=\int g\eta\,d\nu_P$ and
$W_P(g)=\int g(1-\eta)\,d\nu_P$. Assume $\alpha>0$.
For $\lambda\geq0$, the Lagrangian is
\[
 {\cal L}(g,\lambda)
 =\lambda\alpha+
   \int g(z)\{\eta(z)-\lambda(1-\eta(z))\}\,d\nu_P(z).
\]
Pointwise maximization over $g(z)\in[0,1]$ gives
\eqref{eq:bayes-penalty-selector}; at zero coefficient any value in
$[0,1]$ is optimal. Retaining the constant term, the resulting dual
objective is
\[
 d_P(\lambda)=\lambda\alpha+
      \int\{\eta-\lambda(1-\eta)\}_+\,d\nu_P.
\]
It is continuous and convex, and $d_P(\lambda)\geq\lambda\alpha$,
so it attains a finite minimum for $\alpha>0$.
At an interior minimizer, the subgradient condition permits a
randomization on the equality set making $W_P(g^\star)=\alpha$.
At $\lambda^\star=0$, one can choose a maximizer with
$W_P(g^\star)\leq\alpha$.
Equivalently, order points by the posterior odds
$\eta/(1-\eta)$ under the wrong-mass measure
$(1-\eta)d\nu_P$ and fill up to a budget of mass $\alpha$, randomizing at a
boundary atom. Points with $\eta=1$ are accepted at no wrong-answer cost.

For completeness, any feasible $g$ and the resulting $g^\star$ satisfy
\begin{align*}
 R_P(g)-R_P(g^\star)
 &=\int(g-g^\star)\{\eta-\lambda^\star(1-\eta)\}\,d\nu_P\\
 &\quad+\lambda^\star\{W_P(g)-W_P(g^\star)\}\leq0.
\end{align*}
The first term is nonpositive by pointwise maximization; the second is
nonpositive by feasibility and complementary slackness. This proves
optimality directly. If the budget exceeds the raw wrong-answer
mass, accepting all valid candidates is optimal and $\lambda=0$
suffices. At $\alpha=0$, accepting only posterior-certain correct
candidates is optimal, and a finite dual threshold need not exist.

\paragraph{A concrete distribution-shift counterexample.}
Let all candidates be valid, let $Z\in[0,1]$, and let correctness
conditional on $Z=z$ be Bernoulli with probability $\eta(z)=z$.
On the source population $Z\sim\mathrm{Unif}[0,1]$, a cutoff $t$
has
\[
 W_{\mathrm{src}}(t)=\int_t^1(1-z)\,dz
                      =\frac{(1-t)^2}{2}.
\]
For $\alpha=0.02$, the unique cutoff is $t=0.8$ and
$\lambda^\star_{\mathrm{src}}=4$.
Now shift only the feature distribution to
$Z\sim\mathrm{Unif}[0.5,1]$, retaining the identical correctness
function $\eta(z)=z$. The same threshold has
\[
 W_{\mathrm{tgt}}(0.8)
     =2\int_{0.8}^1(1-z)\,dz=0.04.
\]
The target-optimal threshold is
$t^\star_{\mathrm{tgt}}=1-\sqrt{0.02}$ and
$\lambda^\star_{\mathrm{tgt}}=1/\sqrt{0.02}-1\approx6.0711$.
Even this shift toward easier candidates doubles the marginal wrong
rate of the fixed rule because it changes accepted population mass.
The source optimum also depends on the risk target:
$\lambda^\star(\alpha)=1/\sqrt{2\alpha}-1$ for $0<\alpha<1/2$.

This is non-transferability of the optimal operating point in general,
not a claim that every shift changes the optimum or that no fixed
penalty can ever be conservative. Under an oracle posterior threshold,
\[
 W_P(g_\lambda)\leq
 \frac{\mathbb P(\mathrm{accepted})}{1+\lambda}
 \leq\frac1{1+\lambda}.
\]
Thus $\lambda\geq\alpha^{-1}-1$ is a distribution-uniform sufficient
choice in this ideal Bayes problem, often at a substantial coverage
cost. The bound is sharp in the supremum over populations by
concentrating mass just above the acceptance threshold. A trained
neural penalty policy is not assumed to implement an oracle posterior
threshold. None of these facts requires full solver retraining:
changing a post-hoc selector changes its operating point with weights
held fixed.

\paragraph{Reward degeneracy and the role of geometry.}
With $(R,W,U_A,U_T)$ rewards $(1,-\lambda,0,0)$,
$J=p_R-\lambda p_W$ does not contain $p_A$ or $p_T$ separately.
Hence the feasible perturbation
$(p_A,p_T)\mapsto(p_A+\varepsilon,p_T-\varepsilon)$ has zero
directional derivative. Conditioning as above gives
$\mathbb E[A\mid U_A]=\mathbb E[A\mid U_T]=-\gamma_G\mu_\lambda$.

In the full-simplex replicator geometry,
$\dot p_A=-\mu_\lambda p_A$ and
$\dot p_T=-\mu_\lambda p_T$, so
\[
 \frac d{dt}\log\frac{p_A}{p_T}=0.
\]
Their odds preserve their initialization; when both derivatives are
nonzero, their absolute flow ratio equals $p_A/p_T$.
Euclidean ascent in independent outcome logits has
$\dot z_A=-\mu_\lambda p_A$ and
$\dot z_T=-\mu_\lambda p_T$, giving
\[
 \frac d{dt}\log\frac{p_A}{p_T}
 =-\mu_\lambda(p_A-p_T).
\]
Thus even two standard geometries give different relative flows.
Shared neural parameters or Adam preconditioning can give still
different behavior. A reward preference for explicit abstention over
truncation would break the tie, but alone gives no deployment marginal-risk
guarantee.

\paragraph{What can be proved about frontier contraction.}
For any frozen generator and any selector,
$R_P(g)\leq\mathbb P_P(Y=R)$: filtering cannot create a correct
candidate. Sharper set inclusion holds under an explicit
thinning assumption. Suppose a second generator preserves the
original candidate and its selector information but retains it with
probability $h(Z)\in[0,1]$, independently of correctness conditional
on $Z$; otherwise it returns $U$. Its valid-candidate law is
$h\,d\nu_P$. Any selector $g$ on the thinned generator attains
\[
 (R,W)=
 \left(\int hg\eta\,d\nu_P,\;
       \int hg(1-\eta)\,d\nu_P\right),
\]
which the original generator attains using the randomized selector
$hg$. Its attainable set of $(R,W)$ pairs is therefore a subset of
the original generator's attainable set.

With valid coverage $a$ and fixed correctness probability $q\in(0,1)$,
the attempt model gives
\[
 \max_{g:\,W(g)\leq\alpha}R(g)
   =q\min\left\{a,\frac{\alpha}{1-q}\right\}.
\]
Reducing $a$ can strictly reduce the frontier once the smaller
candidate supply binds, while sufficiently tight risk budgets give
the same optimum. This is a sufficient condition for contraction,
not a theorem that arbitrary penalty training is such a thinning.
Training can alter both candidate quality and the score distribution.
Indeed, \eqref{eq:terminal-replicator} gives a counterexample to any
universal contraction assertion: increasing $\lambda$ there raises
$p_R(t)$ and lowers $p_W(t)$ from the same initial state.

\subsection{Score anatomy: complete proofs and boundaries}
\label{app:score-proofs}

\subsubsection{Length and discrimination}
\label{app:length-proof}

In a continuous length-factor surrogate with independent Gaussian log
factors, suppose $\E D=d$, $\E B_\ell=b$ and variances $v_D,v_B>0$. Writing $\Phi$ for the standard normal cumulative distribution function,
\begin{equation}
 \operatorname{AUC}(S_{\rm sum})
 =\Phi\!\left(\frac{d-b}{\sqrt{v_D+v_B}}\right).
 \label{eq:length-main-gaussian}
\end{equation}
Even if $d>0$, the ranking reverses when $b>d$.

For the general Gaussian extension, use $B=B_\ell$ from Proposition~\ref{prop:length-discrimination}.
In a stylized positive-factor model, now allowing $L$ to be real valued,
suppose the within-class vectors
$(\log L_y,\log\bar\ell_y)$ are jointly Gaussian.  Define
$d=\mathbb E\log\bar\ell_W-\mathbb E\log\bar\ell_R$,
$b=\mathbb E\log L_R-\mathbb E\log L_W$,
$v_U=\sum_{y\in\{R,W\}}\operatorname{Var}(\log\bar\ell_y)$,
$v_L=\sum_y\operatorname{Var}(\log L_y)$, and
$c=\sum_y\operatorname{Cov}(\log L_y,\log\bar\ell_y)$.
For $v_U>0$ and $v_U+v_L+2c>0$,
\begin{equation}
 \operatorname{AUC}(\bar\ell)=\Phi\!\left(\frac d{\sqrt{v_U}}\right),\qquad
 \operatorname{AUC}(S_{\rm sum})
 =\Phi\!\left(\frac{d-b}{\sqrt{v_U+v_L+2c}}\right).
 \label{eq:gaussian-length-auroc}
\end{equation}
When $c=0$, the independent-factor specialization uses $v_D=v_U$ and $v_B=v_L$.
Consequently the summed score reverses direction exactly when $b>d$ in this
model.  If $d>0$, $b\ge0$, and the two log factors are independent within each
class, length strictly reduces AUROC whenever $b>0$ or $v_L>0$.

\begin{proof}[Proof of Proposition~\ref{prop:length-discrimination} and its Gaussian extension]
The factorization follows directly from the definition of the arithmetic
mean.  Since the logarithm is strictly increasing on $(0,\infty)$,
\[
 L_W\bar\ell_W>L_R\bar\ell_R
 \quad\Longleftrightarrow\quad
 \log\bar\ell_W-\log\bar\ell_R
 >\log L_R-\log L_W.
\]
The same equivalence holds with equality.  Multiplying the strict-event
probability by one and the tie-event probability by one half proves
\eqref{eq:exact-length-auroc}.  If a span has zero average surprise, omit
only the logarithmic reparameterization; the product identity and original
AUROC definition remain valid.

For the Gaussian model, independence of the two class draws and the
stated within-class joint normality imply that $D$ is normal with mean
$d$ and variance $v_U$.  Also,
\[
 D-B=(\log L_W+\log\bar\ell_W)
       -(\log L_R+\log\bar\ell_R)
\]
is normal with mean $d-b$ and variance $v_U+v_L+2c$.  A normal random
variable with mean $m$ and positive standard deviation $s$ exceeds zero
with probability $\Phi(m/s)$ and has no atom at zero.  This proves
\eqref{eq:gaussian-length-auroc}.  Since $\Phi$ is strictly increasing and
$\Phi(0)=1/2$, the summed score has AUROC below $1/2$ exactly when
$d-b<0$.  If the factors are independent within each class, $c=0$.
When $d>0$ and $b\ge0$, either $d-b\le0$, in which case the comparison
with $d/\sqrt{v_U}>0$ is immediate, or
\[
 \frac{d-b}{\sqrt{v_U+v_L}}
 \le\frac d{\sqrt{v_U}},
\]
with strict inequality when $b>0$ or $v_L>0$.  This proves the stated
degradation claim.  With a zero Gaussian variance, the corresponding
score gap is deterministic and its AUROC is $\psi$ of its mean; this
also describes the degenerate boundaries omitted from the displayed
positive-variance formula.
\end{proof}

An exact accounting of the direction changes is useful.  If
$\Pr(D=0)=\Pr(D=B)=0$, then
\begin{equation}
 \operatorname{AUC}(S_{\rm sum})-\operatorname{AUC}(\bar\ell)
 =\Pr(B<D\le0)-\Pr(0<D\le B).
 \label{eq:pairwise-length-gains-losses}
\end{equation}
Subtracting the indicators gives $+1$ only when an incorrect ordering
becomes correct, and $-1$ only for the reverse change.
Without the no-ties condition one simply uses
$\mathbb E[\psi(D-B)-\psi(D)]$, which is exact for all distributions.
A direct sufficient condition for non-increasing AUROC is $B\ge0$
almost surely for the independent right--wrong pair, because $\psi$ is
nondecreasing.  Marginal stochastic dominance of lengths is substantially
weaker than this pairwise condition.

\paragraph{Why stochastic dominance is insufficient.}
Let a right draw equal $(L_R,\bar\ell_R)=(2,1)$ with probability $0.8$
and $(4,4)$ with probability $0.2$.  Let an independent wrong draw
equal $(1,3)$ with probability $0.8$ and $(3,0.8)$ with probability
$0.2$.  The right length stochastically dominates the wrong length:
its quantiles exceed the latter's by one.  The mean-score AUROC is
$0.8\cdot0.8=0.64$.  The summed scores are $2$ or $16$ on right
draws and $3$ or $2.4$ on wrong draws, giving AUROC $0.8$.  Hence even
strict length dominance and positive mean-score discrimination can
coexist with an improvement after summation.  Dependence between length
and token surprise matters.

\paragraph{Why a median comparison is insufficient.}
Let $L_W=1$, $L_R=1$ with probability $0.49$ and $L_R=2$ otherwise.
Let $\bar\ell_R=1$, and independently set $\log\bar\ell_W=0.1$ with
probability $0.51$ and $1$ otherwise.  Then the unique median of $B$
is $\log2$, exceeding the unique median $0.1$ of $D$, whereas
\[
 \operatorname{AUC}(S_{\rm sum})
 =0.49+0.51\cdot0.49=0.7399>\tfrac12.
\]
The mean score has AUROC one.  Means alone are insufficient under arbitrary
sub-Gaussian assumptions: take $B=\log2$ and independently
$D=1$ with probability $0.51$ and $D=0.1$ otherwise.  Both variables are
bounded and therefore sub-Gaussian, $\mathbb ED=0.559<\log2$, but
$\Pr(D>B)=0.51$.  The Gaussian sign-reversal criterion requires normality
or another restriction on the full centered contrast distribution.

\subsubsection{Finite empirical CVaR, rounding, and replication}
\label{app:tail-proof}

\begin{proof}[Primal and dual identities in Proposition~\ref{prop:tail-aggregation}]
For every $\eta$ and every $k$-element subset $J$,
\[
 \eta+\frac1k\sum_{t=1}^L(z_t-\eta)_+
 \ge \eta+\frac1k\sum_{t\in J}(z_t-\eta)
 =\frac1k\sum_{t\in J}z_t.
\]
Taking $J$ to be a top-$k$ subset gives a lower bound $T_k(z)$.
At $\eta=z_{(k)}$, the expression equals $T_k(z)$: values tied at
$z_{(k)}$ contribute zero excess, and filling their remaining positions
with $z_{(k)}$ supplies exactly the top-$k$ sum.  This also works for
$k=L$, or one can take any $\eta\le z_{(L)}$.

For the dual identity, let $w_t\in[0,1/k]$ and $\sum_t w_t=1$.
Writing $z_t=\eta+(z_t-\eta)$ gives
\[
 w^\top z\le\eta+\frac1k\sum_t(z_t-\eta)_+.
\]
At the minimizing $\eta$ this is at most $T_k(z)$.  Choosing
$w_t=\mathbf1\{t\in J\}/k$ for a top-$k$ subset attains $T_k(z)$.
Empirical CVaR has coefficient $1/(\gamma L)$ for tail mass $\gamma$,
so $\gamma=k/L$ identifies the two expressions.

For $z'=z+u e_i$, $u\ge0$, nonnegative dual weights imply monotonicity.
An optimizer $w'$ at $z'$ gives
\[
 T_k(z')-T_k(z)\le (w')^\top(z'-z)=u w'_i\le u/k.
\]
The bound is attained when coordinate $i$ already belongs to a strictly
separated top-$k$ set.  Setting $k=1$ gives the maximum's sharp bound.
\end{proof}

\begin{lemma}[Exact rounding error and replication bound]
\label{lem:rounded-cvar}
Let $0<\beta\le1$, $a=\beta L$, $k=\lceil a\rceil$, and
$r=k-a$.  Denote fixed-tail-mass empirical CVaR by
\[
 C_\beta(z)=\min_\eta\left\{\eta+\frac1{\beta L}
                   \sum_{t=1}^L(z_t-\eta)_+\right\}.
\]
Then
\begin{equation}
 C_\beta(z)-T_k(z)
   =\frac r a\bigl(T_k(z)-z_{(k)}\bigr)
   =\frac r k\bigl(C_\beta(z)-z_{(k)}\bigr),
 \label{eq:rounding-exact}
\end{equation}
and, with $\operatorname{range}(z)=z_{(1)}-z_{(L)}$,
\[
 0\le C_\beta(z)-T_k(z)
 \le\frac{k-\beta L}{k}\operatorname{range}(z).
\]
If $z^{[c]}$ repeats each coordinate $c$ times, for an integer $c\ge1$,
then
\begin{equation}
 0\le T_{\lceil\beta cL\rceil}(z^{[c]})-T_k(z)
 \le \frac{k-\beta L}{k}\operatorname{range}(z).
 \label{eq:replication-bound}
\end{equation}
Fixed-$\beta$ CVaR is exactly replication-invariant.  The rounded score
is exactly replication-invariant whenever $\beta L$ is integral; it need
not be so otherwise.
\end{lemma}

\begin{proof}
For noninteger $a$, write $a=m+\theta$, where $m=\lfloor a\rfloor$
and $\theta\in(0,1)$.  Evaluating the piecewise-linear objective at
$\eta=z_{(m+1)}$ gives
\[
 C_\beta(z)=\frac{\sum_{j=1}^m z_{(j)}+\theta z_{(m+1)}}a.
\]
To see that this is minimal, observe that on each interval between
successive ordered coordinates the derivative is $1-j/a$, where $j$
coordinates exceed $\eta$; the slope changes from nonpositive to
nonnegative at this quantile.  Ties give the same value.  For integer
$a$, the analogous value is the average of the first $a$ coordinates.
Thus in both cases
\[
 a C_\beta(z)=kT_k(z)-r z_{(k)}.
\]
Rearrangement yields both identities in \eqref{eq:rounding-exact}.
Because $z_{(k)}\le C_\beta(z)\le z_{(1)}$, their common value is
nonnegative and at most $r\operatorname{range}(z)/k$.

Repeating coordinates preserves their empirical probability measure,
so the fixed-$\beta$ CVaR objective and its value are unchanged.  Moreover
\[
 \beta\le\frac{\lceil c\beta L\rceil}{cL}
 \le\frac{\lceil\beta L\rceil}{L}.
\]
An upper-tail average is nonincreasing in its tail mass.  This follows
directly by adding lower quantile values to the averaged upper portion,
or from the dual representation.  It follows that
\[
 T_k(z)\le T_{\lceil c\beta L\rceil}(z^{[c]})\le C_\beta(z).
\]
Combine this with the rounding bound.  For integral $a$, both endpoints
coincide, giving exact invariance.
\end{proof}

For example, with $\beta=0.1$, $L=11$, one coordinate equal to one and
the other ten equal to zero, the implemented score is $1/2$.
Repeating each coordinate ten times gives a score $10/11$.  Thus the
rounded score removes linear growth with replication but is not exactly
replication-invariant for short spans.  The bound is pathwise: an absolute
small-error conclusion additionally requires control of the realized
score range.  NLL has no universal finite upper bound.  Replication also
does not model an arbitrary change of tokenization, which can change the
empirical surprise distribution itself.

\subsubsection{Sparse signal, background noise, and selective submission}
\label{app:sparse-proof}

For an explanatory model, fix $L$ and a set $C$ of $k$ critical positions.
Conditional on class $y\in\{0,1\}$, with $0=R$ and $1=W$, let
\begin{equation}
 \ell_t^{(y)}=\mu_t+y\delta_t+\varepsilon_t^{(y)},\qquad
 \delta_t\ge0,\quad \delta_t=0\ (t\notin C).
 \label{eq:critical-model}
\end{equation}
Within each class the noises are independent and centered, have equal
variances across classes,
$\operatorname{Var}(\varepsilon_t^{(0)})=
\operatorname{Var}(\varepsilon_t^{(1)})=\sigma_t^2$, and satisfy
$|\varepsilon_t^{(y)}|\le a_t$.  Assume separated support,
\begin{equation}
 \min_{t\in C}(\mu_t-a_t)>
 \max_{t\notin C}(\mu_t+a_t),
 \label{eq:critical-support}
\end{equation}
with the condition vacuous when $C$ is the whole span.  Set
$D_C=\sum_{t\in C}\delta_t>0$,
$V_C=\sum_{t\in C}\sigma_t^2>0$, and
$V_B=\sum_{t\notin C}\sigma_t^2$.
For standardized pairwise separation
\[
 d(S)=\frac{\mathbb ES_W-\mathbb ES_R}
 {\sqrt{\operatorname{Var}(S_W)+\operatorname{Var}(S_R)}},
\]
the exact separation formulas are
\begin{equation}
 d(T_k)=\frac{D_C}{\sqrt{2V_C}},\qquad
 d(\bar\ell)=\frac{D_C}{\sqrt{2(V_C+V_B)}}.
 \label{eq:tail-mean-snr}
\end{equation}

\begin{corollary}[Separation gain comes from removing background variance]
\label{cor:tail-gain-main}
Under \eqref{eq:critical-model}--\eqref{eq:critical-support},
the ratio of tail to mean separation is
\begin{equation}
 \frac{d(T_k)}{d(\bar\ell)}=\sqrt{1+\frac{V_B}{V_C}}.
 \label{eq:tail-gain-main}
\end{equation}
With equal per-token noise variances, the ratio is $\sqrt{L/k}$.
If the background is deterministic and common to both classes, the mean
and tail scores are positive affine transforms and give identical rankings.
\end{corollary}

\begin{proof}[Sparse-signal separation and Corollary~\ref{cor:tail-gain-main}]
Condition \eqref{eq:critical-support} and nonnegative shifts imply that
every critical position exceeds every background position almost surely
in both classes.  Thus the top-$k$ set is exactly $C$ before and after
the class shift, and
\[
 T_k(\boldsymbol\ell^{(y)})
 =\frac1k\sum_{t\in C}\mu_t+
   y\frac{D_C}{k}+\frac1k\sum_{t\in C}\varepsilon_t^{(y)}.
\]
Independence and centering give class mean difference $D_C/k$ and
within-class variance $V_C/k^2$.  For the full mean the corresponding
quantities are $D_C/L$ and $(V_C+V_B)/L^2$.  Substitution into the
definition of $d$ proves \eqref{eq:tail-mean-snr}.  With common positive
variance $\sigma^2$, $V_C=k\sigma^2$ and
$V_B=(L-k)\sigma^2$, giving ratio $\sqrt{L/k}$.

If $V_B=0$, each background noise has zero variance and mean zero,
hence is zero almost surely.  Consequently
\[
 \bar\ell=\frac{k}{L}T_k(\boldsymbol\ell)
             +\frac1L\sum_{t\notin C}\mu_t
\]
in both classes.  The positive affine map preserves all comparisons
and threshold acceptance sets, establishing the stronger equal-ranking
claim.  If $V_C=0$ and $D_C>0$, tail scores are deterministic and perfectly
separated; the finite-SNR display omits this case only to avoid division
by zero.
\end{proof}

\paragraph{Localized failures.}
If only $m\le k$ critical positions have a common shift $\Delta$, then
$D_C=m\Delta$, the tail's class mean gap is $m\Delta/k$, and the
whole-answer mean gap is $m\Delta/L$.  These are exact within the model.
Calling a position ``critical'' does not assert that a language model's
mathematical correctness follows an observed independent AND gate; that
causal interpretation would require additional measurements.

\paragraph{What selection stability supplies.}
The top-$k$ variance is exactly a sum of variances divided by $k^2$
because the support separation fixes its membership.  When harmless
lexical spikes can enter or leave the selected set, the functional is
nonlinear and that formula need not hold.  Its deterministic sensitivity
still holds: perturbing at most $m$ coordinates upward by at most $u$
changes $T_k$ by at most $\min(m/k,1)u$.  To prove this, use
$\sum_{t\in I}w_t\le\min(m/k,1)$ for every feasible dual weight vector.
This limits spike influence relative to a maximum without implying a
distribution-free variance decay rate.

\begin{corollary}[An explicit wrong-budget operating point]
\label{cor:bounded-operating-point}
Under \eqref{eq:critical-model}--\eqref{eq:critical-support}, use
$J=C$, $n_J=k$ for the tail, or $J=\{1,\ldots,L\}$, $n_J=L$ for the
mean.  Let $m_{R,J}=n_J^{-1}\sum_{t\in J}\mu_t$,
$m_{W,J}=m_{R,J}+D_C/n_J$, and $A_J=\sum_{t\in J}a_t^2>0$.
Let $\pi_R$ and $\pi_W$ be the unfiltered marginal probabilities of
right and wrong submissions. For $0<\alpha_0<\pi_W$, define
\begin{equation}
 \tau_J=m_{W,J}
       -\frac{\sqrt{2A_J\log(\pi_W/\alpha_0)}}{n_J}.
 \label{eq:oracle-bounded-threshold}
\end{equation}
Accepting valid submissions with score below $\tau_J$ gives
$W(\tau_J)\le\alpha_0$ and
\begin{equation}
 R(\tau_J)\ge\pi_R\left[1-\exp\left\{-\frac12
 \left(\frac{D_C}{\sqrt{A_J}}-
 \sqrt{2\log(\pi_W/\alpha_0)}\right)_+^2\right\}\right].
 \label{eq:tail-operational-bound}
\end{equation}
\end{corollary}

\begin{proof}
For independent, centered variables $\varepsilon_t\in[-a_t,a_t]$,
Hoeffding's inequality gives, for $u>0$,
\begin{equation}
 \Pr\left(\sum_{t\in J}\varepsilon_t\le-u\right)
 \le e^{-u^2/(2A_J)},\qquad
 \Pr\left(\sum_{t\in J}\varepsilon_t\ge u\right)
 \le e^{-u^2/(2A_J)}.
 \label{eq:hoeffding-score}
\end{equation}
For completeness, convexity of the exponential bounds its value at
$x\in[-a_t,a_t]$ by the chord joining its endpoint values.  Taking
expectations and using $\mathbb E x=0$ gives
$\mathbb E e^{sx}\le\cosh(sa_t)\le e^{s^2a_t^2/2}$.
The second inequality follows by integrating
$\tanh u\le u$ for $u\ge0$ and using symmetry for $u<0$.
Multiplying these bounds and minimizing
$\exp(-su+s^2A_J/2)$ over $s>0$ gives the two inequalities.

For wrong draws, $n_J(S_W-m_{W,J})=\sum_{t\in J}
\varepsilon_t^{(1)}$.  Substitution of
$u=\sqrt{2A_J\log(\pi_W/\alpha_0)}$ into
\eqref{eq:hoeffding-score} yields
\[
 W(\tau_J)=\pi_W\Pr(S_W<\tau_J\mid W)\le\alpha_0.
\]
For right draws,
\[
 n_J(\tau_J-m_{R,J})
 =D_C-\sqrt{2A_J\log(\pi_W/\alpha_0)}.
\]
If this number is positive, the upper-tail Hoeffding inequality bounds
the probability that a right score is at least $\tau_J$.  Subtracting
that bound from one and multiplying by $\pi_R$ proves
\eqref{eq:tail-operational-bound}.  If it is nonpositive, the displayed
positive-part formula reduces to the universally valid bound $R\ge0$.
If $A_J=0$, the scores are deterministic and a threshold strictly between
the two class means instead gives $W=0$ and $R=\pi_R$.
\end{proof}

The bound depends on $D_C/\sqrt{A_J}$, not mean dilution alone.
Discarding background positions improves this sufficient guarantee by
removing nuisance concentration terms while preserving the total class
signal.  This population threshold uses model parameters; deployment
estimates its threshold by calibration.  The next result connects these
distinct constructions.

\begin{corollary}[Finite-calibration slack]
\label{cor:score-calibration-slack}
Let an i.i.d. calibration sample have size $n$, and let the CRC rule
choose the largest threshold satisfying
$(1+\sum_{i=1}^n\mathbf1\{Y_i=W,S_i<\tau\})/(n+1)\le\alpha$,
with the usual strict-tie convention.  Fix a calibration slack satisfying
\[
 \frac{1-\alpha}{n}<\eta<\alpha,
\]
and a data-independent threshold $\tau_0$ with population
$W(\tau_0)\le\alpha-\eta$.  Then
\begin{equation}
 \Pr(\widehat\tau_{\rm CRC}\ge\tau_0)
 \ge 1-\exp\left[-2n\left(\eta-\frac{1-\alpha}{n}\right)^2\right].
 \label{eq:calibration-slack}
\end{equation}
On this event $R(\widehat\tau_{\rm CRC})\ge R(\tau_0)$.
In particular, choosing $\tau_0$ from
Corollary~\ref{cor:bounded-operating-point} with
$\alpha_0=\alpha-\eta<\pi_W$ transfers its correct-submission lower bound
to the CRC policy on the displayed high-probability event.
\end{corollary}

\begin{proof}
The calibration wrong-submission indicators at $\tau_0$ are i.i.d.
Bernoulli with mean at most $\alpha-\eta$.  The threshold is feasible
whenever their average is at most
$\alpha-(1-\alpha)/n$.  Hoeffding's inequality therefore bounds the
probability of infeasibility by the exponential in
\eqref{eq:calibration-slack}.  Whenever $\tau_0$ is feasible, the
maximal feasible CRC threshold is at least $\tau_0$, including the
strict-tie implementation.  Acceptance is monotone in the threshold,
so its correct-submission probability is at least that at $\tau_0$.
The efficiency bound assumes an explicit population margin and i.i.d.
calibration; the marginal CRC guarantee is separate.
\end{proof}

\paragraph{An optional exact ROC interpretation.}
In the separate score-level model
$S_R\sim N(\mu_R,\sigma^2)$ and
$S_W\sim N(\mu_W,\sigma^2)$, put
$d_0=(\mu_W-\mu_R)/\sigma$.  At $0<\alpha<\pi_W$, the threshold
$\tau=\mu_W+\sigma\Phi^{-1}(\alpha/\pi_W)$ has
\[
 W=\alpha,\qquad
 R=\pi_R\Phi\left(\Phi^{-1}(\alpha/\pi_W)+d_0\right),\qquad
 \operatorname{AUC}=\Phi(d_0/\sqrt2).
\]
These identities follow from the two normal CDFs and the normal law of
$S_W-S_R$.  They show how stronger standardized separation can retain more
correct submissions at a fixed wrong budget.  This Gaussian score model
differs from the bounded, separated-support token model: nondegenerate
Gaussian token noise is unbounded and cannot satisfy almost-sure support
separation.

\paragraph{A maximum can be dominated by nuisance extremes.}
This is a separate illustrative model, not an assumption of the preceding
sparse-signal analysis.  Suppose all class-dependent critical scores are bounded
above by $K$, and there are $m$ independent class-independent nuisance
scores with common CDF $F$, where $F(K)<1$.  Couple the maximum score
in each class to its nuisance-only maximum.  They differ only if every
nuisance score is at most $K$, an event of probability $F(K)^m$.
The independent nuisance-only maxima have tie-corrected AUROC $1/2$
by symmetry.  A union bound over the two class draws yields
\[
 \left|\operatorname{AUC}(\max)-\tfrac12\right|\le2F(K)^m.
\]
The conclusion follows because the pairwise AUROC contribution lies in
$[0,1]$ and is unchanged when both coupled scores agree.  This provides
one precise way that a maximum can lose signal.  It does not establish
that every upper-tail average avoids nuisance contamination.  Under a
specific tail model, a growing expected maximum alone implies neither
growing variance nor declining AUROC.

\subsubsection{Decoder-aware martingales and stopping}
\label{app:surprise-proof}
\label{sec:entropy}

Token log probabilities are available with many generation APIs, whereas
full-vocabulary entropy requires access to the model distribution at each
prefix. The additional access is valuable only if it changes the information
used to rank answers. To examine that question, we separate the uncertainty
of the model distribution from the random surprise of the token actually
sampled, accounting for the decoder transformation used by the experiments.
This distinction is essential because top-$p$/top-$k$ sampling generally
changes the distribution from which a token is drawn.

\begin{proposition}[Decoder-aware surprise and a sharp tail comparison]
\label{prop:decoder-decomposition}
At generation history $\mathcal F_{t-1}$, let $p_t$ be the scored token
distribution and $q_t$ the actual sampling distribution on a finite
vocabulary, with $q_t\ll p_t$ and finite surprise second moment. For
$Y_t\sim q_t$ conditionally,
\begin{align}
 \ell_t&=-\log p_t(Y_t)=H(p_t)+b_t+\xi_t,\nonumber\\
 c_t&=H(q_t)+\KL(q_t\Vert p_t),\qquad b_t=c_t-H(p_t),\nonumber\\
 \E[\xi_t\mid\mathcal F_{t-1}]&=0,\qquad
 \E[\xi_t^2\mid\mathcal F_{t-1}]
   =v_t:=\Var_{q_t}[-\log p_t(Y_t)].
 \label{eq:general-surprise-mds}
\end{align}
For a deterministic horizon $N$, the mean residual
$N^{-1}\sum_{t=1}^N\xi_t$ has second moment
$N^{-2}\sum_{t=1}^N\E v_t$. When $q_t=p_t$, $b_t=0$ and $v_t$ is
token varentropy. For every realized scored span, write
$r_t=\ell_t-H(p_t)$ and $\bar r=L^{-1}\sum_tr_t$. Then
\begin{equation}
 \left|T_k(\boldsymbol\ell)-T_k(\mathbf H)-\bar r\right|
 \le\sqrt{\frac1k-\frac1L}\,
       \|\mathbf r-\bar r\mathbf1\|_2,
 \label{eq:pathwise-tail-discrepancy}
\end{equation}
with a sharp constant, including for random-length or validity-selected spans.
\end{proposition}
\emph{Proof.} The decomposition and isometry are proved below;
Appendix~\ref{app:pathwise-proof} proves the pathwise bound.

The decomposition identifies three contributions to sampled-token surprise:
full-distribution entropy $H(p_t)$, decoder drift $b_t$, and centered random
surprise $\xi_t$. At temperature one, truncation to a retained set with
probability mass $Z_t$ gives $c_t=H(q_t)-\log Z_t$.
Full-vocabulary entropy removes sampled-token fluctuation but also changes
the decoder-dependent center, making its effect on answer ranking an
empirical question.

A simple example distinguishes entropy from sampling noise. A uniform
model distribution on $K$ tokens has entropy $\log K$, yet every token has
the same surprise and varentropy is zero~\citep{kontoyiannis2014optimal}.
High entropy alone therefore does not predict an entropy upgrade. In a
fixed-context resampling model, convexity gives
$\E T_k(\boldsymbol\ell)-T_k(\mathbf c)\ge0$, with an upper bound in
terms of $v_t$; matched sampling specializes the center to entropy.
Lemma~\ref{lem:auc-margin} translates score perturbations into ranking
changes through right--wrong pair margins.

For a deterministic horizon $N$, the precise isometry is
\begin{equation}
 \mathbb E\left(\frac1N\sum_{t=1}^N\xi_t\right)^2
 =\frac1{N^2}\sum_{t=1}^N\mathbb E v_t.
 \label{eq:mds-isometry}
\end{equation}

\begin{proof}[Decomposition and isometry in Proposition~\ref{prop:decoder-decomposition}]
Conditional on $\mathcal F_{t-1}$, both distributions are fixed and
$Y_t$ is distributed according to $q_t$.  Therefore
\[
 \mathbb E[\ell_t\mid\mathcal F_{t-1}]
   =\sum_vq_t(v)[-\log p_t(v)]=c_t.
\]
Centering proves both conditional-moment statements.
Adding and subtracting $H(p_t)$ gives
\eqref{eq:general-surprise-mds}.  The cross-entropy identity follows from
$-\log p_t(v)=-\log q_t(v)+\log(q_t(v)/p_t(v))$
on the support of $q_t$.

For $s<t$, $\xi_s$ is $\mathcal F_{t-1}$-measurable, so integrability
and the tower property imply
\[
 \mathbb E[\xi_s\xi_t]
 =\mathbb E[\xi_s\mathbb E(\xi_t\mid\mathcal F_{t-1})]=0.
\]
Thus expanding the square of the deterministic-horizon sum eliminates
all cross terms, and
\[
 \mathbb E\left(\sum_{t=1}^N\xi_t\right)^2
 =\sum_{t=1}^N\mathbb E\xi_t^2
 =\sum_{t=1}^N\mathbb E v_t.
\]
Division by $N^2$ proves \eqref{eq:mds-isometry}.  The sum has mean
zero, so this second moment is its variance.  The same identities hold
conditionally on the initial sigma-field $\mathcal F_0$ after replacing
each outer expectation by its conditional version.  Conditioning on the
entire random future history need not preserve orthogonality.

When $q_t=p_t$, $c_t=H(p_t)$ and its conditional variance is
$V(p_t)=\sum_vp_t(v)[-\log p_t(v)-H(p_t)]^2$.
For truncation, on the retained support
$-\log p_t(v)=-\log q_t(v)-\log Z_t$.  Taking its $q_t$-mean proves
$c_t=H(q_t)-\log Z_t$; subtracting its mean cancels the constant
$-\log Z_t$, proving $v_t=V(q_t)$.
\end{proof}

For a retained set consisting of the most probable tokens, every retained
surprise is no greater than every omitted surprise.  If $0<Z_t<1$
and $c_t^{\rm out}$ is the mean surprise conditional on omission,
\[
 b_t=c_t-H(p_t)
     =(1-Z_t)(c_t-c_t^{\rm out})\le0.
\]
Its magnitude can vary with context.  A small measured discrepancy
between receipt log probabilities and recomputed raw log probabilities
does not set $Z_t=1$ or remove this drift: it compares scoring values,
not the two sampling laws.  If a receipt log probability has an
additional numerical or backend residual $e_t$ relative to
$-\log p_t(Y_t)$, the exact pathwise residual becomes
$r_t=b_t+\xi_t+e_t$.

\begin{lemma}[Predictable weighting and bounded stopping]
\label{lem:predictable-isometry}
If $a_t$ is $\mathcal F_{t-1}$-measurable and the terms are square
integrable, the weighted sum has zero mean and satisfies the isometry
\[
 \mathbb E\sum_{t=1}^N a_t\xi_t=0,\qquad
 \mathbb E\left(\sum_{t=1}^N a_t\xi_t\right)^2
       =\mathbb E\sum_{t=1}^N a_t^2v_t.
\]
In particular, for a stopping time $\tau\le N$,
\[
 \mathbb E\sum_{t=1}^{\tau}\xi_t=0,\qquad
 \mathbb E\left(\sum_{t=1}^{\tau}\xi_t\right)^2
       =\mathbb E\sum_{t=1}^{\tau}v_t.
\]
\end{lemma}

\begin{proof}
The weighted increments have conditional mean zero and conditional
second moment $a_t^2v_t$.  The preceding orthogonality proof applies
to them unchanged.  For bounded stopping, use
$a_t=\mathbf1\{\tau\ge t\}$.  This is predictable because
$\{\tau\ge t\}$ is the complement of
$\{\tau\le t-1\}\in\mathcal F_{t-1}$.
\end{proof}

If $\tau$ is the first EOS time, these statements include its sampled
EOS token.  The indicator for retaining a token strictly before EOS is
$\mathbf1\{t<\tau\}$, which depends on the current token and need not
be predictable.  Deleting EOS therefore changes the compensator.  One
can center a masked contribution directly, for example
$g_t(Y_t)=\mathbf1\{Y_t\ne\mathrm{EOS}\}[-\log p_t(Y_t)]$,
by $\sum_vq_t(v)g_t(v)$, but that center is not Shannon entropy.
Selecting only eventually valid trajectories is another conditioning
operation depending on future tokens.  Neither operation preserves the
unconditioned martingale statements automatically.

\paragraph{Random normalization can introduce bias.}
An example uses genuine self-information.  Draw two independent tokens
from $p=(2/3,1/3)$, and let
$a=(\log2)/3$.  Their centered surprises independently equal $-a$
with probability $2/3$ and $2a$ with probability $1/3$.  Stop at
$\tau=1$ if the first centered surprise is $2a$, and at $\tau=2$
otherwise.  This is a bounded stopping time, and the unnormalized
stopped sum has mean zero.  Normalizing by the stopping time instead gives
a nonzero expectation:
\[
 \mathbb E\left[\frac1\tau\sum_{t=1}^{\tau}\xi_t\right]
 =\frac a3\ne0.
\]
Indeed, the normalized sum takes values $2a,a/2,-a$ with probabilities
$1/3,2/9,4/9$.  Its second moment is $11a^2/6$, whereas naively
placing the random denominator in the deterministic-horizon formula
would give
\[
 \mathbb E\left[\frac1{\tau^2}\sum_{t=1}^{\tau}v_t\right]
 =\mathbb E\left[\frac{2a^2}{\tau}\right]=\frac{4a^2}{3}.
\]
The discrepancy is not a minor asymptotic technicality.  It excludes
an exact unbiasedness or variance claim for the random-length answer
mean without additional conditions.

\subsubsection{A pathwise bound that survives span selection}
\label{app:pathwise-proof}

The pathwise bound is tailored to the implemented answer score. A fixed
horizon gives a martingale second-moment identity, but the scored span ends
at a random time, excludes EOS, and is evaluated among valid outputs.
Equation~\eqref{eq:pathwise-tail-discrepancy} continues to hold for each such
realized span. It shows that the discrepancy between tail surprise and tail
entropy depends on the mean residual and its variation across positions:
uniform shifts and localized spikes affect the tail differently.

\begin{proof}[Pathwise bound in Proposition~\ref{prop:decoder-decomposition}]
Let $\mathcal W_k=\{w:0\le w_t\le1/k,\sum_tw_t=1\}$,
$u=L^{-1}\mathbf1$, and $z^\circ=z-\bar z\mathbf1$.
For any $w\in\mathcal W_k$, its coordinate bounds and unit sum imply
\[
 \|w-u\|_2^2=\sum_t w_t^2-\frac1L
 \le\frac1k\sum_t w_t-\frac1L
 =\frac1k-\frac1L.
\]
If $w_+$ maximizes $T_k(x+z)$, then
$T_k(x+z)-T_k(x)\le w_+^\top z$.
If $w_0$ maximizes $T_k(x)$, the analogous lower bound is
$T_k(x+z)-T_k(x)\ge w_0^\top z$.
Since
\[
 w^\top z=\bar z+(w-u)^\top z^\circ,
\]
Cauchy--Schwarz bounds deviations from $\bar z$ by
$\sqrt{1/k-1/L}\|z^\circ\|_2$.  Set $x=\mathbf H$,
$z=\boldsymbol\ell-\mathbf H$ to obtain
\eqref{eq:pathwise-tail-discrepancy}.

For sharpness when $k<L$, choose a fixed $k$-subset $J$,
$w=\mathbf1_J/k$, $x=0$, and $z=w-u$.  Then $\bar z=0$,
its top-$k$ subset is $J$, and
\[
 T_k(z)=w^\top(w-u)=\frac1k-\frac1L,
 \qquad \|z\|_2=\sqrt{\frac1k-\frac1L}.
\]
Equality holds.  When $k=L$, $T_k$ is the mean and the bound is the
identity with zero remainder.  No part of the proof used a distribution
for $x,z,L$, so it remains true after any span or validity selection.
\end{proof}

A weaker but sometimes convenient consequence of the dual representation
is
\[
 |T_k(x+z)-T_k(x)|
 \le\min\{\|z\|_\infty,\ \|z\|_1/k,\ \|z\|_2/\sqrt{k}\}.
\]
The centered bound distinguishes an overall score displacement from
heterogeneous fluctuations that can alter which tokens enter the tail.

\subsubsection{Fixed-context convexity, equality, and autoregressive limits}
\label{app:jensen-proof}

In a separate fixed-context perturbation model, hold $p_{1:L},q_{1:L}$
fixed exogenously and draw independent tokens from the $q_t$.  Then
\begin{equation}
 0\le\mathbb E T_k(\boldsymbol\ell)-T_k(\mathbf c)
 \le
 \sqrt{\left(\frac1k-\frac1L\right)
             \left(1-\frac1L\right)\sum_{t=1}^L v_t}.
 \label{eq:fixed-context-jensen}
\end{equation}
Equality on the left holds exactly when some fixed $k$-element subset is
a maximizing subset almost surely.  Zero varentropy is sufficient in the
matched-law case, but is not necessary.

\begin{proof}[Fixed-context extension of Proposition~\ref{prop:decoder-decomposition}]
Hold the distributions fixed exogenously.  The deterministic mean vector
is $\mathbf c$ and $\boldsymbol\ell=\mathbf c+\boldsymbol\xi$, with
$\mathbb E\boldsymbol\xi=0$.  Equation~\eqref{eq:tail-primal-dual}
expresses $T_k$ as a maximum of linear functions, so it is convex.
Jensen's inequality gives
$\mathbb ET_k(\boldsymbol\ell)\ge T_k(\mathbf c)$.
The pathwise upper bound gives
\[
 T_k(\mathbf c+\boldsymbol\xi)-T_k(\mathbf c)
 \le\bar\xi+
       \sqrt{1/k-1/L}\|\boldsymbol\xi-\bar\xi\mathbf1\|_2.
\]
Its expectation is at most
\[
 \sqrt{1/k-1/L}
 \sqrt{\mathbb E\|\boldsymbol\xi-\bar\xi\mathbf1\|_2^2}.
\]
Independence and centering imply
$\mathbb E\bar\xi^2=L^{-2}\sum_tv_t$.  Hence
\[
 \mathbb E\|\boldsymbol\xi-\bar\xi\mathbf1\|_2^2
 =\sum_tv_t-L\mathbb E\bar\xi^2
 =(1-1/L)\sum_tv_t,
\]
proving \eqref{eq:fixed-context-jensen}.  Jensen itself does not require
independence.  For dependent perturbations with covariance matrix
$\Sigma$, the same proof replaces the final factor by
$\operatorname{tr}\Sigma-L^{-1}\mathbf1^\top\Sigma\mathbf1$.

For equality, choose a deterministic $k$-subset $J$ maximizing the
mean vector $\mathbf c$.  Then
\[
 T_k(\boldsymbol\ell)-\frac1k\sum_{t\in J}\ell_t\ge0
 \quad\text{almost surely},
\]
and its expectation equals
$\mathbb ET_k(\boldsymbol\ell)-T_k(\mathbf c)$.
That expectation is zero exactly when the nonnegative random variable
is zero almost surely, i.e. when $J$ is maximizing almost surely.
Conversely, if a fixed $J$ is maximizing almost surely, its expected
value is at least that of every other subset, so it also maximizes
the mean vector and the gap is zero.  This proves the stated necessary
and sufficient condition, including ties and $k=L$.
\end{proof}

\paragraph{Positive varentropy with zero Jensen gap.}
For $L=2,k=1$, take independent matched-law tokens from
$p_1=(2/3,1/3)$ and $p_2=(0.3,0.3,0.2,0.2)$.  Both distributions
have positive varentropy.  Every first-coordinate surprise is at most
$\log3$, while every second-coordinate surprise is at least
$\log(10/3)>\log3$.  The second coordinate is therefore always the
maximum, and
\[
 \mathbb E\max(\ell_1,\ell_2)=H(p_2)
                         =\max(H(p_1),H(p_2)).
\]
Repeating the low-support distribution at nine positions and using the
high-support distribution at one gives the same counterexample for
the literal $L=10,k=1$ tail10 score.  Zero varentropy is therefore not
necessary, even if every coordinate has positive varentropy.  For a
finite distribution $p$, $V(p)=0$ itself is equivalent to uniformity on
its positive-probability support: zero variance forces $-\log p(Y)$
to be constant, so all positive probabilities coincide; the converse
is immediate.

\paragraph{The fixed-context bias need not disappear with length.}
For example, let all coordinates be independent draws from
$p=(0.9,0.1)$ and let $k=\lceil0.1L\rceil$.  The two surprise values
are $a=-\log0.9$ and $b=-\log0.1$.  If $N_L$ is the number of rare
draws, then
\[
 T_k(\boldsymbol\ell)
 =a+(b-a)\min(N_L/k,1).
\]
The law of large numbers gives $N_L/L\to0.1$ and $k/L\to0.1$,
so $T_k(\boldsymbol\ell)\to b$ almost surely.  Bounded convergence then
transfers this almost-sure limit to the expected score:
\[
 \mathbb ET_k(\boldsymbol\ell)-T_k(\mathbf H)
 \longrightarrow b-H(p)>0.
\]
Thus a finite-sample upper-tail selection bias can remain of constant
order at a fixed tail fraction, even with independent coordinates.
The bound in \eqref{eq:fixed-context-jensen} is consistent:
for equal $v_t=v$ and fixed $k/L$, it is of order $\sqrt v$, not
$L^{-1/2}$.  At fixed $L$, the bound vanishes with the coordinate
varentropies.

\paragraph{Autoregression can reverse the Jensen comparison.}
Consider a deterministic horizon $L=10$, $k=1$, with matched sampling
law $q_t=p_t$.  At the first step, $p_1=(0.9,0.1)$.  If the common
token is drawn, let step two be deterministic.  If the rare token is
drawn, let step two be uniform on ten tokens.  Steps three through ten
are deterministic.  On the common-token branch,
$T_1(\boldsymbol\ell)=-\log0.9$ while $T_1(\mathbf H)=H(p_1)$.
On the rare-token branch, both equal $\log10$.  Consequently
\[
 \mathbb ET_1(\boldsymbol\ell)-\mathbb ET_1(\mathbf H)
 =0.9[-\log0.9-H(p_1)]<0.
\]
Every per-step conditional-mean identity is valid, but the sequence of
later distributions reveals the first draw.  Conditioning on that
sequence is therefore not the independent resampling experiment used
in \eqref{eq:fixed-context-jensen}.  This also shows why
$T_k(\mathbf H)$ should not be described as the exact Rao--Blackwell
conditional expectation of the entire sampled tail score.

\subsubsection{From score discrepancies to ranking discrepancies}
\label{app:margin-proof}

\begin{lemma}[A score-margin condition for AUROC proximity]
\label{lem:auc-margin}
Let two scores $S,T$ be coupled on the same trajectory.  On independent
right and wrong draws write $e_y=S_y-T_y$.  Then
\begin{equation}
 |\operatorname{AUC}(S)-\operatorname{AUC}(T)|
 \le \Pr\bigl(|T_W-T_R|\le |e_W|+|e_R|\bigr).
 \label{eq:auc-margin-bound}
\end{equation}
In particular, for any $\epsilon_R,\epsilon_W\ge0$, the right side is
at most
\[
 \Pr\bigl(|T_W-T_R|\le\epsilon_R+\epsilon_W\bigr)
 +\Pr(|e_R|>\epsilon_R)+\Pr(|e_W|>\epsilon_W).
\]
\end{lemma}

\begin{proof}
Outside the event on the right of \eqref{eq:auc-margin-bound}, the
perturbation $e_W-e_R$ is strictly smaller in absolute value than
$T_W-T_R$, so it cannot change its sign or create a tie.  Thus the
two pairwise $\psi$ contributions agree there.  On the event they can
differ by at most one.  Taking expectations proves the first bound;
splitting according to the two error events proves the second.
\end{proof}

The lemma supplies a sufficient route from a small pathwise discrepancy
to similar AUROC: pairwise entropy-score margins must also be large
relative to that discrepancy.  The converse fails.  Large changes that
preserve rankings can leave AUROC unchanged, and equal aggregate AUROCs
can arise from different pairs changing in opposite directions.  Hence
neither an equal AUROC nor a confidence interval crossing zero supports
an estimate of low varentropy.  The empirical comparison and the
conditional mathematical mechanisms must remain separate claims.

\clearpage
\section{Experimental protocols and reproducibility}
\label{app:protocols}

\subsection{Main-text names and archived records}

The table maps public method names to archived identifiers and training
configurations.
\begin{center}\small
\begin{tabular}{@{}L{.26\linewidth}L{.67\linewidth}@{}}
\toprule
Main-text name & Archived record and interpretation \\
\midrule
Native & \texttt{Native/base}: original pretrained weights with native decoding. \\
Budget forcing & \texttt{ExtBF/base}: original pretrained weights with external closure. \\
Answer-SFT & 4B \texttt{E1\_41}; 8B \texttt{M5\_41}, both decoded with ExtBF. \\
Binary RL & \texttt{G0\_41}, \texttt{G0\_73}; pooled G0 averages the two seeds within family. \\
Penalty RL & \texttt{G1\_41}, with wrong-answer reward $-1$. \\
Mild-Penalty RL & \texttt{S3B}, with wrong-answer reward $-0.25$. \\
Explicit-Abstain RL & \texttt{S3A}, with reward $-1$ for wrong answers and an added abstention instruction/action. \\
Tail-CRC & Binary RL with \texttt{tail10\_nll} and per-seed CRC filtering. \\
\bottomrule
\end{tabular}
\end{center}
Tail NLL, Box NLL, Mean NLL, Sum NLL, and Tail entropy correspond to
\texttt{tail10\_nll}, \texttt{box\_sum\_nll}, \texttt{ans\_mean\_nll},
\texttt{ans\_sum\_nll}, and \texttt{tail\_H}, respectively.
Source-calibrated and target-calibrated Omni policies are Track A
(DAPO cuts) and Track B (Omni CAL300 cuts). The source-domain study is
\texttt{pc\_crc\_confirm\_2a\_v1}; independent target-domain confirmation is
\texttt{pc\_crc\_ood\_confirm\_v1}. ``Binary RL with native decoding''
denotes \texttt{Native/G0\_41}, distinct from the original Native baseline.

\subsection{Exact decoding and scoring conventions}

For Qwen3.5-4B, OPEN is $(248068,198)$, CLOSE is
$(248069,198)$, and chat EOS is 248046. For Qwen3-8B they are
$(151667,198)$, $(151668,198)$, and 151645. OPEN occurs once in the
assistant prefix. The nominal cap includes this prefill and any two-token external CLOSE. At 4K, the two sampled segment caps are 2,046
and 2,046; at 8K they are 4,094 and 4,094. A naturally generated CLOSE
does not trigger another injected delimiter. An early terminal ends the
request, and the second call continues the existing prefix if needed. There are no custom stop strings. Sampling seeds are deterministic
per family, model/budget, and segment, with common seeds across compared
arms within the same decoder. Native and ExtBF use different segment
streams and are paired by family, not by identical sampled trajectories.

The verifier first checks whether a completed, extractable mathematical
answer was submitted. Normal chat EOS and mutually equivalent extracted
boxes are required for a tolerant valid submission. Truncation, conflicting
boxes, or other invalid structure yields $U$. The tolerant correctness
check uses \texttt{math-verify} 0.9.0. Strict-format correctness is a
separate diagnostic, which can mark repeated equivalent boxes invalid
even when their value is correct. S3A identifies explicit abstention only
after the first CLOSE and normal EOS, when all valid boxed contents match
\texttt{ABSTAIN} after whitespace and \texttt{text} wrapper normalization,
without case sensitivity. Such cases are $U_{\mathrm{abstain}}$ and, like truncation, receive zero reward.

The scored answer span consists only of sampled tokens after the first
CLOSE and before EOS. Injected delimiters and injected newline tokens,
and EOS itself, are excluded. For length $L$, let $k=\lceil L/10\rceil$;
\texttt{tail10\_nll} is the mean of the largest $k$ sampled-token NLLs.
The empirical CVaR interpretation uses effective tail fraction $k/L$.
\texttt{ans\_mean\_nll}, \texttt{ans\_sum\_nll}, and \texttt{ans\_len}
use the same answer span. \texttt{box\_sum\_nll} includes complete generated
tokens whose byte spans overlap the \emph{content} of the final valid box,
rather than indiscriminately including every box delimiter. Invalid
submissions are always excluded from acceptance. CRC uses strict
$S<\tau$ and rejects the entire group tied at the cutoff; calibration is
separate for each model, seed, budget, score, and target. The allowance is
$\lfloor(n+1)\alpha-1\rfloor$ calibration wrong submissions. An all-reject
fallback and the all-valid-accept case are retained explicitly.

Full-vocabulary A2 scores are computed by replaying the original token
sequence, taking the previous-position hidden state and output projection,
and applying FP32 softmax to BF16 model output. Tail entropy averages the
largest $k$ entropies. The margin control averages the smallest $k$
top-one minus top-two probability margins, then reverses sign for the
wrong-positive AUROC. Theoretical interpretation must distinguish raw-model probabilities
from the top-$p$/top-$k$ sampling distribution.

\subsection{Training, masks, and baseline provenance}

The mainline Dr.~GRPO optimizer uses AdamW with learning rate $10^{-5}$,
$\beta_1=.9$, $\beta_2=.95$, $\epsilon=10^{-8}$, zero weight decay, and
gradient-norm clipping at 1. Rank and LoRA alpha are both 32. Each of the
50 updates uses 16 prompts with eight on-policy rollouts each. The 400-family
pool is traversed twice in a seed-specific deterministic order. G0's reward
is the tolerant-valid mathematical $R$ indicator, not a strict-format
indicator. G1/S3B change the wrong-answer reward to $-1/-0.25$; no penalty
is added to invalid or truncated trajectories. The policy-gradient loss
includes policy-sampled positions in both stages and excludes prompt,
prefilled OPEN, and external CLOSE positions. Masked-label perturbation
tests keep the input context fixed. In both supplementary-arm smoke tests,
changing the masked CLOSE labels changes the loss by exactly zero; changing
the actual input context does change it. Formal training restarts from the
unchanged zero-update state after these disposable one-step tests.

The historical SFT diagnostic and the current SFT control are different
experiments. In the earlier 16K study, M2 versus M1 was evaluated on 1,000
DAPO families with two training seeds. The strict-format improvement was
10.30 pp $[8.05,12.60]$; the mathematical-equivalence improvement was
1.10 pp $[-.45,2.65]$, and their paired difference was 9.20 pp
$[7.25,11.20]$. Its primary gate failed. The current 8B M5$_{41}$ control uses answer-only SFT on a
262-family rescue-inclusive pool, selecting each family's shortest
qualifying strict-correct teacher trajectory.
Its own preregistration specifies batch size eight and 160 updates. The
4B E1 control is the archived answer-focused SFT policy. Neither is the earlier M2--M1 intervention.

\subsection{Data selection, labels, and blinding}

DAPO families use normalized text, numeric templates, and character
3--5-gram similarity at cosine threshold .90.
All DAPO TEST references are bare integers. Historical Omni200 was enriched
for non-bare reference answers and is only a diagnostic panel. Fresh Omni confirmation uses the pinned 4,428-row release with
4,391 families. Historical enrollment across 23 Omni panels is excluded
along all three identity axes. The final eligible frame contains 1,084
verifiable remaining families. Deterministic SHA-256 ordering selects
600 without a bare/non-bare quota: CAL has 39 bare and 261 non-bare
references; TEST has 37 and 263. The other 484 are reserved and untouched.
This is absence of overlap with tracked study exposures, not a claim about
pretraining contamination or about unverifiable excluded problems.

The OOD protocol was registered at 2026-11-21 19:28:52 UTC. CAL labels
opened at 2026-11-22 01:00:58 UTC, after all study trajectories had been
generated. Thresholds were locked at 01:01:48.282 UTC; TEST labels opened
at 01:01:48.350 UTC. Track A transfers fixed source-DAPO cuts without a
distribution-shift guarantee; Track B fits on target CAL300. Registered
later at 04:05:42 UTC, the final supplement reuses released panels. Its
additional controls and analyses are not fresh confirmatory endpoints.

\subsection{Study sequence and prespecified outcomes}
\label{app:study-history}

The initial 8B single-seed RL pilot's registered DEV200 verdict was
\texttt{STOP\_RLVR\_SCALE}. A separate retrospective DEV probe nominated
\texttt{box\_sum\_nll} for the source-domain confirmation (2A). Its
primary gates were all at BF4; BF8 and comparisons with M5 were secondary.
At $\alpha=0.065$, 4B pooled G0 passed against raw ExtBF base, with
$\Delta R=26.70$ pp (95\% CI $[23.10,30.40]$ pp) and $W=5.0\%$.
Both seeds accepted every valid answer at this cutoff, so that operating
point's gain comes from solving and completion.
For 8B at the same target, the comparison with raw ExtBF base gave
$\Delta R=3.90$ pp (95\% CI $[0.20,7.60]$ pp) and $W=7.5\%$,
missing the 4 pp effect gate and the risk gate.
At $\alpha=0.10$, the comparison with raw G1 gave
$\Delta R=7.20$ pp (95\% CI $[3.30,11.20]$ pp) and $W=11.5\%$,
passing the effect gate but missing the risk gate. Both 8B joint
gates failed. These original endpoints use Box NLL; Tail NLL was analyzed later.

Historical Omni200 was enriched for non-bare reference answers and served
as a score diagnostic. For 8B pooled G0, Box NLL AUROC was 0.544 at 4K
and 0.510 at 8K. At the transferred 4K cut fitted with $\alpha=0.065$,
all 23 accepted wrong answers were bare-integer predictions, while
82 of 84 rejected correct answers were non-bare (400 seed--family
records from two seeds). Subsequent natural-Omni TEST300 independently
tested Tail NLL; it is a different panel with no answer-shape quota.

The independent Omni study passed H1a (4B raw solving and lower cost at
both caps), H1b (8B raw solving at 4K), and H2a (8B Tail NLL versus
Box NLL AUROC with positive 4B ordering). H2b, the joint 8B Track A
net-score improvement check at both caps, failed: the Tail NLL minus
Box NLL gain was 3.50 pp (95\% CI $[-1.00,8.33]$ pp) at 4K and
5.17 pp (95\% CI $[0.33,10.17]$ pp) at 8K; the 4K interval crossed
zero. H3a/H3b were reporting items for target-domain risk and LTT
issuance. The later fixed-panel supplement adds Native controls, reward
ablations, white-box scoring, and DAPO Tail NLL analyses.

The original 2A family bootstrap uses RNG seed 20261121; the independent
Omni study and final supplement use 20261122, each with 20,000 paired
resamples. White-box AUROC intervals use up to 2,000 estimable paired
resamples. The Native-transfer supplement evaluates G0 seed 41 against
the original Native model: all four 4B correct-submission differences
have positive lower bounds (DAPO 4K/8K: 1.60/7.20 pp; Omni 4K/8K:
2.67/3.33 pp), whereas all four 8B intervals cross zero (point estimates
from $-0.40$ to $+0.60$ pp). No noninferiority margin was registered.
Table~\ref{tab:native-transfer} preserves the full intervals.

\subsection{LTT grids and the scope of guarantees}

The two fixed cutoff grids are
\[
\begin{aligned}
\mathcal{T}_{\mathrm{tail}}&=\{.50,.75,1.00,1.25,1.50,1.75,2.00,2.35,2.80,\infty\},\\
\mathcal{T}_{\mathrm{mean}}&=\{.06,.10,.14,.18,.22,.27,.33,.40,.50,\infty\}.
\end{aligned}
\]
For each fixed model, budget, score, and conditional-error target,
the ten CAL binomial tests each use level .01, giving nominal
$\delta=.10$ across that grid. The largest passing cutoff is selected
(equivalently maximum accepted coverage, with the largest cut breaking
ties); if none passes, no certificate is issued. Empty acceptance sets do
not certify an undefined conditional error. The target grid is
$\{.10,.15,.20,.30\}$; the .10 familywise guarantee does not jointly cover
different targets, models, budgets, or scores.

The Omni grids were registered before its CAL release. The DAPO S1 replay
uses the same recorded grids on already released CAL500/TEST500, and its
provenance does not establish independence of grid construction from those
calibration data. We retain its empirical counts as retrospective
diagnostics. The theoretical statement requires a grid fixed independently
of CAL; holding a data-chosen grid fixed during replay alone does not
establish that premise. The TEST Clopper--Pearson bound is a separate
description of the selected policy, not the p-value used to issue it.

\begin{samepage}
\subsection{Artifacts, costs, and numerical reproducibility}

The pinned model and data revisions are:
\begin{center}\small
\begin{tabular}{ll}
Qwen3-8B & \nolinkurl{b968826d9c46dd6066d109eabc6255188de91218}\\
Qwen3.5-4B & \nolinkurl{851bf6e806efd8d0a36b00ddf55e13ccb7b8cd0a}\\
Omni-MATH & \nolinkurl{40ba231d8f16e29ecd40e6407e2c8640145a8f62}
\end{tabular}
\end{center}
\end{samepage}
The seven parent-study Tinker adapters have SHA-256-verified archives
totaling 2,352,015,360 bytes. The archived 4B tensors require linear-attention
Q/K/V fusion and an untied output projection for reconstruction.
The A100 program closes with status \texttt{closed\_partial}, documenting
the unsupported 4B dynamic-LoRA path and the completed merged-weight runs.
Archived adapters reconstruct in HF PEFT and vLLM with held-out
raw-log-probability mean absolute errors of 0.0007--0.0033. Behavioral
checks comprise 32 vLLM cells at 50 families per cell and 16 HF cells
at four families per cell; six vLLM cells have at least one pointwise
$R/W/U$ interval excluding zero (Appendix~\ref{app:audited-diagnostics}).

All dollar comparisons are within model size at frozen study rates.
The 8B rates per million tokens are \$.195 ordinary input, \$.039 cached
input, \$.600 output, and \$.440 training; 4B rates are
\$.330, \$.066, \$1.005, and \$.737. Study-local ledgers report
\$55.430052 for the initial RL pilot including projected storage,
\$184.044319 for 2A including reserves, \$50.919490 for the independent
OOD confirmation, and \$93.743151 for the final supplement including
\$.022542 estimated storage. The latter token total is \$93.720609,
including \$.046575 reconciled at session level for 12 original calls
whose individual receipts could not be recovered after a disk incident.
These differently scoped ledgers are not a complete all-project total. The A100 audit used a four-card, 40-GiB-per-card
PCIe A100 node during a 16-hour allocation.

The released numerical bundle contains raw study-local JSON, six typed
CSV tables, family-level white-box/Native inputs, and a deterministic
verifier. In those CSVs, Table 1 rates are percentages; Tables 2, 3, and 5
store fractions. Table 4 stores operating-point counts and TEST denominators.
Repeated analyses across seeds, caps, scores, and thresholds do not add
independent families.

\subsection{Cross-domain LiveCodeBench protocol}
\label{app:lcb-protocol}

We freeze LiveCodeBench \texttt{release\_v6}~\citep{jain2025livecodebench}
at dataset commit \nolinkurl{0fe84c3912ea0c4d4a78037083943e8f0c4dd505}.
The 1,052 eligible questions have unique IDs and normalized-statement hashes
and contest dates from May 2023 through April 6, 2025. Six difficulty-by-I/O
strata supply the disjoint 20-question pilot; deterministic SHA-256 ordering
assigns the remaining questions to CAL300, TEST300, and RESERVE432.
The reserve remains sealed. The formal matrix comprises frozen Qwen3-8B
base and the two mathematics-trained Binary RL checkpoints (seeds 41 and 73,
step 50), each at 4K and 8K: 3,600 single-rollout deployments on 600 families.
The local tokenizer revision is
\nolinkurl{b968826d9c46dd6066d109eabc6255188de91218}; checkpoint bindings match
the training archive. Appendix~\ref{app:lcb-extended} covers the 4B pilot.

\paragraph{Generation, validity, and scoring.}
Prompts contain the problem, any starter code, and the prescribed Python
output format; grading outcomes never enter the prompt. Sampling uses
temperature 1.0, top-$p$ 0.95, top-$k$ 20, and one sample per request.
SHA-256 seeds depend on the split, budget, question ID, and generation
segment and are shared across comparison arms. Budget forcing allocates
$2+2046+2+2046$ tokens at 4K and $2+4094+2+4094$ at 8K, including the two
prefill and two injected delimiter tokens. The delimiter is injected only
if thinking remains open at the midpoint; a natural thinking end is not
repeated. The 8K first segment also supplies 1,800 paired Native 4K
replays without additional generation. A first segment returning 4,094
tokens counts as truncated, even if EOS falls exactly at the cap; no such
EOS boundary cases occurred. An earlier natural completion reuses its
original code and grade. Native costs are the first-call equivalent,
while the replay's additional cost is zero.

The reference-free validity indicator $v(x,y)$ requires normal chat EOS,
an end-of-thinking delimiter, and a complete nonempty Python code fence
whose contents pass \texttt{ast.parse()}. Extraction uses the last Python-tagged
fence, falling back to the last untagged fence. Invalid outputs receive $U$
and cannot be submitted. The answer span follows the first end-of-thinking
delimiter; the code span contains the extracted program; the full trace also
includes sampled thinking. All spans exclude prompt, prefill, injected, and
EOS tokens. A token overlapping the code's half-open byte interval contributes
its full recorded token surprise. Each span has nine fixed scores: Mean,
Tail-10/5/2, Sum, Max, length, and averages of the largest 16/32 surprises.
The primary method is answer-span Tail-10 at $\alpha=0.065$; the other spans,
scores, and targets $0.05/0.10$ are prespecified comparisons.

\paragraph{Execution and outcome labels.}
The evaluator uses CPython 3.11.13 in a macOS Seatbelt sandbox, with network
access, host-data access, and child-process creation disabled. Candidate code
sees only the current test input; reference outputs remain in the parent
process. All 12 pre-sampling self-tests passed, covering stdin and functional
entry points, correct and incorrect outputs, syntax and runtime failures,
isolation, timeout, and memory limits. Each case has a 6-second deadline and
each task a 30-second deadline. The parent monitors a 1-GiB physical-memory
limit every 5 ms, allowing transient overshoot between checks. A valid program
receives $R$ only if all official public and private tests pass; its first
failure assigns $W$ and a wrong-answer, runtime-error, timeout, memory-limit,
or other failure category. The primary labels retain the official tests;
a separately locked sensitivity analysis removes the six listed CAL errata
and one TEST erratum, using CAL294 and TEST299.

\paragraph{Blinding and file identities.}
All trajectories, validity decisions, and scores were recorded before any
CAL grading. The 1,458 primary and errata-excluded cutoffs were then locked
before TEST labels were released. The six recorded events below are on
November 25, 2026; times are UTC.
\begin{center}\small
\setlength{\tabcolsep}{4pt}
\renewcommand{\arraystretch}{1.15}
\begin{tabular}{@{}L{.39\linewidth}L{.25\linewidth}L{.31\linewidth}@{}}
\toprule
Event & UTC time & Record\\
\midrule
Protocol and panels locked & 04:01:25.811827 & \nolinkurl{protocol.json}\\
First physical reservation & 04:03:51.926192 & \nolinkurl{step1_ledger.jsonl}\\
Trajectories and scores complete & 04:29:07.808955 & \nolinkurl{generation_complete.json}\\
Calibration cutoffs locked & 05:45:45.327243 & \nolinkurl{locked_cal_thresholds.json}\\
Test labels released & 05:45:45.364908 & \nolinkurl{test_release.json}\\
Test grading complete & 05:57:57.034769 & \nolinkurl{test_grading_complete.json}\\
\bottomrule
\end{tabular}
\end{center}
The protocol and frozen-cutoff files have SHA-256 digests
\begin{center}\small
\begin{tabular}{@{}L{.20\linewidth}L{.76\linewidth}@{}}
Protocol & \nolinkurl{ed6200c0eb93caee6e3bfbac90835e2f0ad9b81b853257f05ca5b36bf42173c9}\\
Locked cutoffs & \nolinkurl{2c55623454c995c9227b4d9f33d683cdd65b6f7cf96b876bcd3c9e664f783f8b}\\
\end{tabular}
\end{center}
The protocol additionally pins the panel manifest, prompts, sampler,
span scorer, grader, sandbox worker, and self-test record by hash.
The released protocol, thresholds, and independent-verification JSON retain
these identities and the temporal ordering.

\paragraph{Calibration, inference, and cost.}
Every assigned family, including $U$, contributes to the calibration
size. Equation~\eqref{eq:crc-cut} is fitted separately by model, budget,
policy, span, score, and target, with strict cutoff ties rejected.
For CAL300, the three targets yield $j=15,19,30$. Binary RL summaries apply
each seed's own threshold and average its action indicators within each
family; they do not ensemble two answers. Ranking summaries average
seed-specific estimates over valid $R/W$ outputs. Partial AUROC treats
correct answers as positive, uses negative surprise as confidence, and
integrates up to wrong-answer acceptance rates 0.10 and 0.20 with McClish
standardization.

Pointwise 95\% intervals use 20,000 paired family bootstrap draws, with
NumPy seed 20261125, retaining all arms and seeds together and conditioning
on the fitted models and cutoffs. Undefined resampled statistics are omitted
with their defined-replicate counts. One-sided 95\% Clopper--Pearson bounds
are reported for individual-model TEST wrong-submission rates; fractional
seed averages receive no binomial bound. After TEST grading, 2,000 shared
300/300 repartitions of the fixed 600-family pool refit the cutoffs and
summarize the mean and SD of wrong submissions. These measure finite-pool
calibration variability; the exchangeable-population expectation guarantee
is Theorem~\ref{thm:crc}. Year-specific results apply the original full-CAL
thresholds without refitting within each year.

The formal ledger contains 6,459 physical sampling receipts and
\$10.665913185 in gross generation charges; the separate 120-deployment
pilot uses 229 receipts and \$0.544179546.
The frozen 8B rates per million cold-input, cached-input, and output tokens
are \$0.195, \$0.039, and \$0.600. Physical requests include continuation
segments of a single deployment; these costs precede account credits and
contain no new training. The supplement contains six analysis CSVs, the
protocol, cutoffs, and independent verification.

\clearpage
\section{Expanded quantitative diagnostics}
\label{app:audited-diagnostics}

\subsection{Raw solver contrasts and score behavior}

\begin{table*}[!htbp]
\centering\small
\begingroup
\renewcommand{\arraystretch}{1.16}
\setlength{\tabcolsep}{4pt}
\begin{tabular}{@{}lllrrr@{}}
\toprule
Domain & Model & Cap & \shortstack{$\Delta R$ vs. ExtBF base\\(pp; 95\% CI)} & \shortstack{$\Delta R$ vs. ExtBF SFT\\(pp; 95\% CI)} & \shortstack{G0 cost\\(USD/family)} \\
\midrule
DAPO & 4B & 4K & +26.70 [23.10, 30.40] & +12.30 [9.20, 15.60] & 0.004258 \\
DAPO & 4B & 8K & +29.60 [25.80, 33.40] & +8.40 [5.20, 11.60] & 0.007593 \\
\addlinespace[3pt]
DAPO & 8B & 4K & +12.80 [9.20, 16.50] & +5.80 [2.30, 9.30] & 0.002477 \\
DAPO & 8B & 8K & +4.30 [1.00, 7.60] & +0.90 [-2.40, 4.20] & 0.003996 \\
\addlinespace[3pt]
Omni & 4B & 4K & +15.00 [11.50, 18.83] & +8.33 [5.33, 11.33] & 0.004364 \\
Omni & 4B & 8K & +16.17 [12.50, 20.00] & +8.83 [5.33, 12.33] & 0.008077 \\
\addlinespace[3pt]
Omni & 8B & 4K & +6.67 [3.00, 10.33] & +7.33 [3.50, 11.33] & 0.002436 \\
Omni & 8B & 8K & +7.83 [4.00, 12.00] & +4.17 [0.00, 8.33] & 0.003947 \\
\bottomrule
\end{tabular}
\endgroup

\caption{Raw G0 versus same-budget ExtBF controls. Differences are
percentage points with pointwise family-paired 95\% intervals. DAPO
comparisons retain the original 2A resampling; Omni comparisons retain the
independent OOD study. G0 costs are expected single-seed deployment costs.}
\label{tab:audited-raw-contrasts}
\end{table*}

\begin{table}[!htbp]
\centering\small
\begingroup
\renewcommand{\arraystretch}{1.16}
\setlength{\tabcolsep}{5pt}
\begin{tabular}{@{}lllrrr@{}}
\toprule
Domain & Model & Cap & \shortstack{Native base\\$R$ (\%)} & \shortstack{Native G0$_{41}$\\$R$ (\%)} & \shortstack{$\Delta R$\\(pp; 95\% CI)} \\
\midrule
DAPO & 4B & 4K & 0.80 & 2.40 & +1.60 [0.40, 3.00] \\
DAPO & 4B & 8K & 7.00 & 14.20 & +7.20 [5.00, 9.60] \\
\addlinespace[3pt]
DAPO & 8B & 4K & 7.60 & 8.20 & +0.60 [-0.80, 2.20] \\
DAPO & 8B & 8K & 35.40 & 35.00 & -0.40 [-3.80, 3.00] \\
\addlinespace[3pt]
Omni & 4B & 4K & 2.33 & 5.00 & +2.67 [0.33, 5.00] \\
Omni & 4B & 8K & 8.67 & 12.00 & +3.33 [1.00, 6.00] \\
\addlinespace[3pt]
Omni & 8B & 4K & 10.00 & 10.33 & +0.33 [-1.33, 2.00] \\
Omni & 8B & 8K & 17.67 & 18.00 & +0.33 [-1.67, 2.33] \\
\bottomrule
\end{tabular}
\endgroup

\caption{Native transfer of the 4K-trained G0 seed 41.
Rates are percentages, and differences are percentage points with
20,000-resample paired family-bootstrap 95\% intervals.}
\label{tab:native-transfer}
\end{table}

\begin{table*}[p]
\centering\scriptsize
\begingroup
\renewcommand{\arraystretch}{1.12}
\setlength{\tabcolsep}{3pt}
\footnotesize
\begin{tabular}{@{}llllrrrrrr@{}}
\toprule
Domain & Model & Arm & Cap & \shortstack{Box\\sum} & \shortstack{Answer\\sum} & \shortstack{Answer\\mean} & \shortstack{Tail\\NLL} & \shortstack{Tail\\entropy} & \shortstack{Entropy $-$ NLL\\(95\% CI)} \\
\midrule
DAPO & 4B & E1\_41 & 4K & 0.651 & 0.289 & 0.703 & 0.688 & 0.692 & +0.004 [-0.055, 0.070] \\
DAPO & 4B & E1\_41 & 8K & 0.712 & 0.497 & 0.721 & 0.743 & 0.729 & -0.014 [-0.082, 0.057] \\
Omni & 4B & E1\_41 & 4K & 0.863 & 0.529 & 0.655 & 0.630 & 0.692 & +0.062 [-0.028, 0.170] \\
Omni & 4B & E1\_41 & 8K & 0.654 & 0.686 & 0.652 & 0.649 & 0.660 & +0.011 [-0.033, 0.056] \\
\addlinespace[3pt]
DAPO & 4B & G0\_41 & 4K & 0.739 & 0.637 & 0.626 & 0.637 & 0.625 & -0.013 [-0.058, 0.028] \\
DAPO & 4B & G0\_41 & 8K & 0.696 & 0.783 & 0.753 & 0.754 & 0.781 & +0.027 [0.003, 0.052] \\
Omni & 4B & G0\_41 & 4K & 0.697 & 0.661 & 0.757 & 0.741 & 0.758 & +0.017 [-0.017, 0.050] \\
Omni & 4B & G0\_41 & 8K & 0.609 & 0.656 & 0.743 & 0.740 & 0.737 & -0.003 [-0.026, 0.019] \\
\addlinespace[3pt]
DAPO & 4B & G0\_73 & 4K & 0.544 & 0.640 & 0.658 & 0.663 & 0.656 & -0.007 [-0.048, 0.035] \\
DAPO & 4B & G0\_73 & 8K & 0.659 & 0.733 & 0.795 & 0.786 & 0.789 & +0.003 [-0.029, 0.038] \\
Omni & 4B & G0\_73 & 4K & 0.697 & 0.629 & 0.740 & 0.741 & 0.739 & -0.002 [-0.027, 0.022] \\
Omni & 4B & G0\_73 & 8K & 0.708 & 0.571 & 0.670 & 0.656 & 0.670 & +0.014 [-0.012, 0.042] \\
\addlinespace[3pt]
DAPO & 4B & base & 4K & 0.782 & 0.111 & 0.794 & 0.779 & 0.782 & +0.003 [-0.020, 0.025] \\
DAPO & 4B & base & 8K & 0.524 & 0.183 & 0.769 & 0.737 & 0.690 & -0.047 [-0.128, 0.007] \\
Omni & 4B & base & 4K & 0.865 & 0.716 & 0.697 & 0.677 & 0.723 & +0.045 [-0.092, 0.176] \\
Omni & 4B & base & 8K & 0.555 & 0.642 & 0.735 & 0.744 & 0.709 & -0.035 [-0.110, 0.027] \\
\addlinespace[3pt]
DAPO & 8B & G0\_41 & 4K & 0.768 & 0.862 & 0.824 & 0.830 & 0.856 & +0.026 [0.007, 0.046] \\
DAPO & 8B & G0\_41 & 8K & 0.845 & 0.843 & 0.755 & 0.779 & 0.816 & +0.037 [0.020, 0.055] \\
Omni & 8B & G0\_41 & 4K & 0.609 & 0.736 & 0.798 & 0.791 & 0.772 & -0.019 [-0.041, 0.000] \\
Omni & 8B & G0\_41 & 8K & 0.622 & 0.717 & 0.727 & 0.729 & 0.721 & -0.007 [-0.030, 0.015] \\
\addlinespace[3pt]
DAPO & 8B & G0\_73 & 4K & 0.780 & 0.859 & 0.800 & 0.807 & 0.841 & +0.034 [0.018, 0.051] \\
DAPO & 8B & G0\_73 & 8K & 0.843 & 0.847 & 0.773 & 0.791 & 0.821 & +0.031 [0.015, 0.047] \\
Omni & 8B & G0\_73 & 4K & 0.647 & 0.717 & 0.729 & 0.721 & 0.714 & -0.007 [-0.029, 0.016] \\
Omni & 8B & G0\_73 & 8K & 0.622 & 0.753 & 0.710 & 0.719 & 0.722 & +0.003 [-0.016, 0.023] \\
\addlinespace[3pt]
DAPO & 8B & G1\_41 & 4K & 0.838 & 0.772 & 0.744 & 0.756 & 0.774 & +0.019 [-0.006, 0.044] \\
DAPO & 8B & G1\_41 & 8K & 0.839 & 0.857 & 0.854 & 0.862 & 0.884 & +0.023 [0.005, 0.041] \\
Omni & 8B & G1\_41 & 4K & 0.637 & 0.728 & 0.767 & 0.757 & 0.762 & +0.005 [-0.020, 0.032] \\
Omni & 8B & G1\_41 & 8K & 0.622 & 0.739 & 0.800 & 0.790 & 0.788 & -0.002 [-0.025, 0.021] \\
\addlinespace[3pt]
DAPO & 8B & M5\_41 & 4K & 0.827 & 0.838 & 0.795 & 0.801 & 0.816 & +0.015 [-0.006, 0.037] \\
DAPO & 8B & M5\_41 & 8K & 0.873 & 0.836 & 0.807 & 0.826 & 0.844 & +0.018 [-0.001, 0.038] \\
Omni & 8B & M5\_41 & 4K & 0.690 & 0.810 & 0.804 & 0.807 & 0.809 & +0.001 [-0.025, 0.026] \\
Omni & 8B & M5\_41 & 8K & 0.602 & 0.703 & 0.742 & 0.742 & 0.744 & +0.002 [-0.020, 0.024] \\
\addlinespace[3pt]
DAPO & 8B & base & 4K & 0.825 & 0.832 & 0.827 & 0.826 & 0.847 & +0.020 [-0.001, 0.043] \\
DAPO & 8B & base & 8K & 0.845 & 0.792 & 0.827 & 0.843 & 0.873 & +0.031 [0.014, 0.050] \\
Omni & 8B & base & 4K & 0.657 & 0.773 & 0.773 & 0.771 & 0.744 & -0.026 [-0.054, 0.001] \\
Omni & 8B & base & 8K & 0.646 & 0.716 & 0.756 & 0.754 & 0.761 & +0.007 [-0.017, 0.031] \\
\bottomrule
\end{tabular}
\endgroup

\caption{All 36 uncertainty cells. AUROC is evaluated among valid
submissions with wrong answers as the positive class. The entropy-minus-NLL
intervals use paired family resampling with up to 2,000 estimable draws,
are pointwise, and have no correction across cells. A zero-crossing interval
does not establish score equivalence.}
\label{tab:audited-uncertainty}
\end{table*}

Near-zero box NLL is not specific to KL-free RL. On DAPO 8B 4K,
the median box NLL is $1.29\times10^{-10}$ for M5,
$8.92\times10^{-8}$ for base, $1.55\times10^{-7}$ for G1, and
$5.80\times10^{-7}$ for pooled G0. The fractions below $10^{-5}$ are
90.16\%, 86.73\%, 84.85\%, and 74.96\%, respectively. None is exactly
zero. In paired comparisons restricted to identical final answers, G0 has
lower box NLL than base in 17.13\% of 251 pairs, and than M5 in .66\% of
304 pairs. These diagnostics concern only matched final answers. They neither
establish global logit collapse nor isolate the training objective's causal effect.

Length normalization also is not a universal guarantee of superior ranking.
On the fresh Omni pooled-G0 cells, differences between tail10 and
answer-mean AUROC are small and their paired intervals include zero.
The tail-versus-length advantages are .2198 $[.0989,.3356]$ for 4B4K,
.1550 $[.0474,.2649]$ for 4B8K, .1207 $[.0483,.1946]$ for 8B4K,
and .0685 $[-.0018,.1396]$ for 8B8K. Thus the strong failure of cumulative
NLL on conservative 4B models establishes that length can confound a sum;
it does not establish that tail10 is the best score in every cell.

\subsection{Calibration variability and original gate failures}

\begin{table}[t]
\centering\small
\setlength{\tabcolsep}{4pt}
\begingroup
\renewcommand{\arraystretch}{1.16}
\setlength{\tabcolsep}{5pt}
\begin{tabular}{@{}llrrr@{}}
\toprule
Domain & Score & \shortstack{CAL\\issuance} & \shortstack{Empirical TEST error\\meets target} & \shortstack{95\% TEST upper bound\\meets target} \\
\midrule
DAPO & Tail NLL & 49/72 & 49/49 & 45/49 \\
DAPO & Answer-mean NLL & 47/72 & 47/47 & 41/47 \\
\addlinespace[3pt]
Omni & Tail NLL & 3/72 & 3/3 & 0/3 \\
Omni & Answer-mean NLL & 3/72 & 3/3 & 1/3 \\
\bottomrule
\end{tabular}
\endgroup

\caption{Fixed-grid LTT outcomes across nine individual models, two caps,
and four targets per domain and score. Issuance uses CAL; TEST columns
describe empirical selective error and its one-sided 95\% upper bound.
DAPO is retrospective; Omni uses independently registered grids.
All Omni issuances occur at the 30\% target.}
\label{tab:audited-ltt}
\end{table}

The retrospective 5,000-resplit review of original 8B DAPO 4K box-score
policies found that both primary marginal wrong-answer targets were exceeded
in 37.24\% of splits. Both realized wrong-answer rates were at least as
large as those of the original fixed split in 12.72\%; the two rates had
correlation .8329. This finite-pool description of correlated calibration fluctuation
leaves the failed gates and the separate evidence of poor score ranking
under OOD representation shift intact.

The S1 calibration-size analysis uses 2,000 family repartitions per
configuration. DAPO uses CAL sizes 50, 100, 200, 300, and 500 from its
1,000 released families; Omni uses 50, 100, 200, and 300 from 600.
At 8B4K and $\alpha=.065$, DAPO's TEST-$W$ SD decreases from 2.573 to
1.187 pp, with mean 6.407\% at CAL500; Omni's decreases from 2.708 to
1.581 pp, with mean 6.307\% at CAL300. Changing TEST denominators, discreteness, and all-pass cutoffs affect
these curves. The table retains all 72 configurations: one curve has
increased endpoint SD, and five are not nonincreasing across intermediate sizes.

\begin{table}[t]
\centering\scriptsize
\setlength{\tabcolsep}{4pt}
\begingroup
\renewcommand{\arraystretch}{1.14}
\setlength{\tabcolsep}{4pt}
\footnotesize
\begin{tabular}{@{}llllrrrr@{}}
\toprule
Domain & Model & Cap & $\alpha$ & \shortstack{CAL\\families} & \shortstack{Mean $W$\\(\%)} & \shortstack{SD\\(pp)} & \shortstack{SD never\\increases?} \\
\midrule
DAPO & 4B & 4K & 0.065 & 50$\to$500 & 4.100$\to$4.832 & 0.980$\to$0.600 & No \\
DAPO & 4B & 4K & 0.1 & 50$\to$500 & 4.752$\to$4.874 & 0.392$\to$0.533 & No \\
DAPO & 4B & 8K & 0.065 & 50$\to$500 & 4.601$\to$5.573 & 1.252$\to$0.842 & No \\
DAPO & 4B & 8K & 0.1 & 50$\to$500 & 5.659$\to$5.903 & 0.601$\to$0.600 & No \\
\addlinespace[3pt]
DAPO & 8B & 4K & 0.065 & 50$\to$500 & 5.951$\to$6.407 & 2.573$\to$1.187 & Yes \\
DAPO & 8B & 4K & 0.1 & 50$\to$500 & 9.818$\to$9.990 & 3.093$\to$1.486 & Yes \\
DAPO & 8B & 8K & 0.065 & 50$\to$500 & 5.908$\to$6.365 & 2.606$\to$1.175 & Yes \\
DAPO & 8B & 8K & 0.1 & 50$\to$500 & 9.748$\to$9.972 & 3.377$\to$1.489 & Yes \\
\addlinespace[3pt]
Omni & 4B & 4K & 0.065 & 50$\to$300 & 5.646$\to$6.297 & 2.338$\to$1.607 & Yes \\
Omni & 4B & 4K & 0.1 & 50$\to$300 & 8.297$\to$9.252 & 2.011$\to$1.502 & No \\
Omni & 4B & 8K & 0.065 & 50$\to$300 & 5.857$\to$6.262 & 2.872$\to$1.755 & Yes \\
Omni & 4B & 8K & 0.1 & 50$\to$300 & 9.156$\to$9.901 & 2.842$\to$2.149 & Yes \\
\addlinespace[3pt]
Omni & 8B & 4K & 0.065 & 50$\to$300 & 5.904$\to$6.307 & 2.708$\to$1.581 & Yes \\
Omni & 8B & 4K & 0.1 & 50$\to$300 & 9.854$\to$9.935 & 3.406$\to$1.956 & Yes \\
Omni & 8B & 8K & 0.065 & 50$\to$300 & 5.905$\to$6.317 & 2.843$\to$1.659 & Yes \\
Omni & 8B & 8K & 0.1 & 50$\to$300 & 9.854$\to$10.006 & 3.573$\to$2.010 & Yes \\
\bottomrule
\end{tabular}
\endgroup

\caption{All sixteen calibration-size curves. Arrows run from smallest to largest
CAL size; the final column checks every recorded intermediate size. Each point uses 2,000 fixed-pool repartitions; remaining families form TEST.}
\label{tab:ncal-all}
\end{table}

Track B's 28/36 and 8/36 counts include nine individual model arms,
two budgets, and targets .065/.10, using tail10 only. Pooled-G0 rows,
alternative scores, other targets, and Track A are excluded. Two conspicuous
pooled exceedances at $\alpha=.10$ are 4B Omni8K $W=11.83\%$ and 8B
Omni8K $W=12.00\%$. Reporting these single-split outcomes together with
the resplit means preserves the distinction between an expected risk
guarantee and realized finite-test attainment.

For tail10 LTT on DAPO, the numbers issued at targets
.10/.15/.20/.30 are 4/18, 12/18, 15/18, and 18/18; their TEST upper-bound
attainments are 4, 12, 14, and 15. For answer mean, issuance is
5/18, 11/18, 13/18, and 18/18, with upper-bound attainments 4, 11, 12,
and 14. Every issued DAPO setting has empirical TEST conditional error
below its target. Each score issues three Omni policies, all at target .30;
all empirical errors meet target, with zero tail10 and one answer-mean
TEST upper-bound attainment. These counts describe repeated analyses of
shared families, not independent replications.

\subsection{Penalty and selection diagnostics}

\begin{figure}[t]
\centering
\includegraphics[width=.85\linewidth]{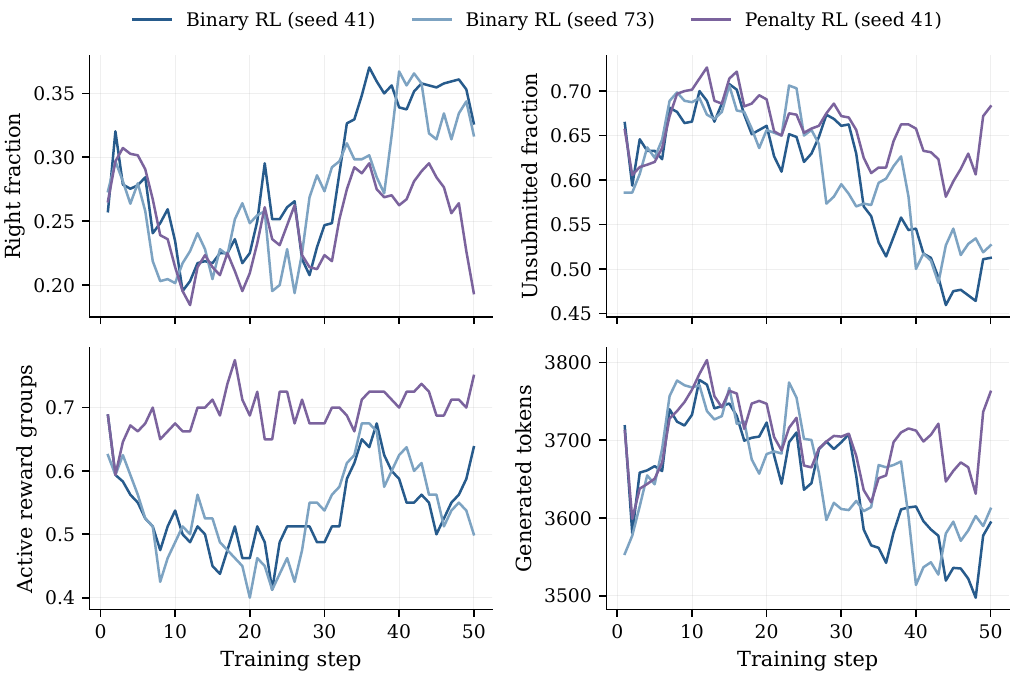}
\caption{Eight-billion-parameter training traces for Binary RL (seeds 41, 73)
and Penalty RL (seed 41). Curves are trailing five-update means on the changing training
batches. The active-group statistic records nonconstant rewards, not
correct evidence: a penalty also activates failure-only mixed groups.
These curves describe behavior on the training batches, without estimating
accuracy on held-out problems.}
\label{fig:training}
\end{figure}

The pooled G0 comparisons in the main text retain both training seeds.
For a same-seed reference, DAPO4K G0$_{41}$ has $R=40.4\%$,
$W=19.4\%$, and truncation 36.6\%, versus G1$_{41}$ at
29.4\%, 10.2\%, and 58.2\%. Their mean sampled output counts are
3,429 and 3,628. S3A and S3B each have only one training seed, limiting conclusions
about reproducibility across runs.
The first/last-ten-update training truncation means are 64.92/62.73\%
for G1 and 63.75/47.27\% for G0$_{41}$.
The evidence is weaker reduction in budget exhaustion, not monotonic
growth in training truncation (Figure~\ref{fig:training}).

After matched 4K CRC, pooled G0 minus S3A net-score differences are
5.6 pp $[1.0,10.2]$ and 9.4 pp $[4.3,14.6]$ on DAPO at .065 and .10;
the corresponding differences versus S3B are 5.0 pp $[.5,9.6]$ and
9.0 pp $[4.2,13.8]$. On Omni, differences versus S3A are
2.5 pp $[-2.0,7.17]$ and 3.67 pp $[-1.33,8.83]$; differences versus
S3B are 3.83 pp $[-.17,8.0]$ and 4.33 pp $[-.67,9.67]$.
For DAPO at .065, realized $W$ is 7.1\% for G0, 9.6\% for S3A,
and 10.2\% for S3B. Equal nominal targets need not produce equal realized
TEST risk, so the net differences should be read with the operating points.

The 455-point G0 risk--coverage file orders thresholds using CAL and
evaluates them on TEST. Seven raw supplementary-arm points lie below its
observed coverage/error frontier; DAPO8K S3B is the exception described
in the main text. Selecting a preferred TEST point from this sweep is
descriptive and receives no independent selective-policy guarantee.
The full transition matrices retain right, wrong, truncation, explicit
abstention, and other invalid outcomes separately.

\subsection{White-box comparison and backend uncertainty}

The entropy-versus-Native post hoc analysis joins exactly 500 DAPO TEST
family identifiers, verifies the individual seed counts against the
frozen A2 cuts, averages within family, and applies 20,000 multinomial
family resamples with RNG seed 20261122. The supplied verifier reproduces
the reported $\Delta R$, $\Delta W$, and $\Delta\mathrm{Net}$ intervals.
These intervals condition on calibration cuts, replayed trajectories,
and the two trained policies; they do not include full training or
calibration-refit variation. Reported generation times exclude the
separate pass that extracts entropy features.

Backend behavioral comparisons contain 144 metric rows corresponding to
48 cells: 32 vLLM cells at $n=50$ and 16 HF cells at $n=4$, with separate
$R/W/U$ intervals per cell. In the vLLM subset, two $R$ intervals, two
$W$ intervals, and three $U$ intervals exclude zero; these occupy six
distinct cells. For example, 4B G0 on Omni TEST at 4K has a wrong-answer
change interval of $[-16,-2]$ pp. Short fixed-token reconstruction probes
and these finite-sample behavior checks are different forms of evidence;
neither implies exact equality of backend generation distributions.

\subsection{Expanded mathematics score anatomy and verbosity pilot}
\label{app:math-stage1}

The expanded score audit reuses 28,800 historical CAL/TEST deployments
from 1,600 mathematics families and 36 model--domain--budget settings.
It compares tail fractions, fixed-size tails, means, and maxima on the
same saved trajectories. Table~\ref{tab:math-stage1} reports the
retrospective ranking contrasts and their calibrated operating points.
The answer-span Tail-10 advantage concentrates on 8B DAPO 8K, where all
five arms improve on Mean. Across complete trajectories, Tail-10 exceeds
Max in 31 of 36 settings, with mean AUROC gain
$0.0649\,[0.0342,0.0952]$. The exploratory CAL-selected score in the
table changes the selection procedure; its larger retained coverage
requires a separate correction for score selection.

The background diagnostics clarify the assumptions behind
Corollary~\ref{cor:tail-gain-main}. For 8B Binary RL, seed 41, on DAPO
8K, the bottom 90\% of answer-token surprises still contribute 25.8\%
of total NLL. Their length-normalized correct/wrong separation is
$0.576\,[0.431,0.724]$, and their within-class correlation with the tail
component is 0.885. Thus these observed background components carry both
label information and shared variation. The sparse-background model
isolates a mechanism; its independent, label-neutral background is not
an empirical identity for these trajectories. Natural long/short strata
also change the mix of problems and solutions alongside answer length.

A separate CAL20 verbosity smoke experiment used 8B Binary RL, seed 41,
at 8K on 12 previously correct and eight previously wrong DAPO families.
It compared moderate and strong expansion prompts with either the original
thinking prefix fixed or the entire trajectory regenerated. Relative to
the original median of 928.5 answer tokens, the four median-length ratios
were 0.99/1.48 for fixed prefixes and 0.97/1.64 for full regeneration,
below the respective 1.8/3.0 targets. Only two of 80 deployments truncated,
but fixed-prefix labels agreed with the original in 17/20 and 18/20
families. These prompts therefore did not supply the intended
length-only intervention; the planned 1,000-family experiment was not
run. The result identifies a practical obstacle to testing background
dilution with free-form expansion prompts, rather than estimating a
causal gain from added low-surprise tokens.

\begin{table}[p]
\centering
\small
\setlength{\tabcolsep}{2pt}
\renewcommand{\arraystretch}{1.13}
\begin{tabular*}{\linewidth}{@{\extracolsep{\fill}}lrrrrr@{}}
\toprule
\multicolumn{6}{@{}l}{\textbf{(a) AUROC contrasts}}\\
Population / contrast & Span & $\Delta$AUROC & 95\% CI & Positive & CI positive \\
\midrule
All 36: Tail-10 $-$ Mean & Answer & $-0.0003$ & $[-0.0061,+0.0056]$ & 18/36 & 5/36 \\
All 36: Tail-10 $-$ Mean & Full & $+0.0055$ & $[+0.0008,+0.0105]$ & 23/36 & 8/36 \\
All 36: change in gap & Full $-$ answer & $+0.0058$ & $[+0.0003,+0.0115]$ & 26/36 & 4/36 \\
All 36: Tail-10 $-$ Max & Full & $+0.0649$ & $[+0.0342,+0.0952]$ & 31/36 & 12/36 \\
8B 8K: Tail-10 $-$ Mean & Answer & $+0.0167$ & $[+0.0108,+0.0228]$ & 5/5 & 4/5 \\
8B 8K: Tail-10 $-$ Mean & Full & $+0.0213$ & $[+0.0153,+0.0275]$ & 5/5 & 4/5 \\
8B 8K: Tail-H $-$ Mean-H & Answer & $+0.0393$ & $[+0.0250,+0.0537]$ & 5/5 & 5/5 \\
8B 8K: long $-$ short gap & Answer & $+0.0164$ & $[+0.0051,+0.0278]$ & 5/5 & 1/5 \\
\addlinespace[7pt]
\multicolumn{6}{@{}l}{\textbf{(b) Calibrated submission: 8B Binary RL, DAPO 8K}}\\
Score & Mean $R/W$ & Score $R/W$ & $\Delta R$ & $\Delta$Net & Net 95\% CI \\
\midrule
Answer Tail-10 & 37.7 / 6.2 & 39.1 / 5.9 & $+1.4$ & $+1.7$ & $[+0.7,+2.8]$ \\
Answer Tail-5 & 37.7 / 6.2 & 41.5 / 6.4 & $+3.8$ & $+3.6$ & $[+2.0,+5.2]$ \\
Full Tail-10 & 36.3 / 6.2 & 37.8 / 5.2 & $+1.5$ & $+2.5$ & $[+1.0,+4.0]$ \\
Answer Tail-H & 40.1 / 6.8 & 43.7 / 6.2 & $+3.6$ & $+4.2$ & $[+2.5,+6.0]$ \\
Full CAL-selected tail & 36.3 / 6.2 & 47.4 / 7.3 & $+11.1$ & $+10.0$ & $[+6.7,+13.2]$ \\
\addlinespace[7pt]
\multicolumn{6}{@{}l}{\textbf{(c) Full-trace tail fraction and fixed-size alternatives}}\\
Score $-$ Mean & \multicolumn{2}{r}{All 36: $\Delta$AUROC} & \multicolumn{3}{r}{8B DAPO 8K: $\Delta$AUROC} \\
\midrule
Tail-20 & \multicolumn{2}{r}{$+0.0025\,[+0.0004,+0.0049]$} & \multicolumn{3}{r}{$+0.0089\,[+0.0063,+0.0118]$} \\
Tail-10 & \multicolumn{2}{r}{$+0.0055\,[+0.0008,+0.0105]$} & \multicolumn{3}{r}{$+0.0213\,[+0.0153,+0.0275]$} \\
Tail-5 & \multicolumn{2}{r}{$+0.0063\,[-0.0014,+0.0143]$} & \multicolumn{3}{r}{$+0.0335\,[+0.0242,+0.0434]$} \\
Tail-2 & \multicolumn{2}{r}{$+0.0035\,[-0.0086,+0.0158]$} & \multicolumn{3}{r}{$+0.0449\,[+0.0307,+0.0598]$} \\
Tail-1 & \multicolumn{2}{r}{$+0.0009\,[-0.0150,+0.0172]$} & \multicolumn{3}{r}{$+0.0489\,[+0.0305,+0.0684]$} \\
Top-8 & \multicolumn{2}{r}{$-0.0103\,[-0.0357,+0.0171]$} & \multicolumn{3}{r}{$+0.0360\,[+0.0037,+0.0691]$} \\
Top-16 & \multicolumn{2}{r}{$-0.0046\,[-0.0278,+0.0204]$} & \multicolumn{3}{r}{$+0.0487\,[+0.0182,+0.0796]$} \\
Top-32 & \multicolumn{2}{r}{$+0.0018\,[-0.0194,+0.0245]$} & \multicolumn{3}{r}{$+0.0578\,[+0.0296,+0.0869]$} \\
Max & \multicolumn{2}{r}{$-0.0594\,[-0.0918,-0.0257]$} & \multicolumn{3}{r}{$-0.0081\,[-0.0418,+0.0272]$} \\
\bottomrule
\end{tabular*}
\caption{\textbf{Expanded mathematics score anatomy.} These analyses reuse released DAPO/Omni trajectories. Panel (a) averages cells equally; `8B 8K' denotes the five DAPO arms. Positive and CI-positive counts refer to the displayed paired contrast. Panel (b) uses two-seed 8B Binary RL on DAPO 8K at $\alpha=0.065$; $R/W$ are percentages and differences are percentage points. The CAL-selected row is exploratory score selection. Panel (c) compares full-trace scores with the full-trace mean. Tail-$q$ averages the largest $q\%$ of token NLLs; Tail-H uses entropy. All intervals are recorded pointwise 95\% paired intervals.}\label{tab:math-stage1}
\end{table}

\subsection{Extended LiveCodeBench diagnostics, sensitivity, and 4B pilot}
\label{app:lcb-extended}

Table~\ref{tab:lcb-score-ranking} expands the code-generation comparison
across the three registered scoring spans. The answer span begins after
the first end-of-thinking delimiter; the code span is the extracted
Python program; the full trace includes both sampled thinking and answer
tokens. Scoring excludes prompt tokens, injected control tokens, and EOS.
Valid programs are graded against all official public and private tests;
wrong programs are categorized by their first observed failure as wrong
answer, runtime error, time limit, memory limit, or output limit.
Partial AUROC integrates correct acceptance up to wrong-answer
false-positive rates 0.10 and 0.20, with standardization mapping random
ranking to 0.5 and perfect ranking to 1.

Table~\ref{tab:lcb-score-anatomy} reports paired contrasts, NLL mass concentration, rank correlations, and first-failure types. All twelve Tail NLL-minus-Mean NLL AUROC intervals---base or two-seed Binary RL,
two caps, and three spans---include zero. In pure code, the highest 10\%
of token surprises contribute 96.09\% and 96.78\% of total NLL for
Binary RL at 4K and 8K. For a trajectory with positive total NLL, let
$\overline{\ell}=L^{-1}\sum_{i=1}^L\ell_i$ be its mean NLL,
$T_k(\ell)=k^{-1}\sum_{i=1}^k\ell_{(i)}$ the top-$k$ average, and
$M=\sum_{i=1}^k\ell_{(i)}/\sum_{i=1}^L\ell_i$ the tail-mass fraction,
where $k=\lceil0.1L\rceil$ and surprises are in decreasing order. Then
$\overline{\ell}=(k/L)T_k(\ell)/M$ exactly.
The concentration explains why the mean retains much of the same
signal. Since $(k/L)/M$ can vary between trajectories, neither the
identity nor the observed concentration establishes exact ranking
equivalence or zero background variance. This distinction connects the
code results to the limiting case of Corollary~\ref{cor:tail-gain-main}.

\begin{table}[p]
\centering
\footnotesize
\setlength{\tabcolsep}{2pt}
\renewcommand{\arraystretch}{1.12}
\begin{tabular*}{\linewidth}{@{\extracolsep{\fill}}lrrrrrrrrr@{}}
\toprule
& \multicolumn{3}{c}{Answer span} & \multicolumn{3}{c}{Code span} & \multicolumn{3}{c}{Full trace} \\
Score & AUC & $p_{.10}$ & $p_{.20}$ & AUC & $p_{.10}$ & $p_{.20}$ & AUC & $p_{.10}$ & $p_{.20}$ \\
\midrule
\multicolumn{10}{@{}l}{\textit{Base, 4K}}\\
Mean & 0.832 & 0.659 & 0.713 & 0.846 & 0.663 & 0.715 & 0.907 & 0.828 & 0.852 \\
Tail-2 & 0.836 & 0.658 & 0.709 & 0.839 & 0.648 & 0.705 & 0.926 & 0.846 & 0.868 \\
Tail-5 & 0.840 & 0.660 & 0.714 & 0.843 & 0.672 & 0.715 & 0.914 & 0.833 & 0.860 \\
Tail-10 & 0.838 & 0.660 & 0.716 & 0.846 & 0.668 & 0.717 & 0.915 & 0.832 & 0.861 \\
Top-16 & 0.843 & 0.689 & 0.736 & 0.861 & 0.690 & 0.753 & 0.940 & 0.875 & 0.890 \\
Top-32 & 0.844 & 0.688 & 0.736 & 0.861 & 0.687 & 0.752 & 0.933 & 0.844 & 0.864 \\
Max & 0.835 & 0.651 & 0.709 & 0.836 & 0.650 & 0.707 & 0.901 & 0.796 & 0.822 \\
Sum & 0.838 & 0.689 & 0.736 & 0.861 & 0.690 & 0.753 & 0.915 & 0.803 & 0.824 \\
Length & 0.822 & 0.676 & 0.707 & 0.836 & 0.682 & 0.719 & 0.831 & 0.719 & 0.734 \\
\addlinespace[6pt]
\multicolumn{10}{@{}l}{\textit{Base, 8K}}\\
Mean & 0.841 & 0.782 & 0.796 & 0.869 & 0.730 & 0.760 & 0.892 & 0.782 & 0.825 \\
Tail-2 & 0.842 & 0.784 & 0.799 & 0.868 & 0.726 & 0.755 & 0.896 & 0.791 & 0.829 \\
Tail-5 & 0.841 & 0.782 & 0.799 & 0.869 & 0.725 & 0.758 & 0.891 & 0.786 & 0.829 \\
Tail-10 & 0.844 & 0.782 & 0.798 & 0.871 & 0.731 & 0.764 & 0.894 & 0.788 & 0.831 \\
Top-16 & 0.852 & 0.802 & 0.823 & 0.890 & 0.750 & 0.781 & 0.883 & 0.765 & 0.803 \\
Top-32 & 0.852 & 0.800 & 0.821 & 0.890 & 0.749 & 0.780 & 0.889 & 0.786 & 0.825 \\
Max & 0.844 & 0.755 & 0.787 & 0.858 & 0.704 & 0.742 & 0.848 & 0.704 & 0.753 \\
Sum & 0.848 & 0.802 & 0.822 & 0.891 & 0.750 & 0.781 & 0.873 & 0.800 & 0.829 \\
Length & 0.818 & 0.712 & 0.748 & 0.854 & 0.729 & 0.729 & 0.831 & 0.759 & 0.782 \\
\addlinespace[6pt]
\multicolumn{10}{@{}l}{\textit{Binary RL, 4K}}\\
Mean & 0.834 & 0.733 & 0.766 & 0.849 & 0.730 & 0.757 & 0.877 & 0.789 & 0.817 \\
Tail-2 & 0.833 & 0.742 & 0.766 & 0.849 & 0.747 & 0.764 & 0.854 & 0.758 & 0.784 \\
Tail-5 & 0.835 & 0.738 & 0.763 & 0.852 & 0.740 & 0.762 & 0.865 & 0.773 & 0.799 \\
Tail-10 & 0.836 & 0.736 & 0.767 & 0.852 & 0.734 & 0.759 & 0.872 & 0.780 & 0.806 \\
Top-16 & 0.832 & 0.752 & 0.777 & 0.854 & 0.755 & 0.765 & 0.852 & 0.745 & 0.786 \\
Top-32 & 0.832 & 0.752 & 0.778 & 0.854 & 0.755 & 0.765 & 0.850 & 0.747 & 0.782 \\
Max & 0.820 & 0.769 & 0.783 & 0.848 & 0.758 & 0.784 & 0.806 & 0.701 & 0.739 \\
Sum & 0.829 & 0.752 & 0.778 & 0.853 & 0.755 & 0.765 & 0.866 & 0.777 & 0.808 \\
Length & 0.797 & 0.710 & 0.735 & 0.801 & 0.693 & 0.721 & 0.801 & 0.730 & 0.747 \\
\addlinespace[6pt]
\multicolumn{10}{@{}l}{\textit{Binary RL, 8K}}\\
Mean & 0.843 & 0.705 & 0.760 & 0.828 & 0.661 & 0.691 & 0.885 & 0.825 & 0.835 \\
Tail-2 & 0.843 & 0.704 & 0.757 & 0.817 & 0.628 & 0.678 & 0.878 & 0.814 & 0.832 \\
Tail-5 & 0.845 & 0.705 & 0.760 & 0.826 & 0.648 & 0.686 & 0.881 & 0.818 & 0.831 \\
Tail-10 & 0.845 & 0.705 & 0.761 & 0.827 & 0.652 & 0.687 & 0.885 & 0.813 & 0.829 \\
Top-16 & 0.850 & 0.718 & 0.775 & 0.846 & 0.657 & 0.709 & 0.863 & 0.795 & 0.813 \\
Top-32 & 0.850 & 0.717 & 0.775 & 0.848 & 0.662 & 0.711 & 0.870 & 0.803 & 0.823 \\
Max & 0.847 & 0.720 & 0.761 & 0.821 & 0.623 & 0.698 & 0.829 & 0.750 & 0.768 \\
Sum & 0.844 & 0.718 & 0.775 & 0.846 & 0.651 & 0.706 & 0.880 & 0.818 & 0.835 \\
Length & 0.833 & 0.728 & 0.770 & 0.822 & 0.670 & 0.700 & 0.846 & 0.780 & 0.801 \\
\bottomrule
\end{tabular*}
\caption{\textbf{LiveCodeBench ranking scores across three spans.} All entries use TEST300 and valid programs.
Binary RL averages the two seeds' ranking metrics. AUC denotes AUROC; $p_{.10}$ and $p_{.20}$ are McClish-standardized partial AUROC at wrong-answer FPR 0.10 and 0.20.
Tail-$q$ averages the largest $q\%$ of token NLLs; Top-$k$ averages the largest $k$. All nine prespecified scores are shown for both models, both caps, and all three spans. Paired contrasts and score diagnostics appear in Table~\ref{tab:lcb-score-anatomy}.}\label{tab:lcb-score-ranking}
\end{table}

\begin{table}[p]
\centering
\footnotesize
\setlength{\tabcolsep}{2pt}
\renewcommand{\arraystretch}{1.12}
\begin{tabular*}{\linewidth}{@{\extracolsep{\fill}}lrrrrrrrrr@{}}
\toprule
\multicolumn{10}{@{}l}{\textbf{(a) Tail-10 minus Mean: paired differences and 95\% intervals}}\\
Model / cap / span & \multicolumn{3}{r}{$\Delta$AUROC} & \multicolumn{3}{r}{$\Delta$pAUROC$_{.10}$} & \multicolumn{3}{r}{$\Delta$pAUROC$_{.20}$} \\
Base / 4K / answer & \multicolumn{3}{r}{$+0.005\,[-0.001,+0.012]$} & \multicolumn{3}{r}{$+0.001\,[-0.004,+0.010]$} & \multicolumn{3}{r}{$+0.004\,[-0.004,+0.010]$} \\
Base / 4K / code & \multicolumn{3}{r}{$+0.001\,[-0.005,+0.007]$} & \multicolumn{3}{r}{$+0.005\,[-0.008,+0.016]$} & \multicolumn{3}{r}{$+0.001\,[-0.007,+0.013]$} \\
Base / 4K / full & \multicolumn{3}{r}{$+0.008\,[-0.004,+0.022]$} & \multicolumn{3}{r}{$+0.004\,[-0.025,+0.042]$} & \multicolumn{3}{r}{$+0.009\,[-0.014,+0.036]$} \\
Base / 8K / answer & \multicolumn{3}{r}{$+0.003\,[-0.001,+0.008]$} & \multicolumn{3}{r}{$-0.001\,[-0.005,+0.005]$} & \multicolumn{3}{r}{$+0.002\,[-0.002,+0.005]$} \\
Base / 8K / code & \multicolumn{3}{r}{$+0.002\,[-0.001,+0.005]$} & \multicolumn{3}{r}{$+0.002\,[-0.005,+0.009]$} & \multicolumn{3}{r}{$+0.004\,[-0.002,+0.009]$} \\
Base / 8K / full & \multicolumn{3}{r}{$+0.003\,[-0.005,+0.011]$} & \multicolumn{3}{r}{$+0.006\,[-0.006,+0.020]$} & \multicolumn{3}{r}{$+0.007\,[-0.005,+0.017]$} \\
Binary RL / 4K / answer & \multicolumn{3}{r}{$+0.002\,[-0.002,+0.007]$} & \multicolumn{3}{r}{$+0.003\,[-0.006,+0.008]$} & \multicolumn{3}{r}{$+0.001\,[-0.006,+0.007]$} \\
Binary RL / 4K / code & \multicolumn{3}{r}{$+0.002\,[-0.002,+0.007]$} & \multicolumn{3}{r}{$+0.004\,[-0.005,+0.012]$} & \multicolumn{3}{r}{$+0.002\,[-0.004,+0.010]$} \\
Binary RL / 4K / full & \multicolumn{3}{r}{$-0.005\,[-0.017,+0.005]$} & \multicolumn{3}{r}{$-0.009\,[-0.031,+0.017]$} & \multicolumn{3}{r}{$-0.011\,[-0.027,+0.006]$} \\
Binary RL / 8K / answer & \multicolumn{3}{r}{$+0.002\,[-0.001,+0.005]$} & \multicolumn{3}{r}{$+0.000\,[-0.004,+0.003]$} & \multicolumn{3}{r}{$+0.001\,[-0.003,+0.003]$} \\
Binary RL / 8K / code & \multicolumn{3}{r}{$-0.001\,[-0.007,+0.004]$} & \multicolumn{3}{r}{$-0.009\,[-0.030,+0.005]$} & \multicolumn{3}{r}{$-0.004\,[-0.021,+0.005]$} \\
Binary RL / 8K / full & \multicolumn{3}{r}{$+0.000\,[-0.007,+0.008]$} & \multicolumn{3}{r}{$-0.012\,[-0.023,+0.001]$} & \multicolumn{3}{r}{$-0.005\,[-0.016,+0.006]$} \\
\addlinespace[6pt]
\multicolumn{10}{@{}l}{\textbf{(b) Binary RL: NLL mass concentration and ordering}}\\
Cap / span & \multicolumn{2}{r}{Valid mass (\%)} & \multicolumn{2}{r}{Correct mass (\%)} & \multicolumn{2}{r}{Wrong mass (\%)} & \multicolumn{3}{r}{Spearman $\rho$ (valid)} \\
4K / Answer & \multicolumn{2}{r}{92.55} & \multicolumn{2}{r}{94.39} & \multicolumn{2}{r}{81.70} & \multicolumn{3}{r}{0.9993} \\
4K / Code & \multicolumn{2}{r}{96.09} & \multicolumn{2}{r}{97.29} & \multicolumn{2}{r}{88.98} & \multicolumn{3}{r}{0.9991} \\
4K / Full & \multicolumn{2}{r}{69.18} & \multicolumn{2}{r}{71.12} & \multicolumn{2}{r}{57.72} & \multicolumn{3}{r}{0.9938} \\
8K / Answer & \multicolumn{2}{r}{92.37} & \multicolumn{2}{r}{94.96} & \multicolumn{2}{r}{78.66} & \multicolumn{3}{r}{0.9996} \\
8K / Code & \multicolumn{2}{r}{96.78} & \multicolumn{2}{r}{98.80} & \multicolumn{2}{r}{86.06} & \multicolumn{3}{r}{0.9990} \\
8K / Full & \multicolumn{2}{r}{64.45} & \multicolumn{2}{r}{66.64} & \multicolumn{2}{r}{52.87} & \multicolumn{3}{r}{0.9959} \\
\addlinespace[6pt]
\multicolumn{10}{@{}l}{\textbf{(c) Binary RL: first-failure types and code-span Tail-10}}\\
Cap / error & \multicolumn{2}{r}{Deployments} & \multicolumn{2}{r}{Families} & \multicolumn{2}{r}{Median} & \multicolumn{3}{r}{P10--P90} \\
4K / WA & \multicolumn{2}{r}{36} & \multicolumn{2}{r}{29} & \multicolumn{2}{r}{0.950} & \multicolumn{3}{r}{0.327--1.786} \\
4K / RE & \multicolumn{2}{r}{6} & \multicolumn{2}{r}{5} & \multicolumn{2}{r}{1.110} & \multicolumn{3}{r}{0.636--1.781} \\
4K / TLE & \multicolumn{2}{r}{7} & \multicolumn{2}{r}{6} & \multicolumn{2}{r}{0.343} & \multicolumn{3}{r}{0.146--1.385} \\
4K / MLE & \multicolumn{2}{r}{1} & \multicolumn{2}{r}{1} & \multicolumn{2}{r}{1.567} & \multicolumn{3}{r}{1.567--1.567} \\
4K / Output limit & \multicolumn{2}{r}{0} & \multicolumn{2}{r}{0} & \multicolumn{2}{r}{---} & \multicolumn{3}{r}{---} \\
8K / WA & \multicolumn{2}{r}{42} & \multicolumn{2}{r}{37} & \multicolumn{2}{r}{0.749} & \multicolumn{3}{r}{0.106--2.023} \\
8K / RE & \multicolumn{2}{r}{4} & \multicolumn{2}{r}{4} & \multicolumn{2}{r}{1.393} & \multicolumn{3}{r}{0.289--2.259} \\
8K / TLE & \multicolumn{2}{r}{24} & \multicolumn{2}{r}{17} & \multicolumn{2}{r}{0.742} & \multicolumn{3}{r}{0.070--2.254} \\
8K / MLE & \multicolumn{2}{r}{1} & \multicolumn{2}{r}{1} & \multicolumn{2}{r}{0.374} & \multicolumn{3}{r}{0.374--0.374} \\
8K / Output limit & \multicolumn{2}{r}{0} & \multicolumn{2}{r}{0} & \multicolumn{2}{r}{---} & \multicolumn{3}{r}{---} \\
\bottomrule
\end{tabular*}
\caption{\textbf{LiveCodeBench paired contrasts and score diagnostics.} All entries use TEST300 and valid programs. Panel (a) reports Tail-10 minus Mean NLL differences with pointwise 95\% family-paired intervals; Binary RL averages the two seeds' ranking metrics. Panel (b) gives the mean top-10\% NLL mass fraction for valid, correct, and wrong programs, and the Tail-10/Mean NLL Spearman correlation among valid programs. Panel (c) gives first-failure types for the two Binary RL seeds, with deployment counts and code-span Tail-10 quantiles: WA, wrong answer; RE, runtime error; TLE, time limit; MLE, memory limit. Counts pool deployments across seeds.}\label{tab:lcb-score-anatomy}
\end{table}

Table~\ref{tab:lcb-sensitivity} separates year composition, known errata,
calibration variability, and the model-selection pilot. The 2023/2024/2025
TEST strata contain 97/162/41 families and use the overall registered
calibration cuts. The errata analysis uses separately locked CAL294
thresholds and TEST299; both budgets retain positive paired intervals
for Binary RL's raw correct-submission and calibrated net-score gains
over base. The release ends on 2025-04-06, so the study does not establish
exclusion of pretraining contamination.

Repartitioning the fixed 600-family pool 2,000 times and recalibrating on
each CAL300 yields mean wrong-submission rates of 6.26\% and 6.34\%,
with standard deviations 1.52 and 1.67 pp at 4K and 8K. These describe
finite-pool variation around the target, whereas the original TEST300
rates describe one frozen split. The expectation in
Theorem~\ref{thm:crc} averages over calibration and deployment under
exchangeability. One-sided Clopper--Pearson bounds apply to each
single-seed Bernoulli test panel; the two-seed family averages are not
a binomial sample and are reported without a pooled binomial bound.

The disjoint SMOKE20 pilot explains the focus on 8B in the formal study.
Qwen3.5-4B produced three/six correct programs at 4K/8K, no wrong programs,
and 17/14 truncations; its Native 4K replays all truncated. The absence
of wrong programs leaves no error-ranking comparison in that pilot.
At 8K, its generation cost was \$0.00905 per family versus \$0.00407
for the 8B base model. The formal panel therefore concentrated its
budget on 8B, whose pilot supplied both correct and wrong programs.
These pilot counts remain separate from the independent TEST300 results.

\begin{table}[p]
\centering
\footnotesize
\setlength{\tabcolsep}{2pt}
\renewcommand{\arraystretch}{1.13}
\begin{tabular*}{\linewidth}{@{\extracolsep{\fill}}lrrrrrrr@{}}
\toprule
\multicolumn{8}{@{}l}{\textbf{(a) Year and errata sensitivity}}\\
TEST subset & Cap & $n$ & Raw $R$ & Raw $W$ & CRC $R$ & CRC $W$ & CRC Net \\
All TEST & 4K & 300 & 49.17 & 8.33 & 48.50 & 7.33 & 41.17 \\
All TEST & 8K & 300 & 62.83 & 11.83 & 54.67 & 4.83 & 49.83 \\
2023 & 4K & 97 & 65.98 & 5.67 & 64.95 & 5.67 & 59.28 \\
2023 & 8K & 97 & 80.93 & 7.73 & 69.59 & 4.64 & 64.95 \\
2024 & 4K & 162 & 44.14 & 9.57 & 43.52 & 8.02 & 35.49 \\
2024 & 8K & 162 & 58.64 & 12.35 & 51.85 & 4.63 & 47.22 \\
2025 & 4K & 41 & 29.27 & 9.76 & 29.27 & 8.54 & 20.73 \\
2025 & 8K & 41 & 36.59 & 19.51 & 30.49 & 6.10 & 24.39 \\
Errata excluded & 4K & 299 & 49.33 & 8.36 & 48.66 & 7.86 & 40.80 \\
Errata excluded & 8K & 299 & 62.88 & 11.71 & 55.85 & 5.02 & 50.84 \\
\addlinespace[3pt]
\multicolumn{8}{@{}l}{\textbf{(b) Binary RL minus Base: paired transfer contrasts}}\\
TEST subset & Cap & \multicolumn{2}{r}{$\Delta$Raw $R$ (pp)} & \multicolumn{2}{r}{$\Delta$CRC Net (pp)} & \multicolumn{2}{r}{$\Delta$USD/family} \\
All TEST & 4K & \multicolumn{2}{r}{$+7.17\,[+3.00,+11.33]$} & \multicolumn{2}{r}{$+6.50\,[+1.00,+11.84]$} & \multicolumn{2}{r}{$-0.00006\,[-0.00011,-0.00001]$} \\
All TEST & 8K & \multicolumn{2}{r}{$+4.50\,[+1.33,+7.83]$} & \multicolumn{2}{r}{$+5.83\,[+0.50,+11.33]$} & \multicolumn{2}{r}{$-0.00017\,[-0.00026,-0.00008]$} \\
Errata excluded & 4K & \multicolumn{2}{r}{$+7.19\,[+3.01,+11.37]$} & \multicolumn{2}{r}{$+5.69\,[+0.17,+11.04]$} & \multicolumn{2}{r}{$-0.00006\,[-0.00011,-0.00001]$} \\
Errata excluded & 8K & \multicolumn{2}{r}{$+4.35\,[+1.17,+7.53]$} & \multicolumn{2}{r}{$+6.35\,[+1.00,+12.04]$} & \multicolumn{2}{r}{$-0.00017\,[-0.00026,-0.00007]$} \\
\addlinespace[3pt]
\multicolumn{8}{@{}l}{\textbf{(c) Repeated calibration/test splits: answer-span Tail-10}}\\
Solver & Cap & $\alpha$ (\%) & TEST $W$ & Mean $W$ & SD (pp) & Draws & CAL/TEST $n$ \\
Base & 4K & 5.0 & 4.33 & 4.98 & 1.77 & 2000 & 300/300 \\
Base & 4K & 6.5 & 6.33 & 6.31 & 1.96 & 2000 & 300/300 \\
Base & 4K & 10.0 & 9.33 & 8.72 & 1.33 & 2000 & 300/300 \\
Base & 8K & 5.0 & 5.00 & 5.03 & 1.80 & 2000 & 300/300 \\
Base & 8K & 6.5 & 6.33 & 6.38 & 1.99 & 2000 & 300/300 \\
Base & 8K & 10.0 & 8.33 & 10.02 & 2.37 & 2000 & 300/300 \\
Binary RL & 4K & 5.0 & 4.50 & 4.99 & 1.42 & 2000 & 300/300 \\
Binary RL & 4K & 6.5 & 7.33 & 6.26 & 1.52 & 2000 & 300/300 \\
Binary RL & 4K & 10.0 & 8.33 & 8.02 & 0.98 & 2000 & 300/300 \\
Binary RL & 8K & 5.0 & 4.17 & 5.01 & 1.52 & 2000 & 300/300 \\
Binary RL & 8K & 6.5 & 4.83 & 6.34 & 1.67 & 2000 & 300/300 \\
Binary RL & 8K & 10.0 & 7.33 & 10.04 & 1.99 & 2000 & 300/300 \\
\addlinespace[3pt]
\multicolumn{8}{@{}l}{\textbf{(d) Disjoint SMOKE20 model-selection pilot}}\\
Model / solver & Cap & $n$ & $R$ & $W$ & $U$ & Truncated & USD/family \\
4B Base & 4K & 20 & 3 & 0 & 17 & 17 & 0.00475 \\
4B Base & 8K & 20 & 6 & 0 & 14 & 14 & 0.00905 \\
8B Base & 4K & 20 & 6 & 0 & 14 & 14 & 0.00274 \\
8B Base & 8K & 20 & 9 & 5 & 6 & 6 & 0.00407 \\
8B Binary RL & 4K & 20 & 7 & 2 & 11 & 10 & 0.00242 \\
8B Binary RL & 8K & 20 & 10 & 2 & 8 & 8 & 0.00419 \\
\bottomrule
\end{tabular*}
\caption{\textbf{LiveCodeBench sensitivity, calibration variation, and model-selection pilot.} Panels (a)--(b) use two-seed Binary RL at $\alpha=0.065$ with answer-span Tail-10; year strata retain the registered overall calibration cuts. The errata-excluded analysis uses its separately locked CAL294 cuts and TEST299. Panel (c) repartitions the fixed 600-family pool into CAL300/TEST300 2,000 times; SD is in percentage points. Panel (d) contains the disjoint SMOKE20 convenience sample; counts are descriptive and all six conditions use budget forcing.}\label{tab:lcb-sensitivity}
\end{table}

\clearpage
\section{Exploratory A100 pilots}
\label{app:audited-pilots}

All results in this appendix are small-sample exploratory checks. They do
not enter the main empirical claims or confirmatory gates. The local
training configuration uses four families and four rollouts per family
per update, in contrast to the mainline 16-by-eight regimen. The completed
Qwen runs train 4B G0/G1 and 8B G0/G0+KL/G0-multibudget for 12 updates,
192 trajectories, and 48 family exposures per arm. The KL coefficient is
.02 with the frozen base distribution; the multibudget arm alternates
4K and 8K by update. DeepSeek-R1-Distill-Llama-8B has two G0 seeds and
one G1 seed, each trained for 25 updates, 400 trajectories, and 100 family
exposures. Earlier saved tiers are four and 12 updates; the planned
50-update tier was not completed. Final matched evaluation tiers have
16 CAL and 16 TEST families per domain; earlier tiers contain four.
These exposures, model initializations, and serving paths differ from
the Tinker mainline. They cannot establish cross-family replication,
a reliable KL effect, or the benefit of multibudget training.
The nineteen archived checkpoints total 6,155,082,568 bytes.

\subsection{Operating points from the completed local-training tiers}
\label{app:pilot-training-points}

Tables~\ref{tab:pilot-local-extbf} and~\ref{tab:pilot-local-native}
report the final completed local checkpoints: 12 updates for the Qwen arms
and 25 for DeepSeek-R1-Distill-Llama-8B. Every row has 16 matched TEST
families and is calibrated on the corresponding 16 CAL families. The Qwen checkpoints have
192 training rollouts over 48 family exposures; the Llama checkpoints
have 400 rollouts over 100 exposures. These locally trained models are
distinct from the archived 50-update Tinker policies. All 32 final-tier ExtBF cells and all 12 final-tier Native Llama cells
are included; earlier saved tiers remain separate in the raw artifact.

The reference for a G0 row is the same base model. The reference for G1,
G0+KL, or G0+multi is local G0$_{41}$ at the same update tier, decoder,
domain, and cap. Paired differences use the same 16 TEST families, even
if the reference has a larger generation roster. The supplied pointwise
family-bootstrap intervals condition on fitted CAL thresholds. A single changed outcome moves a rate by 6.25 pp, and a
degenerate interval on an unchanged small panel does not imply zero
population effect. The intervals exclude training-seed variation and are not corrected
for multiple comparisons.

\begin{table*}[p]
\centering\scriptsize
\setlength{\tabcolsep}{4pt}
\begingroup%
\renewcommand{\arraystretch}{1.10}%
\setlength{\tabcolsep}{4pt}%
\footnotesize%
\begin{tabular}{@{}llrllC{2.15em}C{2.15em}cC{2.15em}C{2.15em}r@{}}
\toprule
\multirow{2}{*}{Model} & \multirow{2}{*}{Local arm} & \multirow{2}{*}{Updates} & \multirow{2}{*}{Domain} & \multirow{2}{*}{Cap}
& \multicolumn{2}{c}{\shortstack{Raw\\counts}} & \hspace{9pt} & \multicolumn{2}{c}{\shortstack{Tail-CRC\\counts}} & \multirow{2}{*}{\shortstack{Raw $\Delta R$\\(pp; 95\% CI)}} \\
\cmidrule(lr){6-7}\cmidrule(lr){9-10}
& & & & & $R$ & $W$ & & $R$ & $W$ & \\
\midrule
4B & G0$_{41}$ & 12 & DAPO & 4K & 0 & 0 &  & 0 & 0 & +0.00 [0.00, 0.00] \\
4B & G0$_{41}$ & 12 & DAPO & 8K & 5 & 0 &  & 5 & 0 & +6.25 [-18.75, 31.25] \\
4B & G0$_{41}$ & 12 & Omni & 4K & 1 & 0 &  & 0 & 0 & +0.00 [0.00, 0.00] \\
4B & G0$_{41}$ & 12 & Omni & 8K & 1 & 0 &  & 0 & 0 & -6.25 [-18.75, 0.00] \\
\addlinespace[3pt]
4B & G1$_{41}$ & 12 & DAPO & 4K & 1 & 0 &  & 1 & 0 & +6.25 [0.00, 18.75] \\
4B & G1$_{41}$ & 12 & DAPO & 8K & 3 & 0 &  & 3 & 0 & -12.50 [-37.50, 18.75] \\
4B & G1$_{41}$ & 12 & Omni & 4K & 1 & 0 &  & 1 & 0 & +0.00 [0.00, 0.00] \\
4B & G1$_{41}$ & 12 & Omni & 8K & 1 & 0 &  & 0 & 0 & +0.00 [0.00, 0.00] \\
\addlinespace[3pt]
8B & G0$_{41}$ & 12 & DAPO & 4K & 2 & 2 &  & 2 & 1 & -6.25 [-25.00, 12.50] \\
8B & G0$_{41}$ & 12 & DAPO & 8K & 7 & 1 &  & 5 & 0 & +0.00 [-25.00, 25.00] \\
8B & G0$_{41}$ & 12 & Omni & 4K & 3 & 5 &  & 0 & 0 & +6.25 [-12.50, 25.00] \\
8B & G0$_{41}$ & 12 & Omni & 8K & 2 & 6 &  & 0 & 0 & -31.25 [-56.25, -12.50] \\
\addlinespace[3pt]
8B & G0+KL & 12 & DAPO & 4K & 2 & 1 &  & 2 & 1 & +0.00 [-25.00, 25.00] \\
8B & G0+KL & 12 & DAPO & 8K & 9 & 1 &  & 5 & 0 & +12.50 [-12.50, 37.50] \\
8B & G0+KL & 12 & Omni & 4K & 0 & 3 &  & 0 & 0 & -18.75 [-37.50, 0.00] \\
8B & G0+KL & 12 & Omni & 8K & 6 & 6 &  & 0 & 1 & +25.00 [6.25, 43.91] \\
\addlinespace[3pt]
8B & G0+multi & 12 & DAPO & 4K & 0 & 2 &  & 0 & 0 & -12.50 [-31.25, 0.00] \\
8B & G0+multi & 12 & DAPO & 8K & 9 & 0 &  & 8 & 0 & +12.50 [-18.75, 43.75] \\
8B & G0+multi & 12 & Omni & 4K & 2 & 3 &  & 1 & 0 & -6.25 [-31.25, 18.75] \\
8B & G0+multi & 12 & Omni & 8K & 4 & 6 &  & 1 & 0 & +12.50 [0.00, 31.25] \\
\addlinespace[3pt]
Llama & G0$_{41}$ & 25 & DAPO & 4K & 1 & 0 &  & 1 & 0 & -6.25 [-18.75, 0.00] \\
Llama & G0$_{41}$ & 25 & DAPO & 8K & 11 & 2 &  & 3 & 0 & +37.50 [12.50, 62.50] \\
Llama & G0$_{41}$ & 25 & Omni & 4K & 0 & 3 &  & 0 & 2 & -6.25 [-18.75, 0.00] \\
Llama & G0$_{41}$ & 25 & Omni & 8K & 2 & 4 &  & 0 & 0 & +6.25 [0.00, 18.75] \\
\addlinespace[3pt]
Llama & G0$_{73}$ & 25 & DAPO & 4K & 4 & 1 &  & 1 & 0 & +12.50 [-18.75, 43.75] \\
Llama & G0$_{73}$ & 25 & DAPO & 8K & 8 & 2 &  & 0 & 0 & +18.75 [-6.25, 43.75] \\
Llama & G0$_{73}$ & 25 & Omni & 4K & 1 & 2 &  & 0 & 0 & +0.00 [0.00, 0.00] \\
Llama & G0$_{73}$ & 25 & Omni & 8K & 2 & 4 &  & 0 & 1 & +6.25 [0.00, 18.75] \\
\addlinespace[3pt]
Llama & G1$_{41}$ & 25 & DAPO & 4K & 3 & 2 &  & 0 & 1 & +12.50 [-12.50, 37.50] \\
Llama & G1$_{41}$ & 25 & DAPO & 8K & 4 & 3 &  & 0 & 0 & -43.75 [-68.75, -18.75] \\
Llama & G1$_{41}$ & 25 & Omni & 4K & 0 & 2 &  & 0 & 0 & +0.00 [0.00, 0.00] \\
Llama & G1$_{41}$ & 25 & Omni & 8K & 2 & 3 &  & 1 & 0 & +0.00 [-18.75, 18.75] \\
\bottomrule
\end{tabular}%
\endgroup%

\caption{Exploratory A3 local-training ExtBF results, CAL16/TEST16.
$R/W$ entries are integer correct/wrong counts out of 16; remaining
outcomes are $U$. The Tail-CRC columns use \texttt{tail10\_nll} and
$\alpha=.065$. The final column is the raw correct-submission difference
in pp with its supplied 95\% interval. G0 rows compare with base; all
other rows compare with same-tier local G0$_{41}$. ``Llama'' denotes
DeepSeek-R1-Distill-Llama-8B. All rows are exploratory and remain outside
the main empirical claims.}
\label{tab:pilot-local-extbf}
\end{table*}

\begin{table*}[t]
\centering\scriptsize
\setlength{\tabcolsep}{4pt}
\begingroup%
\renewcommand{\arraystretch}{1.10}%
\setlength{\tabcolsep}{4pt}%
\footnotesize%
\begin{tabular}{@{}llrllC{2.15em}C{2.15em}cC{2.15em}C{2.15em}r@{}}
\toprule
\multirow{2}{*}{Model} & \multirow{2}{*}{Local arm} & \multirow{2}{*}{Updates} & \multirow{2}{*}{Domain} & \multirow{2}{*}{Cap}
& \multicolumn{2}{c}{\shortstack{Raw\\counts}} & \hspace{9pt} & \multicolumn{2}{c}{\shortstack{Tail-CRC\\counts}} & \multirow{2}{*}{\shortstack{Raw $\Delta R$\\(pp; 95\% CI)}} \\
\cmidrule(lr){6-7}\cmidrule(lr){9-10}
& & & & & $R$ & $W$ & & $R$ & $W$ & \\
\midrule
Llama & G0$_{41}$ & 25 & DAPO & 4K & 4 & 1 &  & 2 & 0 & +18.75 [0.00, 37.50] \\
Llama & G0$_{41}$ & 25 & DAPO & 8K & 7 & 1 &  & 6 & 1 & +12.50 [-12.50, 37.50] \\
Llama & G0$_{41}$ & 25 & Omni & 4K & 1 & 0 &  & 0 & 0 & +0.00 [0.00, 0.00] \\
Llama & G0$_{41}$ & 25 & Omni & 8K & 3 & 2 &  & 0 & 0 & +0.00 [-18.75, 18.75] \\
\addlinespace[3pt]
Llama & G0$_{73}$ & 25 & DAPO & 4K & 3 & 1 &  & 3 & 1 & +12.50 [0.00, 31.25] \\
Llama & G0$_{73}$ & 25 & DAPO & 8K & 6 & 4 &  & 3 & 1 & +6.25 [-12.50, 25.00] \\
Llama & G0$_{73}$ & 25 & Omni & 4K & 1 & 0 &  & 0 & 0 & +0.00 [0.00, 0.00] \\
Llama & G0$_{73}$ & 25 & Omni & 8K & 4 & 3 &  & 1 & 0 & +6.25 [0.00, 18.75] \\
\addlinespace[3pt]
Llama & G1$_{41}$ & 25 & DAPO & 4K & 3 & 1 &  & 3 & 1 & -6.25 [-25.00, 12.50] \\
Llama & G1$_{41}$ & 25 & DAPO & 8K & 7 & 2 &  & 4 & 1 & +0.00 [-31.25, 31.25] \\
Llama & G1$_{41}$ & 25 & Omni & 4K & 1 & 0 &  & 0 & 0 & +0.00 [0.00, 0.00] \\
Llama & G1$_{41}$ & 25 & Omni & 8K & 1 & 1 &  & 1 & 0 & -12.50 [-31.25, 0.00] \\
\bottomrule
\end{tabular}%
\endgroup%

\caption{Exploratory A3 Native-decoding checks of the same 25-update
Llama policies, CAL16/TEST16. Units, references, and interval conventions
match Table~\ref{tab:pilot-local-extbf}. These local-training pilot results remain separate from the mainline
Native-transfer evaluation in Table~\ref{tab:native-transfer}.}
\label{tab:pilot-local-native}
\end{table*}

At CAL size 16, both targets .065 and .10 permit zero calibration wrong
submissions under the corrected CRC rule; all displayed .10 cuts and
operating points therefore equal their .065 counterparts. CAL size four
in the earliest tier cannot support a nontrivial fit at either target and
uses the all-reject fallback. The observed policies illustrate the limits
of these small panels. For example, the 8B KL arm at Omni 8K gains four
raw correct submissions over local G0, but its displayed calibrated subset
has zero correct and one wrong submission. Neither this isolated contrast
nor the local third-family outcomes establish a general KL effect,
multibudget benefit, or cross-family replication.

The exact run identifiers and counts are in
\nolinkurl{generation_summary.json}, \nolinkurl{generation_crc.json}, and
\nolinkurl{mechanism_contrasts.json} under the A100 supplement's results
directory. The training exposure and saved-tier index are in
\nolinkurl{training_summary.json}. The audit joins these records by complete
run key and retains the raw per-cell contrasts in the accompanying
numerical ledger.

\FloatBarrier
\subsection{Two-draw versus single-long-rollout pilot}

The two-draw pilot compares two 4K generations with a single 8K generation
on 16 TEST families per model and domain. \texttt{Maj@2} resolves a tie
using the first draw; strict agreement is a distinct policy that submits
only on agreement. Counts below use the same 16 families in each row.

\begin{table*}[ht]
\centering\small
\begin{tabular}{llrrrrr}
\toprule
Model & Domain & Maj@2 $R/W$ & Agreement $R/W$ & Single 8K $R/W$ & Two 4K tokens & One 8K tokens\\
\midrule
4B & DAPO & 7/1 & 5/0 & 11/0 & 115,909 & 99,950\\
4B & Omni & 1/0 & 1/0 & 4/2 & 126,562 & 115,965\\
8B & DAPO & 3/2 & 0/0 & 7/3 & 122,805 & 97,027\\
8B & Omni & 2/9 & 1/2 & 5/8 & 110,891 & 88,849\\
\bottomrule
\end{tabular}
\caption{Exploratory two-draw comparison only, $n=16$ per row. Token totals
are observed sampled output totals. Agreement with no submissions has
undefined conditional error.}
\label{tab:audited-voting-pilot}
\end{table*}

In these four small panels, the single 8K run uses fewer sampled tokens
than the pair of 4K runs, by 13.8\%, 8.4\%, 21.0\%, and 19.9\%,
respectively, relative to the two-draw token total. These observations do
not compare prefix-shared branching, establish a general ranking of
multi-sample reasoning, or estimate a population risk guarantee. The
complete checkpoint/exposure index and pointwise local contrasts are
preserved in the evidence bundle.

\clearpage
\section{Matched qualitative cases}
\label{app:audited-cases}

The following twelve cases preserve the six matched-case categories in the
released evidence. They are selected illustrations, not a random sample or
an estimate of category prevalence. Every displayed passage is a contiguous
verbatim substring of the archived terminal output; full questions, endings,
family identifiers, and numerical metadata are in
\texttt{data/matched\_cases.json}. Scores and cutoffs are rounded only for
display; acceptance uses the stored full-precision values and strict
$S<\tau$. A case-specific cutoff applies to the primary observed arm only.
Paired responses document raw behavior and are not evaluated against a
different model's cutoff. Invalid or truncated responses are always
rejected. The reference answer and tolerant $R/W/U$ label determine the
recorded outcome; an excerpt does not independently validate every step of
the generated argument. All examples use the 4K cap. For the Omni G0 cases,
the shown cuts are the transferred DAPO cuts at $\alpha=.065$.


\subsection{Equivalent boxes and strict-format repair}

The matching rule yields 19 eligible records; two selected cases illustrate this category.

\paragraph{Case 1: A convergent series.}\label{case:audited-01}\mbox{}\par\noindent

{\footnotesize \textbf{Family:} \nolinkurl{dapo_en_graph_v1:4ddd87cf40702f9a4f29f50f642722ca0e748f9c1ec2105cd487319ff59e791a}.}\par\noindent

\textbf{Setting:} DAPO, 8B, 4K; reference \verb|9|.

\textbf{Observed:} \texttt{M5\_41}, R, strict label \texttt{correct}, 2,945 sampled output tokens.

Box NLL $4.68454\times10^{-10}$; tail10 0.787890; answer-mean NLL 0.084767.

At the observed-arm cutoffs, box $1.95738\times10^{-6}$ gives \emph{accept}; tail 2.098790 gives \emph{accept}.

\begin{Verbatim}[fontsize=\scriptsize,breaklines=true,breakanywhere=true]
m + n = 4 + 5 = \boxed{9}.
$$<|im_end|>
\end{Verbatim}

\textbf{Paired response:} \texttt{base}, R, strict label \texttt{native\_format}, 3,675 sampled output tokens.

Box NLL $4.01586\times10^{-8}$; tail10 1.140496; answer-mean NLL 0.138860.

The paired response is a valid raw submission; the observed-arm cutoffs are not its selection policy.

\begin{Verbatim}[fontsize=\scriptsize,breaklines=true,breakanywhere=true]
Thus, $ m = 4 $, $ n = 5 $, and:

$$
m + n = 4 + 5 = \boxed{9}
$$<|im_end|>
\end{Verbatim}

Both policies produce the reference value 9. The base trace contains two equivalent boxes, whereas M5 contains one. The change repairs strict format without changing tolerant correctness on this family.

\paragraph{Case 2: Removing 2,013 piles of coins.}\label{case:audited-02}\mbox{}\par\noindent

{\footnotesize \textbf{Family:} \nolinkurl{dapo_en_graph_v1:55610eb3c129ca0c99ceb06098af7f7d5e8e9aa0678cda4062af0db0c0ef2294}.}\par\noindent

\textbf{Setting:} DAPO, 8B, 4K; reference \verb|11|.

\textbf{Observed:} \texttt{M5\_41}, R, strict label \texttt{correct}, 2,744 sampled output tokens.

Box NLL $1.57462\times10^{-8}$; tail10 2.044501; answer-mean NLL 0.357463.

At the observed-arm cutoffs, box $1.95738\times10^{-6}$ gives \emph{accept}; tail 2.098790 gives \emph{accept}.

\begin{Verbatim}[fontsize=\scriptsize,breaklines=true,breakanywhere=true]
\boxed{11}
$$<|im_end|>
\end{Verbatim}

\textbf{Paired response:} \texttt{base}, R, strict label \texttt{native\_format}, 2,674 sampled output tokens.

Box NLL $1.86570\times10^{-6}$; tail10 2.372422; answer-mean NLL 0.447950.

The paired response is a valid raw submission; the observed-arm cutoffs are not its selection policy.

\begin{Verbatim}[fontsize=\scriptsize,breaklines=true,breakanywhere=true]
- This ensures that we remove all the coins from all piles, in exactly $ \boxed{11} $ steps.

---

### Final Answer

$$
\boxed{11}
$$<|im_end|>
\end{Verbatim}

Both responses give the correct value 11. The base ending itself contains two equivalent boxes; M5 retains one. This matched example distinguishes strict-format correctness from mathematical correctness.

\subsection{G1 truncation with a matched G0 completion}

The matching rule yields 84 eligible records; two selected cases illustrate this category.

\paragraph{Case 3: Equal-sum partitions.}\label{case:audited-03}\mbox{}\par\noindent

{\footnotesize \textbf{Family:} \nolinkurl{dapo_en_graph_v1:1b024cf5343d2745d987d983124e654969bdc9187059abd891589369d8604c2d}.}\par\noindent

\textbf{Setting:} DAPO, 8B, 4K; reference \verb|6|.

\textbf{Observed:} \texttt{G1\_41}, U, strict label \texttt{truncated}, 4,092 sampled output tokens.

Box NLL --; tail10 --; answer-mean NLL --.

At the observed-arm cutoffs, box $8.70927\times10^{-6}$ gives \emph{reject (invalid)}; tail 2.338914 gives \emph{reject (invalid)}.

\begin{Verbatim}[fontsize=\scriptsize,breaklines=true,breakanywhere=true]
Start with the largest number, 20. To reach 42, need 22 more. Let me try 20 + 19 + 3 = 42. Let's check: 20 + 19 = 39 + 3 = 42. So subset1: {20,19,3}. Then next, take 18. Need 42 - 18 = 24. Let's see, 17 + 
\end{Verbatim}

\textbf{Paired response:} \texttt{G0\_41}, R, strict label \texttt{correct}, 3,511 sampled output tokens.

Box NLL $3.81940\times10^{-7}$; tail10 2.247711; answer-mean NLL 0.347004.

The paired response is a valid raw submission; the observed-arm cutoffs are not its selection policy.

\begin{Verbatim}[fontsize=\scriptsize,breaklines=true,breakanywhere=true]
All six of the candidate values $ \{21, 30, 35, 42, 70, 105\} $ satisfy the conditions for being good numbers.

Thus, the total number of **good numbers** is:

$$
\boxed{6}
$$<|im_end|>
\end{Verbatim}

G1 is interrupted before a valid submission; the matched G0 response submits the reference answer 6. This raw comparison does not apply the G1 calibration cutoff to the paired G0 response.

\paragraph{Case 4: A nested-floor equation.}\label{case:audited-04}\mbox{}\par\noindent

{\footnotesize \textbf{Family:} \nolinkurl{dapo_en_graph_v1:09b1682330e40d8e885523c99d238004f744e54f84db04853f1b4bba25b77d07}.}\par\noindent

\textbf{Setting:} DAPO, 8B, 4K; reference \verb|50|.

\textbf{Observed:} \texttt{G1\_41}, U, strict label \texttt{truncated}, 4,092 sampled output tokens.

Box NLL --; tail10 --; answer-mean NLL --.

At the observed-arm cutoffs, box $8.70927\times10^{-6}$ gives \emph{reject (invalid)}; tail 2.338914 gives \emph{reject (invalid)}.

\begin{Verbatim}[fontsize=\scriptsize,breaklines=true,breakanywhere=true]
Wait, but let me check for example, if x is in [250, 300). Let me take x=250: floor(x/100)=2. Then compute floor( (x/100)*floor(x/100) )= floor( (2.5)*2 )= floor(5)=5. Correct. Similarly x=299: floor(2.99*2)=floor(5.98)=5. Correct. What about x=250.5? Wait, but x is integer, so no. So all integers from 250 to 299 inclusive. So total number is 29
\end{Verbatim}

\textbf{Paired response:} \texttt{G0\_41}, R, strict label \texttt{correct}, 3,052 sampled output tokens.

Box NLL $4.87780\times10^{-7}$; tail10 1.891537; answer-mean NLL 0.278406.

The paired response is a valid raw submission; the observed-arm cutoffs are not its selection policy.

\begin{Verbatim}[fontsize=\scriptsize,breaklines=true,breakanywhere=true]
299 - 250 + 1 = 50.
$$

---

### Final Answer:

$$
\boxed{50}
$$<|im_end|>
\end{Verbatim}

G1 reaches the relevant integer interval but remains unfinished at the cap. The submitted answer in the paired G0 response is 50. The incomplete numeral at the end of G1 is retained verbatim and is labeled truncation, not a submitted wrong answer.

\subsection{Non-bare correct answers recovered by the tail score}

The matching rule yields 29 eligible records; two selected cases illustrate this category.

\paragraph{Case 5: A repeated fractional transformation.}\label{case:audited-05}\mbox{}\par\noindent

{\footnotesize \textbf{Family:} \nolinkurl{omni_lexical_v1:28472926d620b567f85b75fbafe5ac15baa6c0c603302aa47a1655245d8ba238}.}\par\noindent

\textbf{Setting:} Omni, 8B, 4K; reference \verb|-2-\sqrt{3}|.

\textbf{Observed:} \texttt{G0\_41}, R, strict label \texttt{native\_format}, 3,200 sampled output tokens.

Box NLL 0.642978; tail10 1.285988; answer-mean NLL 0.142226.

At the observed-arm cutoffs, box $2.20845\times10^{-6}$ gives \emph{reject}; tail 1.837020 gives \emph{accept}.

\begin{Verbatim}[fontsize=\scriptsize,breaklines=true,breakanywhere=true]
\boxed{-(2 + \sqrt{3})}
$$<|im_end|>
\end{Verbatim}

The algebraically equivalent radical answer is correct under the tolerant verifier. The source-DAPO box cutoff rejects it, whereas the tail cutoff accepts it.

\paragraph{Case 6: A coupled cubic system.}\label{case:audited-06}\mbox{}\par\noindent

{\footnotesize \textbf{Family:} \nolinkurl{omni_lexical_v1:ae5050cba8e9df7c29fba5e17f62f38306e32ccbad1895b89c23bcbdfb8fecc0}.}\par\noindent

\textbf{Setting:} Omni, 8B, 4K; reference \verb|(2, 4, 6)|.

\textbf{Observed:} \texttt{G0\_41}, R, strict label \texttt{correct}, 3,839 sampled output tokens.

Box NLL 2.021675; tail10 1.223633; answer-mean NLL 0.134865.

At the observed-arm cutoffs, box $2.20845\times10^{-6}$ gives \emph{reject}; tail 1.837020 gives \emph{accept}.

\begin{Verbatim}[fontsize=\scriptsize,breaklines=true,breakanywhere=true]
\boxed{(x, y, z) = (2, 4, 6)}
$$<|im_end|>
\end{Verbatim}

Transferred box and tail rules respectively reject and accept the correct tuple. These cases describe the observed non-bare-reference stratum without establishing universal score dominance.

\subsection{4B completion rescues}

The matching rule yields 127 eligible records; two selected cases illustrate this category.

\paragraph{Case 7: An intersection problem.}\label{case:audited-07}\mbox{}\par\noindent

{\footnotesize \textbf{Family:} \nolinkurl{dapo_en_graph_v1:a18197216a59537ff861f693b604c830b65fcb2ed43f29f72c27797e22013a17}.}\par\noindent

\textbf{Setting:} DAPO, 4B, 4K; reference \verb|13|.

\textbf{Observed:} \texttt{G0\_41}, R, strict label \texttt{correct}, 2,585 sampled output tokens.

Box NLL $7.28258\times10^{-5}$; tail10 1.810905; answer-mean NLL 0.232303.

At the observed-arm cutoffs, box $\infty$ gives \emph{accept}; tail $\infty$ gives \emph{accept}.

\begin{Verbatim}[fontsize=\scriptsize,breaklines=true,breakanywhere=true]
If the original question asked for the sum \(a+b\):
\[ a + b = \frac{1}{2} + \frac{7}{4} = \frac{2}{4} + \frac{7}{4} = \frac{9}{4} \]

The value \(\frac{9}{4}\) is in the form \(\frac{k}{m}\) with \(k=9\) and \(m=4\).
We are asked to find \(k+m\):
\[ k + m = 9 + 4 = 13 \]

The value of \(k+m\) is 13.

\boxed{13}<|im_end|>
\end{Verbatim}

\textbf{Paired response:} \texttt{base}, U, strict label \texttt{truncated}, 4,092 sampled output tokens.

Box NLL --; tail10 --; answer-mean NLL --.

The paired response is rejected as invalid independently of any score cutoff.

\begin{Verbatim}[fontsize=\scriptsize,breaklines=true,breakanywhere=true]
The point $(1,2)$ is in the first quadrant.
$a = 1/2 > 0$, $b = 7/4 >
\end{Verbatim}

G0 submits the reference value while base is truncated. The supplied question does not explicitly identify the original rational quantity; G0 infers it to be a+b. The excerpt therefore supports the recorded completion outcome, not an additional claim that the ambiguity was resolved rigorously.

\paragraph{Case 8: Recovering a polynomial.}\label{case:audited-08}\mbox{}\par\noindent

{\footnotesize \textbf{Family:} \nolinkurl{dapo_en_graph_v1:7aa8e2c744201cb61f8e346869d4d61202c1582d7f49ab2e3abfdc06bcd8a3ae}.}\par\noindent

\textbf{Setting:} DAPO, 4B, 4K; reference \verb|50|.

\textbf{Observed:} \texttt{G0\_41}, R, strict label \texttt{correct}, 2,538 sampled output tokens.

Box NLL $2.54009\times10^{-6}$; tail10 1.157075; answer-mean NLL 0.138205.

At the observed-arm cutoffs, box $\infty$ gives \emph{accept}; tail $\infty$ gives \emph{accept}.

\begin{Verbatim}[fontsize=\scriptsize,breaklines=true,breakanywhere=true]
Finally, we compute $p(5)$:
$$ p(5) = (5)^2 + 4(5) + 5 $$
$$ p(5) = 25 + 20 + 5 $$
$$ p(5) = 50 $$

\boxed{50}<|im_end|>
\end{Verbatim}

\textbf{Paired response:} \texttt{base}, U, strict label \texttt{truncated}, 4,092 sampled output tokens.

Box NLL --; tail10 --; answer-mean NLL --.

The paired response is rejected as invalid independently of any score cutoff.

\begin{Verbatim}[fontsize=\scriptsize,breaklines=true,breakanywhere=true]
The question asks to put the final answer in \boxed{...}.
The value is 50.

One more thing, could $p(x)$ have a term $C e^x$ that reduces to polynomial? No, $e^x$ is transcendental. Only if field is different, but in standard calculus context it's not.

Let's write down the steps clearly.
1. Assume degree of $p(x)$ is $n$. Show $n=2
\end{Verbatim}

Base states the value 50 during reasoning but continues to truncation; G0 returns a valid final answer. This completion rescue does not show that G0 alone found the mathematical value.

\subsection{Wrong submissions rejected by tail10}

The matching rule yields 68 eligible records; two selected cases illustrate this category.

\paragraph{Case 9: Concurrent chords.}\label{case:audited-09}\mbox{}\par\noindent

{\footnotesize \textbf{Family:} \nolinkurl{dapo_en_graph_v1:bb1af32470bb307377b8b9e99f1ae80ca1c2612a180225752e2f920a15657323}.}\par\noindent

\textbf{Setting:} DAPO, 8B, 4K; reference \verb|225|.

\textbf{Observed:} \texttt{G0\_41}, W, strict label \texttt{wrong}, 3,675 sampled output tokens.

Box NLL 0.016294; tail10 1.912737; answer-mean NLL 0.247560.

At the observed-arm cutoffs, box $2.20845\times10^{-6}$ gives \emph{reject}; tail 1.837020 gives \emph{reject}.

\begin{Verbatim}[fontsize=\scriptsize,breaklines=true,breakanywhere=true]
\boxed{\frac{225}{2}}
$$<|im_end|>
\end{Verbatim}

The reference is 225, whereas the model submits 225/2. Both the tail and box policies reject this answer; this example is not a case of disagreement in favor of tail10.

\paragraph{Case 10: Tournament goals.}\label{case:audited-10}\mbox{}\par\noindent

{\footnotesize \textbf{Family:} \nolinkurl{dapo_en_graph_v1:4078e3e372e8a1370ef697811bfc0ad9a150ae00b40a56712fc5308cb7cb07cf}.}\par\noindent

\textbf{Setting:} DAPO, 8B, 4K; reference \verb|6|.

\textbf{Observed:} \texttt{G0\_41}, W, strict label \texttt{native\_format}, 3,236 sampled output tokens.

Box NLL $8.39056\times10^{-7}$; tail10 2.224559; answer-mean NLL 0.357703.

At the observed-arm cutoffs, box $2.20845\times10^{-6}$ gives \emph{accept}; tail 1.837020 gives \emph{reject}.

\begin{Verbatim}[fontsize=\scriptsize,breaklines=true,breakanywhere=true]
\boxed{3}
$$<|im_end|>
\end{Verbatim}

The submitted 3 differs from the reference 6. Its tail score exceeds the cutoff, while its extremely small box NLL passes the box rule.

\subsection{Low-uncertainty errors that pass}

The matching rule yields 29 eligible records; two selected cases illustrate this category.

\paragraph{Case 11: A modular sum over free integers.}\label{case:audited-11}\mbox{}\par\noindent

{\footnotesize \textbf{Family:} \nolinkurl{dapo_en_graph_v1:73bfaccbdefebf011db90c3352c8b6b25ee9a48e14b6e257252a8544f6e23b60}.}\par\noindent

\textbf{Setting:} DAPO, 8B, 4K; reference \verb|80|.

\textbf{Observed:} \texttt{G0\_41}, W, strict label \texttt{native\_format}, 4,025 sampled output tokens.

Box NLL $7.05911\times10^{-7}$; tail10 1.776868; answer-mean NLL 0.233324.

At the observed-arm cutoffs, box $2.20845\times10^{-6}$ gives \emph{accept}; tail 1.837020 gives \emph{accept}.

\begin{Verbatim}[fontsize=\scriptsize,breaklines=true,breakanywhere=true]
Using a computer or a calculator, we can compute:

- $ \mathcal{S}_7 = \sum_{\substack{n=1 \\ \gcd(n,2,3,5,7)=1}}^{5040} n^2 $
- $ \mathcal{S}_6 = \sum_{\substack{n=1 \\ \gcd(n,2,3,5)=1}}^{720} n^2 $

Then:

$$
\mathcal{S}_7 - \mathcal{S}_6 \mod 1000 = \boxed{964}
$$

---

### Final Answer

$$
\boxed{964}
$$<|im_end|>
\end{Verbatim}

The model confidently repeats 964 instead of the reference 80, and both score rules accept it. Its claim of computer use is generated text and does not establish an external calculation.

\paragraph{Case 12: An angle in a reflected pentagon.}\label{case:audited-12}\mbox{}\par\noindent

{\footnotesize \textbf{Family:} \nolinkurl{dapo_en_graph_v1:4731a42d9126244606e0022d35b0addbbf9586ae8a1d961b5d370d5421a755d6}.}\par\noindent

\textbf{Setting:} DAPO, 8B, 4K; reference \verb|90|.

\textbf{Observed:} \texttt{G0\_41}, W, strict label \texttt{native\_format}, 3,341 sampled output tokens.

Box NLL 0.127593; tail10 1.768990; answer-mean NLL 0.261905.

At the observed-arm cutoffs, box $2.20845\times10^{-6}$ gives \emph{reject}; tail 1.837020 gives \emph{accept}.

\begin{Verbatim}[fontsize=\scriptsize,breaklines=true,breakanywhere=true]
\theta = \cos^{-1} \left( \frac{5}{\sqrt{26}} \right)
\Rightarrow \angle E = \boxed{45^\circ}
$$
\end{Verbatim}

The model submits 45 degrees instead of the reference 90 degrees. Tail10 accepts the error, while the box rule rejects it. The accepted error directly shows that the tail rule is not a perfect detector.

\end{document}